%% file: main.tex
\documentclass[sigconf, nonacm, balance=false]{acmart}
\usepackage{popets}

\setcopyright{popets}
\copyrightyear{YYYY}

\acmYear{YYYY}
\acmVolume{YYYY}
\acmNumber{X}
\acmDOI{XXXXXXX.XXXXXXX}
\acmISBN{}
\acmConference{Proceedings on Privacy Enhancing Technologies}
\usepackage{custompackages}

\begin{document}

\title[DP-SNP-TIHMM]{DP-SNP-TIHMM: Differentially Private, Time-Inhomogeneous Hidden Markov Models for Synthesizing Genome-Wide Association Datasets}

%%%%%%%%%%%%%%%% Authors' Info %%%%%%%%%%%%%%%%%
\author{Shadi Rahimian}
\affiliation{%
  \institution{CISPA Helmholtz Center for Information Security}
  \city{Saarbr{\"u}cken}
  \country{Germany}
}
\email{shadi.rahimian@cispa.de}

\author{Mario Fritz}
\affiliation{%
  \institution{CISPA Helmholtz Center for Information Security}
  \city{Saarbr{\"u}cken}
  \country{Germany}}
\email{fritz@cispa.de}


\renewcommand{\shortauthors}{}

\begin{abstract}
Single nucleotide polymorphism (SNP) datasets are fundamental to genetic studies but pose significant privacy risks when shared. The correlation of SNPs with each other makes strong adversarial attacks such as masked-value reconstruction, kin, and membership inference attacks possible. Existing privacy-preserving approaches either apply differential privacy to statistical summaries of these datasets or offer complex methods that require post-processing and the usage of a publicly available dataset to suppress or selectively share SNPs.

In this study, we introduce an innovative framework for generating synthetic SNP sequence datasets using samples derived from time-inhomogeneous hidden Markov models (TIHMMs). To preserve the privacy of the training data, we ensure that each SNP sequence contributes only a bounded influence during training, enabling strong differential privacy guarantees. Crucially, by operating on full SNP sequences and bounding their gradient contributions, our method directly addresses the privacy risks introduced by their inherent correlations.

Through experiments conducted on the real-world 1000 Genomes dataset, we demonstrate the efficacy of our method using privacy budgets of $\varepsilon \in [1, 10]$ at $\delta=10^{-4}$. Notably, by allowing the transition models of the HMM to be dependent on the location in the sequence, we significantly enhance performance, enabling the synthetic datasets to closely replicate the statistical properties of non-private datasets. This framework facilitates the private sharing of genomic data while offering researchers exceptional flexibility and utility.
\end{abstract}


\maketitle

\input{body/introduction}

\input{body/background}
\input{body/method}

\input{body/experiments}

\input{body/related-work}

\section{Challenges}
While our approach is promising, certain limitations must be considered. First, the effective sequence length that can be modeled is currently bounded by available computational resources, as training HMMs over long SNP sequences remains computationally demanding. Techniques such as HMM merging~\cite{stolcke1994best} offer a promising avenue to scale to longer sequences without retraining from scratch, though their applicability to our framework requires further investigation.

Second, the presence of related individuals in genomic datasets introduces dependencies that may violate the independence assumptions underpinning differential privacy guarantees. This issue is inherent to all differentially private methods applied to genetic data and is not specific to our pipeline. One promising solution is group differential privacy, which adjusts the privacy budget based on an assumed upper bound on the number of closely related individuals (see, e.g., ~\cite{almadhoun2020differential}).

Third, the preprocessing step of SNP selection and the possibility of the emergence of novel variants in larger or more diverse datasets pose privacy risks. Our proposed approach of applying differential privacy to the gradients of locus-dependent sequential models provides a promising path forward. It can be directly applied to full DNA sequences, thereby eliminating the need for SNP selection and mitigating issues arising from emerging or previously unobserved variants.

Overall, while these challenges merit continued exploration, they do not diminish the practical viability of our framework. On the contrary, they open up exciting directions for enhancing scalability and robustness in future work.

\section{Conclusion}
In this work, we present a novel framework for privacy-preserving generation of synthetic genomic data, specifically focusing on the release of complete SNP sequences. By bounding the gradient updates during training, our approach effectively controls the privacy risk associated with linkage disequilibrium and SNP correlations, enabling the release of realistic, sequence-level genomic data under formal differential privacy guarantees.

Our framework introduces a shift in perspective from traditional approaches, which primarily focus on releasing aggregate GWAS statistics or rely on public auxiliary information to determine which SNPs to suppress or disclose. While such methods provide strong utility guarantees within their targeted scope, often optimizing for accurate $p$-values of a small subset of SNPs, they are inherently limited in flexibility. In contrast, our goal is to enable broader exploratory analyses by releasing fully synthetic datasets that retain key statistical signals, without the need for external genomic knowledge or selective SNP suppression.

Although our model is not without limitations, it represents an important step toward scalable and practical solutions for private genomic data sharing. As the field of genomics continues to advance rapidly, so too must our methods for safeguarding privacy. We believe that the direction initiated by this work lays a valuable foundation for future research at the intersection of synthetic data generation, differential privacy, and genomic utility.

\myparagraph{Ethics statement} This study utilizes data from the 1000 Genomes Project, a publicly available resource generated with informed consent~\cite{IGSR} for broad research use. The dataset contains fully anonymized genomic information along with limited demographic metadata, specifically, sex and ethnic/geographic background. No additional personal or identifiable information is included. At no point did our research involve attempts to re-identify individuals or interact with human subjects, and no new data was collected.

Our use of the dataset was solely for evaluating the performance of our differentially private algorithm, with the goal of advancing privacy-preserving genomic analysis. All data use adhered strictly to the terms and ethical guidelines provided by the 1000 Genomes Project. We remain committed to the responsible and ethical handling of sensitive genomic data.

\begin{acks}
This work is partially funded by Medizininformatik-Plattform\\ 
``Privatsph{\"a}ren-schutzende Analytik
in der Medizin" (PrivateAIM), grant No. 01ZZ2316G, and Bundesministeriums f{\"u}r Bildung und
Forschung (PriSyn), grant No. 16KISAO29K.
The work was also supported by ELSA – European Lighthouse on Secure and Safe AI funded by the European Union under grant agreement No. 101070617. Views and opinions expressed are, however, those of the authors only and do not necessarily reflect those of the European Union or European Commission. Neither the European Union nor the European Commission can be held responsible for them. 
\end{acks}

\bibliographystyle{ACM-Reference-Format}
\bibliography{references}

\appendix
\input{body/appendix}

\end{document}

%% file: body/introduction.tex
\section{Introduction}
\label{sec:intro}
Genome-Wide Association Studies (GWAS) are powerful tools in genetics that aim to identify associations between genetic variants and phenotypic traits, such as diseases, physical characteristics, or other biological markers. By analyzing the genetic data of thousands of individuals, GWAS searches the genome for loci, specific positions on chromosomes, where genetic variations are correlated with particular traits. These studies typically involve case-control designs, where the genomes of individuals with a specific trait (cases) are compared to those without it (controls), or quantitative trait designs, which analyze traits that vary across a spectrum, like height or cholesterol levels.

The success of GWAS has revolutionized our understanding of the genetic basis of complex traits and diseases, enabling researchers to identify genetic risk factors for conditions such as Alzheimer's disease, diabetes, and cancer~\cite{uffelmann2021genome}.

The genome can be thought of as a long sequence of nucleotides, with 4 possible nucleobases (A, T, C, or G) at each locus. Single Nucleotide Polymorphisms (SNPs) are the most common type of genetic variation studied in GWAS. A SNP represents a change in a single nucleotide at a specific position in the genome. While individual SNPs may not always directly cause a trait, their statistical correlation with the trait provides clues about nearby causal variants. This is possible because of linkage disequilibrium (LD), the tendency of SNPs near each other on the genome to be inherited together~\cite{reich2001linkage}. 

While LD is a powerful tool for genetic research, it introduces significant privacy challenges. SNPs in LD are correlated, meaning that knowledge of one SNP can reveal information about nearby SNPs. This correlation has been exploited in privacy attacks to infer sensitive genetic information, such as missing value reconstruction attacks~\cite{nyholt2009jim}, kin genomic attacks~\cite{ayday2017inference}, membership inference attacks~\cite{homer2008resolving, shringarpure2015privacy} and more sophisticated attacks that use a combination of all of this information~\cite{humbert2013addressing, deznabi2017inference}.

Differential privacy (DP)~\cite{dwork2006differential} has become a standard and widely adopted framework for ensuring privacy in datasets and statistics derived from them. However, the vast number of SNPs in the human genome, often numbering in the tens of millions~\cite{international2001map, 10002015global}, and their correlations due to linkage disequilibrium pose significant challenges for developing high-utility, differentially private techniques tailored to SNP data.

Existing DP approaches for genome-wide association studies primarily focus on either releasing private statistics from datasets~\cite{fienberg2011privacy, uhlerop2013privacy, johnson2013privacy}, such as the $p$-values of top-$k$ SNPs, or relaxing the definition of DP to account for SNP correlations~\cite{yilmaz2022genomic, humbert2014reconciling, yilmaz2020preserving}, enabling the release of a noisy subset of SNPs. While the first approach restricts researchers to predefined statistics, limiting exploratory analyses, the second approach sacrifices formal DP guarantees and often requires complex pre- and post-processing steps, as well as auxiliary knowledge, such as publicly available linkage disequilibrium patterns. These limitations underscore the need for more robust and flexible solutions to ensure privacy in genomic studies.

Inspired by the state-of-the-art imputation techniques for missing SNPs in genomic datasets (e.g. MaCH~\cite{li2010mach}, Minimac~\cite{das2016next}, Beagle~\cite{ayres2012beagle} and SHAPEIT~\cite{delaneau2014integrating}), we utilize hidden Markov models~\cite{rabiner1986introduction} in our work. These imputation softwares are mostly based on the Li-Stephens~\cite{li2003modeling} model of genetic recombination, which suggests that by training a hidden Markov model(HMM) on SNP sequences from individuals in a dataset, the model can learn to impute the missing SNPs at specific loci in a new individual. 

Our methodology involves training a hidden Markov model end-to-end on SNP sequences from individuals in a dataset using stochastic gradient descent. To ensure privacy during model training, we employ the differentially private stochastic gradient descent (DP-SGD) technique~\cite{abadi2016deep}. By training directly on SNP sequences, our approach effectively addresses locus-dependent linkage disequilibrium, providing privacy guarantees for the entire sequence. Once trained, the HMM can be used to generate differentially private synthetic datasets by sampling from the model. These sanitized synthetic datasets serve as publicly shareable proxies of the original data, enabling the calculation of meaningful statistics while safeguarding the privacy of individuals in the original dataset.

As opposed to the original Li–Stephens model, which employs a time-homogeneous transition scheme, we introduce a time inhomogeneous (locus-dependent) transition model. While a single transition model can capture broad genome-wide patterns, SNP sequences need not exhibit repeating structures that are well described by a uniform model. By allowing locus-specific transitions, we better preserve local correlations and behaviors, leading to a closer match between the samples from our time-inhomogeneous HMM and the original dataset.

% As opposed to the original Li-Stephens model that uses time-homogeneous transition models for their HMM, we propose a time-inhomogeneous (locus-dependent) transition model. This comes from our assumption that although the full genome shows patterns, looking only at the SNP sequences does not necessarily have to show repeating patterns that can be learned by one ubiquitous transition model, and we would rather preserve the locus-specific correlations and behavior. We find that this modification significantly improves the performance of the HMM in matching the original dataset's behavior. 

We run our experiments on SNP sequences from the 1000 Genome project and use the classic genetic distance metrics to measure the closeness of the original population to the synthetic population.  We show that our proposed differentially private time-inhomogeneous hidden Markov model can be sampled to produce a synthetic dataset that mimics the behavior of the non-private dataset at an acceptable privacy regime $(\varepsilon \in [1, 10], \delta = 10^{-4})$.

To summarize, our contributions are:

\begin{itemize}
    \item We present a novel framework for generating synthetic SNP datasets using locus-dependent sequential models trained with differential privacy, enabling the privacy-preserving release of genetic data.  
    \item We introduce the time-inhomogeneous HMM and systematically evaluate its performance across different hidden state sizes ($H$), sample sizes, sequence lengths, and privacy regimes.  
    \item Our method removes the need for post-processing or external public datasets as auxiliary information, thereby streamlining the generation workflow.  
    \item We provide a comprehensive assessment of synthetic data quality using multiple measures, including allele frequency preservation, Nei's genetic distance, correlation structure matching (LD panels), and downstream SNP association analysis.  
    \item We empirically demonstrate how model complexity ($H$) and privacy level ($\varepsilon$) govern the trade-off between utility (e.g., downstream tasks and imputation fidelity) and privacy.  
\end{itemize}

% \begin{itemize}
    % \item Our method provides flexibility by allowing data providers to release synthetic datasets rather than being restricted to predetermined statistics. These datasets can be shared publicly and utilized for exploratory analyses without limitations.
%     \item To the best of our knowledge, we are the first to train a private generative model on SNP sequences in an end-to-end manner, enabling us to produce synthetics datasets to facilitate genome-wide association studies.
%     \item Our method provides flexibility by allowing data providers to release synthetic datasets rather than being restricted to predetermined statistics. These datasets can be shared publicly and utilized for exploratory analyses without limitations.
%     \item Our approach eliminates the need for post-processing or reliance on public datasets as auxiliary information, streamlining the workflow. 
%     \item Through differentially private training on SNP sequences, we address the privacy challenges posed by correlated SNPs, commonly referred to as linkage disequilibrium. 
% \end{itemize}

% In GWAS, SNPs are used as genetic markers because they are abundant, relatively stable across generations, and often associated with traits or diseases. The association between a SNP and a phenotype is measured statistically, often using logistic regression or linear regression models, depending on whether the trait is categorical (e.g., presence or absence of a disease) or continuous (e.g., height or weight).

%% file: body/background.tex
\section{Background}
\label{sec:background}
We begin by providing an overview of single nucleotide polymorphisms (SNPs) and their role in genome-wide association studies (GWAS). Next, we briefly introduce hidden Markov models (HMMs), which serve as a foundational statistical tool in genetic data analysis. Finally, we present an overview of differential privacy, the privacy-preserving framework employed in this work to ensure the confidentiality of SNP datasets.

\subsection{SNP Genome-Wide Association Studies}
\label{sec:bg-snps}
We first begin with some genetic background. Humans have 22 pairs of homologous chromosomes and a pair of sex chromosomes. These chromosomes consist of long sequences of nucleotides, each represented by one of four nucleobases: Adenine (A), Thymine (T), Cytosine (C) or Guanine (G). Each homologous pair consists of one chromosome inherited from the mother and one from the father, with both chromosomes containing the same genes (sequence of nucleotides with specific functions) in the same loci. Collectively, these sequences constitute the human genome, which encapsulates the entirety of an individual's genetic material. There are about 3 billion bases in the human genome, of which an estimated $99.5\%$ is common to all humans. The remaining $0.5\%$ accounts for the genetic variation responsible for individual differences, including traits such as eye color, susceptibility to certain diseases, and other characteristics.

Single nucleotide polymorphisms (SNPs) are the most prevalent form of genetic variation in the human genome, occurring approximately once every 300 nucleobases on average~\cite{kruglyak2001variation}. These variations involve a substitution of a single nucleotide at a specific locus in the DNA sequence. For example, when an A in the reference genome is replaced with a G. We call these different versions of the nucleobase \textbf{alleles}. The \textbf{major allele} is the more frequent nucleobase in the population, and the \textbf{minor allele} the less frequent.

\begin{figure}[H]
    \centering
    \includegraphics[width=0.9\linewidth]{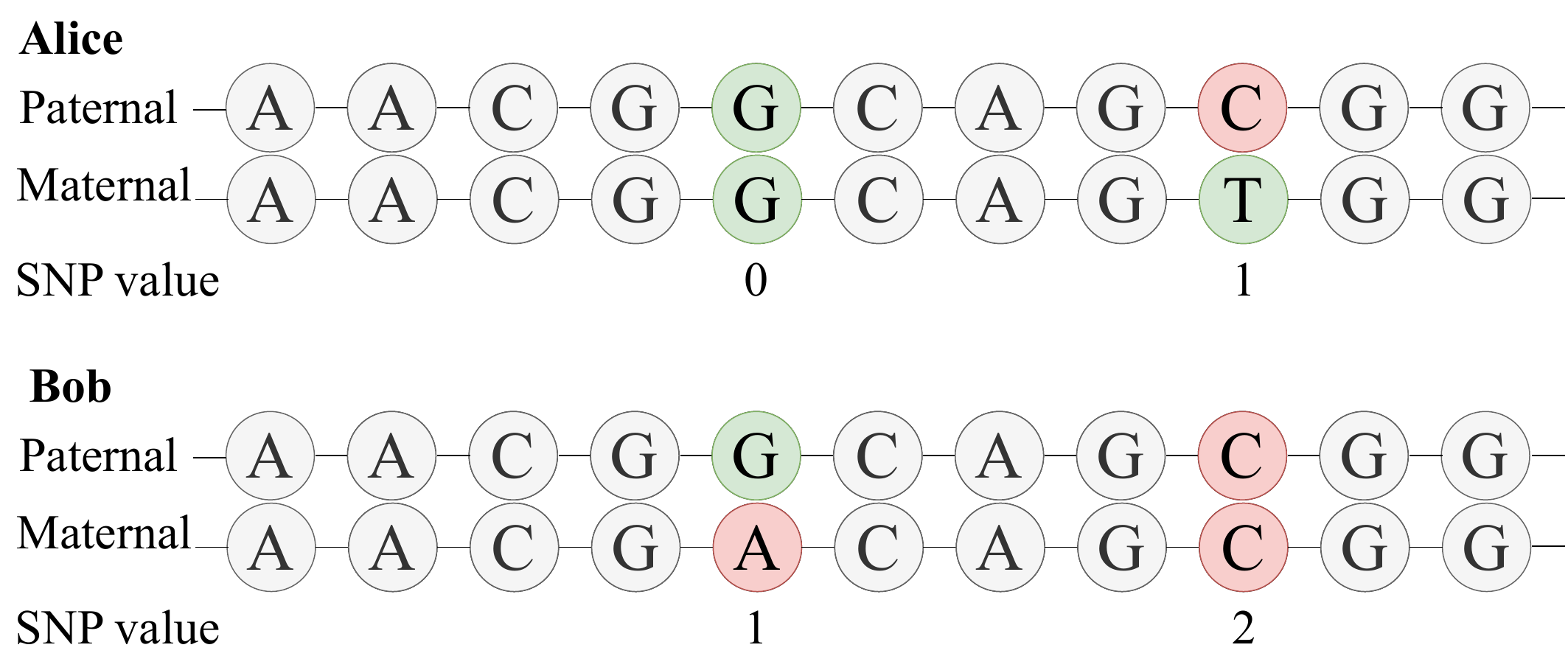}
    \caption{The same segment of a chromosome for Alice and Bob. The major allele is shown in green and the minor allele in red.}
    \label{fig:snp-toy}
\end{figure}

SNP genotypes are commonly represented numerically, with 0 indicating the presence of two major alleles in both homologous chromosomes, 1 representing one major and one minor allele, and 2 indicating two minor alleles in both homologous chromosomes of the individual. Figure~\ref{fig:snp-toy} provides an example for a small sequence of the genome.
It has been shown that the association between SNPs is non-random, with SNPs physically closer to each other being more likely to be inherited together. This correlation between alleles in a population is formally known as \textbf{linkage disequilibrium} or LD~\cite{stephens2001haplotype}. These correlation patterns can be complex and go beyond simple pair-wise dependencies and are affected by factors such as population distribution and isolation, region of origin, and position on the genome~\cite{li2003modeling, reich2001linkage, macdonald1991complex, stephens2001haplotype}.

We can utilize the single nucleotide polymorphisms data to find associations of genes with phenotypes (traits) in what is known as genome-wide association studies, or GWAS~\cite{uffelmann2021genome}. These studies involve systematically scanning the genome of large populations to detect SNPs that differ in allele frequency between case and control groups or along a continuous trait distribution and use statistical techniques to pinpoint SNPs significantly correlated with a phenotype. 

% \begin{itemize}
%     \item what are GWAS X
%     \item what is SNP X
%     \item major minor allele X
%     \item linkage disequilibrium X
% \end{itemize}

%%%%%%%%%%%%%%%%%%%%%%%%%%%%%%%%%%%%%%%%%%%%%%%%%%%%%%%%%%%%%%%%%%%%%%%%%%%%%%%%%

\subsection{Hidden Markov Models}
\label{sec:bg-hmm}
\begin{figure}[H]
    \centering
    \includegraphics[width=0.7\linewidth]{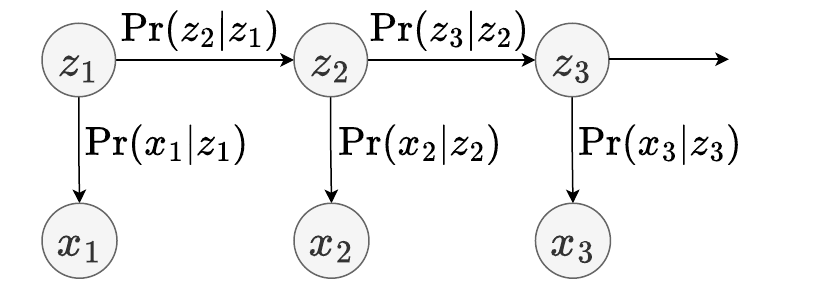}
    \caption{Causal graph of a hidden Markov model.}
    \label{fig:hmm-toy}
\end{figure}

Hidden Markov Models (HMMs)~\cite{rabiner1986introduction} are statistical models used to represent systems that transition between hidden states over time, with observable outputs dependent on those states. Figure~\ref{fig:hmm-toy} shows the probabilistic dependencies of an HMM, where $x_i$s are the observed outcomes at $t=i$ and the unknown processes that result in observables are captured in hidden states $z_i$s. The sequence has a finite length of $L$, so $\mathbf{z}=\{z_1, z_2, ...,z_L\}$ and $\mathbf{x}=\{x_1, x_2, ..., x_L\}$, and each hidden state can take one of the finite set of $H$ values, that is, $h \in \{1, 2, ..., H\}$. HMM is characterized by three sets of trainable parameters:
\begin{itemize}
    \item The state prior $\pi_{z_1=h} :=p(z_1=h)$, which is the probability of starting in state $h$.
    \item The transition model $\tau_{z_i=h', z_{i+1}=h} := \Pr(z_{i+1}=h|z_{i}=h')$, represents the probability of jumping from a hidden state $h'$ to a hidden state $h$.
    \item The emission model $\epsilon_{z_i=h}(x_i) := \Pr(x_i|z_i=h)$ captures the probability of generating observable $x_i$ when the system is in hidden state $h$.
\end{itemize}

The likelihood of the model for an observed sequence $\mathbf{x}$ is given by $\Pr(\mathbf{x}; \theta)$ where $\theta$ constitutes all the trainable parameters. We can calculate this likelihood efficiently, using dynamic programming in what is known as the \textbf{forward algorithm}: 
{\small
\begin{align*}
    &\alpha_k(z_k):= \Pr(z_k, x_{1:k})=\sum_{z_{k-1}=1}^{H}\Pr(z_{k-1}, z_k, x_{1:k})\\
    &=\sum_{z_{k-1}=1}^{H}\Pr(x_k|z_{k-1}, z_k, x_{1:k-1})\Pr(z_k|z_{k-1}, x_{1:k-1})\\
    &\times\Pr(z_{k-1},x_{1:k-1})\\
    &= \epsilon_{z_k}(x_k) \sum_{z_{k-1}=1}^{H}\tau_{z_{k-1}, z_k}\alpha_{k-1}(z_{k-1});\\ 
    &\alpha_1(z_1) = \pi_{z_1}\epsilon_{z_1}(x_1)
\end{align*}
}
where $x_{1:k}$ denotes the observed sequence from $t=1$ till $t=k$ and we use conditional independencies of HMM to arrive at the last line. This algorithm is prone to underflow due to multiplying a long chain of small probabilities, so in practice, the above equations are converted to the log domain. The forward algorithm requires ${\Theta}(H^2L)$ operations. The final likelihood of the complete sequence can be calculated as the summation over all the possible hidden states for the last $\alpha_L$:
\begin{equation}
\label{eq:final-likelihood}
    \Pr(\mathbf{x}; \theta) = \sum_{z_L=1}^H \alpha_{L}(z_L)
\end{equation}

\subsection{Differential Privacy}
\label{sec:bg-dp}
Differential privacy (DP)~\cite{dwork2006differential} is a rigorous mathematical framework that ensures the privacy of individuals in a dataset by guaranteeing that the outcome of a computation is not significantly affected by the inclusion or exclusion of any single individual's data. 

\begin{definition}[Differential Privacy (DP)~\cite{dwork2006differential}]
\label{definition:dp}
A randomized mechanism $\mathcal{M}$ satisfies $(\varepsilon, \delta)$-differential privacy if, for any two neighboring datasets $D$ and $D'$ differing in at most one element, and for any subset of possible outputs $S$:
\begin{equation}
\Pr[\mathcal{M}(D) \in S] \leq e^\varepsilon \Pr[\mathcal{M}(D') \in S] + \delta,\nonumber
\end{equation}
\end{definition}
where $\varepsilon$ quantifies the privacy loss, with smaller values providing stronger privacy guarantees, and $\delta$ represents the probability of the mechanism failing to provide $\varepsilon$-level privacy. 

In the bounded differential privacy model, $D'$ is derived from $D$ by modifying the value of exactly one data point. In contrast, the unbounded DP defines $D'$ as differing from $D$ by the addition or removal of a single data point. In this paper, we adopt the \textbf{unbounded} differential privacy framework exclusively.  

A common method to ensure $(\varepsilon, \delta)$-DP is the Gaussian mechanism, which adds noise sampled from a Gaussian distribution to the output of a function. To apply the Gaussian mechanism, we first define the global sensitivity of the function.

\begin{definition}[$L_2$ Global Sensitivity]
\label{def:lp-sensitivity}
For an arbitrary function $f: \mathcal{D} \rightarrow \mathbb{R}^k$, and all possible neighboring datasets $D$ and $D'$, the $L_2$-sensitivity of $f$ is defined as:
\begin{equation}
    \Delta_2 f = \max_{D, D'} \|f(D) - f(D')\|_2,\nonumber
\end{equation}
where $\|.\|_2$ denotes the $L_2$-norm.
\end{definition}

\begin{theorem}[Gaussian Mechanism~\cite{Dwork2014book}]
\label{def:gaussian-mechanism}
Let $f$ be a function with $L_2$-sensitivity $\Delta_2 f$. The \textit{Gaussian mechanism} defines a randomized algorithm $\mathcal{M}(D)$ that returns:
\begin{equation}
\mathcal{M}(D) = f(D) + \mathcal{N}(0, \sigma^2 I),\nonumber
\end{equation}
where $\mathcal{N}(0, \sigma^2 I)$ is a multivariate Gaussian distribution with zero mean and covariance $\sigma^2 I$. The standard deviation $\sigma$ is calibrated based on the target privacy guarantees, and in particular, scales proportionally with $\Delta_2 f$.
\end{theorem}

% The Gaussian mechanism admits tight analysis under R\'enyi Differential Privacy (RDP)~\cite{mironov2017renyi}, which provides a convenient framework for composing multiple mechanisms and converting the resulting guarantees into $(\varepsilon, \delta)$-DP bounds. In practice, noise calibration is often performed in the RDP framework and then translated into the desired $(\varepsilon, \delta)$ parameters.

Differential privacy is immune to post-processing, meaning that any transformation of the output of a differentially private mechanism $\mathcal{M}$ cannot degrade its privacy guarantees.
\begin{theorem}[Post-Processing Immunity~\cite{Dwork2014book}]
\label{theorem:postprocessing}
If a mechanism $\mathcal{M}$ satisfies $(\varepsilon, \delta)$-differential privacy, and $g$ is any arbitrary function, then the composition $g(\mathcal{M}(D))$ also satisfies $(\varepsilon, \delta)$-differential privacy.
\end{theorem}

%% file: body/method.tex
\section{Method}
\label{sec:method}
In this section, we explain our proposed method, which uses differential privacy to privately train our improved hidden Markov model.

\myparagraph{System Model} We focus on a centralized setting where a trusted data collector holds genomic information and SNP sequences from individuals. This is a practical assumption as, with the current technology, genome sequencing is only possible through sequencing services such as medical and research centers ~\citep[e.g,][]{broadinstitute, NIHUndiagnosed, GenomicsEngland} or commercial sequencing platforms \citep[e.g.,][]{Nebula, Veritas, 23andMe}. 

\myparagraph{Threat Model}
The attacker is assumed to have full access to the trained model and its outputs. 
% The objective of the data collector is to release private genetic information that enables meaningful exploratory genomic studies, while simultaneously safeguarding the privacy of the individuals whose data contributes to the dataset. 

% \begin{figure}[ht!]
\begin{figure*}[ht!]
    \centering
    \includegraphics[width=0.7\linewidth]{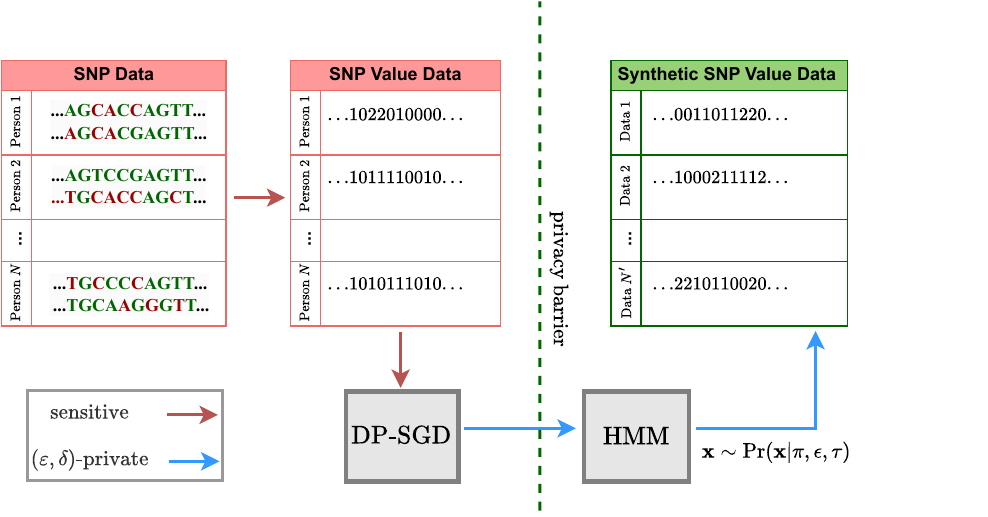}
    \caption{Workflow of our proposed framework. Any component after the privacy barrier is $(\varepsilon, \delta)-$DP.}
    \label{fig:workflow}
\end{figure*}
% \end{figure}

\myparagraph{Privacy issue of SNP datasets}
Consider the sum of SNP values across the dataset at each locus. These counts can be transformed into allele frequencies and subsequently used in downstream analyses, such as top-$k$ associated SNP selection, a core component of genome-wide association studies. For a single locus, the addition or removal of one individual changes the count by at most $2$, yielding both an $L_1$ sensitivity and an $L_2$ sensitivity of $2$. However, since modifying an individual's data simultaneously affects all loci in a sequence of length $L$, the overall sensitivities scale with $L$: the global $L_1$ sensitivity is $2L$, while the global $L_2$ sensitivity is $2\sqrt{L}$. Consequently, a naive differentially private mechanism that perturbs each locus independently would require noise calibrated to these inflated sensitivities, leading to outputs with a vanishing signal-to-noise ratio and little meaningful information.

\myparagraph{Our proposed solution}
Considering this challenge, our goal is to ensure the privacy of SNP datasets while maximizing flexibility for researchers, which is crucial given the exploratory nature of many topics in genomics.

HMMs form the foundation of several state-of-the-art SNP imputation methods and tools~\cite{li2010mach, das2016next, ayres2012beagle, delaneau2014integrating}, primarily leveraging the Li-Stephens~\cite{li2003modeling} model of genetic recombination to impute missing SNPs in individual datasets. In this work, we propose, for the first time, using HMMs to generate synthetic SNP datasets. Figure~\ref{fig:workflow} illustrates the workflow of our proposed approach, which we detail further in the following.

\myparagraph{HMM model and training:}  As discussed in Section~\ref{sec:bg-hmm}, HMMs effectively capture complex and unknown sequence correlations within their hidden states, leveraging their probabilistic graph structure. In our context, the observable outcomes are SNP sequences, where at each locus we have discrete outcomes $x_i\in\{0,1,2\}$. Correlations between loci are encoded in the hidden states, with the number of hidden states $H$ treated as a hyperparameter. 

The state prior, transition model, and emission model are matrices with values reflecting the probabilities and are learned during the training of the model. 

Traditionally, the transition probabilities of an HMM are time-homogeneous, meaning that the transitions between hidden states do not depend on the time $t$ in the sequence, that is, $\forall i:\   \Pr(z_{i+1}=h|z_i=h') = \Pr (h|h')= \tau_{h'h}$. Since our goal is not to learn repeating patterns throughout the SNP sequence and we would rather preserve the locus-specific correlations and behavior, we suggest using a \textbf{time-inhomogeneous transition model}: $\tau_{h'h}(i)= \Pr(z_{i+1}=h|z_i=h')$. The time-inhomogeneous HMM can be represented by a sequence of time-dependent (in our context, dependent on the locus in the SNP sequence) transition matrices. For a sequence length of $L$, we have: 
% {\small
\begin{align*}
    &\mathcal{T} = [\tau(1), \tau(2), ..., \tau(L-1)],\\
    &\forall i: \tau(i) = \begin{bmatrix}
    \tau_{11} (i) & \tau_{12}(i), & \cdots & \tau_{1H}(i)\\
    % \tau_{21} (i) & \tau_{22}(i), & \cdots & \tau_{2H}(i)\\
    \vdots\\
    \tau_{H1} (i) & \tau_{H2}(i), & \cdots & \tau_{HH}(i)\\        
    \end{bmatrix}_{H\times H}    
\end{align*}
% }

We keep the emission models homogeneous over the sequence, that is, $\Pr(x_i|z_i=h) = \Pr(x_i|h) = \epsilon_h(x_i)$ and since the observable outcomes are discrete, we have:
% {\small
\begin{align*}
\mathcal{E}= \begin{bmatrix}
  \epsilon_1(x_i = 0)& \epsilon_1(x_i=1)& \epsilon_1(x_i=2)\\ 
  % \epsilon_2(x_i = 0)& \epsilon_2(x_i=1)& \epsilon_2(x_i=2)\\ 
  \vdots\\
  \epsilon_H(x_i = 0)& \epsilon_H(x_i=1)& \epsilon_H(x_i=2)\\ 
\end{bmatrix}_{H\times3}
\end{align*}
% }

The training process, outlined in Algorithm~\ref{alg:training}, minimizes the negative log-likelihood of SNP sequences. Gradients with respect to model parameters $(\pi, \mathcal{T}, \mathcal{E})$ are calculated (using e.g. Pytorch's autograd) and updated using stochastic gradient descent (SGD). To ensure privacy, we employ DP-SGD~\cite{abadi2016deep}, which clips gradients by their $l_2$-norm to bound global sensitivity and applies the Gaussian mechanism (Theorem~\ref{def:gaussian-mechanism}). This guarantees differential privacy for the trained model. The overall privacy budget across training epochs is tracked using the Rényi Differential Privacy (RDP) accountant~\cite{mironov2019r, abadi2016deep}. By training the model on entire SNP sequences and bounding gradients globally, local SNP dependencies and linkage disequilibrium are inherently addressed.

% \mario{the description is not very explicit on how the gradients are computed and in which order they are updated. do you think this is trivial, common knowledge, obvious? or should there be more details added? obviously also the promise to release code would help to some extend.}

\begin{algorithm}[tb]
   \caption{Differentially Private Time-inhomogeneous HMM}
   \label{alg:training}
% {
% {\small
\begin{algorithmic}
   \State {\bfseries Input:} Dataset $\mathcal{X} \in \{0, 1, 2\}^ {N\times L}$ of $N$ samples of length $L$, collection of learnable parameters $\theta$ (state prior $\pi_{1\times H}$, emission matrix $\mathcal{E}_{H\times 3}$, transition matrix $\mathcal{T}_{H\times H\times L}$), batch size $B$, gradient clipping bound $C$, number of epochs $T$, learning rate $\eta_t$.
   \State Initialize the elements of probability matrices $\pi, \mathcal{E}, \mathcal{T}$.
   \For{$t = 1 \rightarrow T$}
   % \IF{$x_i > x_{i+1}$}
   % \STATE Swap $x_i$ and $x_{i+1}$
   \State Take random data points $B_t$ with sampling probability $B/N$
   \For{each $n\in B_t$}
   \State \textbf{Forward algorithm}
   \State $\alpha_{h, 1} = \pi_h \epsilon_h(x_1)$
   \For{$l = 2 \rightarrow L$}
   \State $\alpha_{h, l} = \epsilon_h(x_l) \sum_{h'} \tau_{h'h}(l-1) \alpha_{h', l-1}$
   \EndFor
   \State $\mathcal{L} (\theta_t, \mathbf{x}^n) = - \log \sum_h \alpha_{h,L}$
   \State \textbf{Compute gradient}
   \State $\mathbf{g_t}(\mathbf{x}^n)\leftarrow \nabla_{\theta_t}\mathcal{L}(\theta_t, \mathbf{x}^n)$
   \State \textbf{Clip gradient}
   \State $\mathbf{\bar g}_t(\mathbf{x}^n) \leftarrow \mathbf{g}_t(\mathbf{x}^n)/\max(1, \frac{\|\mathbf{g}_t(\mathbf{x}^n)\|_2}{C})$
   \EndFor
   \State \textbf{Add noise}
   \State $\mathbf{\tilde g}_t\leftarrow (\sum_n \mathbf{\bar g}_t(\mathbf{x}^n) + \mathcal{N} (0, \sigma^2C^2\mathbf{I}))$
   \State \textbf{Descent}
   \State $\theta_{t+1} \leftarrow \theta_t - \eta_t \mathbf{\tilde g}_t$
   \EndFor
   \State {\bfseries Output:} Final, private model parameters $\theta_T$ and overall privacy cost $(\varepsilon, \delta)$ calculated via Renyi-DP accountant~\cite{mironov2019r, abadi2016deep}.
   % \UNTIL{$noChange$ is $true$}
\end{algorithmic}
% }
\end{algorithm}
\myparagraph{Synthetic dataset} After training, the model satisfies DP guarantees. By the post-processing immunity of DP (Theorem~\ref{theorem:postprocessing}), any output derived from the trained model also adheres to these guarantees. We propose generating sanitized synthetic datasets by sampling sequences from the trained HMM.

To sample a sequence of length $L$: 
\textbf{1) Initialize:} Select an initial hidden state $z_1$ using the learned state prior $\pi$. \textbf{2) Emission Sampling:} Sample $x_1$ from the emission probabilities $\epsilon_{z_1}(x)$. \textbf{3) Transition Sampling:} Sample $z_2$ from the learned transition matrix $\tau(1)$. \textbf{4) Repeat:} For $i = 2, \dots, L$, sample $x_i$ from $\epsilon_{z_i}(x)$ and $z_{i+1}$ from $\tau(i)$.

% \begin{itemize}
%     \item \textbf{Initialize:} Select an initial hidden state $z_1$ using the learned state prior $\pi$.
%     \item \textbf{Emission Sampling:} Sample $x_1$ from the emission probabilities $\epsilon_{z_1}(x)$.
%     \item \textbf{Transition Sampling:} Sample $z_2$ from the learned transition matrix $\tau(1)$.
%     \item \textbf{Repeat:} For $i = 2, \dots, L$, sample $x_i$ from $\epsilon_{z_i}(x)$ and $z_{i+1}$ from $\tau(i)$.
% \end{itemize}

% This approach ensures the synthetic sequences preserve the statistical and temporal properties of the original SNP dataset while maintaining DP guarantees.

%% file: body/experiments.tex
\section{Experiments}
\label{sec:experiments}
In this section, we first introduce the dataset and the evaluation metrics used to assess the performance of our hidden Markov models. We then describe our differential privacy baseline, namely the generalized randomized response mechanism. To establish a reference point, we conduct preliminary experiments with non-private HMMs, highlighting their baseline performance and the improvements gained through our proposed time-inhomogeneous model. Finally, we present our core experiments, in which we combine differential privacy with the time-inhomogeneous HMM and evaluate the quality of the resulting synthetic private dataset.

\subsection{Dataset}
For our experiments, we use the integrated phased biallelic SNP dataset of the 1000 Genomes Project\footnote{\href{https://ftp.1000genomes.ebi.ac.uk/vol1/ftp/data_collections/1000_genomes_project/release/20190312_biallelic_SNV_and_INDEL/}{1000 genomes project data collections}}~\cite{fairley2020international}. This dataset contains the genetic variations of 2,548 individuals in a biallelic (major/minor allele) variant call format (VCF). Since this is a public dataset and the aim of the project is to provide reference panels for other studies, no phenotype or label is included. \textbf{In fact, there are currently no large and publicly available SNP datasets that come with characteristic labels. This directly stems from the privacy concerns for such datasets and highlights the urgent need to provide privacy solutions for these types of data.}

We use python's \texttt{scikit-allel\footnote{\href{https://scikit-allel.readthedocs.io/en/stable/}{scikit-allel}}} package to pre-process and handle data. Firstly, \textit{singletons} are removed from the dataset. These are the loci on the genome where only one individual in the dataset registers for a variation. We remove these loci since no correlation can be learned from only one datapoint by our models. Lastly, we convert the major/minor allele type of the diploid to an alternate total count of 0, 1, or 2 for two major alleles, one major and one minor allele, and two minor alleles, respectively. 

% \mario{strictly speaking this preprocessing leaks information of the data. but I guess - it is for the purpose of this study - and should be dealt with differently in deployment. do we want to comment on this?}

%%%%%%%%%%%%%%%%%%%%%%%%%%%%%%%%%%%%%%%%%%%%%%%%%%%%%%%%%%%%%%%
%%%%%%%%%%%%%%%%%%%%%%%%%%%%%%%%%%%%%%%%%%%%%%%%%%%%%%%%%%%%%%%
\subsection{Performance Measures}

The lack of labels for public SNP datasets is a challenge that the community is facing, so we employ commonly used metrics to evaluate both the fidelity and generalizability of our synthetic SNP sequence generation. Statistical fidelity ensures the synthetic dataset closely resembles the real dataset, while generalizability verifies that the method does not merely memorize the training data but remains robust in novel scenarios.

To assess statistical fidelity, we compute minor allele frequencies at each SNP locus and use them to calculate population-level distances (\textit{Euclidean, Manhattan}, and \textit{Nei's} genetic distance) between the real and synthetic datasets. 

For generalizability, we analyze the histogram of Euclidean distances between each synthetic record and its closest neighbor in the real dataset. A low frequency of very small distances indicates reduced memorization of the training data.

% Since our dataset contains no label or biological plausibility, we need to use metrics that can measure the fidelity as well as the generalizability of our synthetic SNP sequence generation. We inspect statistical fidelity to ensure that our synthetic dataset resembles the real dataset. Generalizability measures are to ensure that our method does not only memorize the training dataset but is robust in novel scenarios. 

\subsubsection{Frequency}
Frequency of alleles in a population is one of the most fundamental properties that can be studied. For population $A$, the frequency of the minor allele $m$ at locus $i$ is defined as:
\begin{equation}
   f^m_{A, i} = \frac{1\times n^{mM}_{i} + 2\times n^{mm}_{i}}{2\times N_A}
   \label{eq:freqm}
\end{equation}
where $n^{mM}_{i}$ is the number of individuals with one minor allele at locus $i$, $n^{mm}_{i}$ is the number of individuals with two minor alleles at locus $i$, and $2\times N_{A}$ is the total number of alleles across $N$ diploid individuals observed at each locus. 
The frequency of the major allele $M$ can similarly be calculated as:
\begin{equation}
   f^M_{A, i} = \frac{1\times n^{mM}_{i} + 2\times n^{MM}_{i}}{2 \times N_A}\nonumber
\end{equation}
and we have $\forall i: f^m_{A,i} + f^M_{A,i} = 1\ \text{and}\  0\leq f^m_{A,i}, f^M_{A,i} \leq1$. So we can compare the minor or major allele frequencies between the two populations interchangeably. For consistency, throughout our paper, we always calculate the minor allele frequencies.  
    
\subsubsection{Euclidean Distance}
Calculating the frequencies at each locus is helpful; however, we might want to have a measure of distance across the whole SNP sequence. The normalized Euclidean distance between two populations $A$ and $B$ is defined as:
{%
\begin{eqnarray}
    D_{Eu} (A,B) = \sqrt{\frac{1}{L}\sum_{i=1}^L(f^m_{B, i} - f^m_{A,i})^2} = \sqrt{\frac{1}{L}\sum_{i=1}^L(f^M_{B,i} - f^M_{A,i})^2}\nonumber
    \label{eq:euclidean-dist}
\end{eqnarray}}
where $L$ is the length of the SNP sequence and $f_i$ is the frequency at locus $i$. The normalization factor $\frac{1}{L}$ makes sure that the distance is always between 0 and 1. As shown, this metric is symmetric in the choice of major or minor allele.

\subsubsection{Czekanowski (Manhattan) Distance}
Another useful metric to inspect is the Czekanowski or Manhattan distance, which also summarizes the distance between two sequences. The normalized Manhattan distance between two populations $A$ and $B$ is defined as:
\begin{eqnarray}
    D_{Cz}(A,B) = \frac{1}{L}\sum_{i=1}^L |f^m_{B,i} - f^m_{A,i}| = \frac{1}{L} \sum_{i=1}^L|f^M_{B,i} -f^A_{A,i}|\nonumber
\end{eqnarray}
where again $L$ is the length of the SNP sequence and $f_i$ is the frequency at locus $i$. This metric is also normalized between 0 and 1 and is symmetric with respect to the choice of major or minor allele.

\subsubsection{Nei's Standard Genetic Distance~\cite{nei1972genetic, katada2004nei}}
One of the most widely used and evolutionarily meaningful measures of genetic divergence between populations is Nei’s standard genetic distance. This metric is probability-based and reflects the likelihood that two alleles, randomly drawn from two different populations, are identical in state. Unlike Euclidean or Manhattan distances, which quantify direct differences in allele frequencies, Nei’s distance incorporates both between-population divergence and within-population similarity. Notably, under assumptions of genetic drift and mutation, Nei's genetic distance increases approximately linearly with time, making it particularly suitable for modeling evolutionary divergence.

The probability of two randomly chosen alleles from population $A$ being the same allele (either minor or major) at locus $i$ is $p_{A,i} = (f^m_{A,i})^2 + (f^M_{A,i})^2$ and it is $p_{B,i} = (f^m_{B,i})^2 + (f^M_{B,i})^2$ for population $B$. The probability of identity when one allele is chosen from population $A$ and one is chosen from population $B$ is $p_{AB,i} = f^m_{A,i}f^m_{B,i} + f^M_{A,i}f^M_{B,i}$.
The normalized identity of genes between $A$ and $B$ at locus $i$ is defined as:
\begin{equation}
I_i = \frac{p_{AB,i}}{\sqrt{p_{A,i}p_{B,i}}}\nonumber
\end{equation}
where, $I_i=1$ if the two populations have the same alleles in identical frequencies, and $I_i=0$ if they have no common allele. 
The genetic distance between $A$ and $B$ over all loci is defined as:
\begin{equation}
    D_{Nei}(A,B) = - \ln \frac{P_{AB}}{\sqrt{P_{A} P_{B}}}\nonumber
\end{equation}
where $P_{A} = \sum_{i=1}^L p_{A,i}, P_B = \sum_{i=1}^L p_{B,i}$ and $P_{AB} = \sum_{i=1}^L p_{AB,i}$. When the allele frequencies in the two populations are identical, we have $D_{Nei} = -\ln(1) = 0$, and the value approaches infinity as the dissimilarities between the populations grow. Notice that Nei's standard genetic distance does not satisfy the triangle inequality of a metric. This distance is also symmetric with respect to the choice of minor and major alleles. 

\subsubsection{Euclidean Distance to the Closest Record (DCR)}
So far, our utility measures have covered methods that can be used to measure the similarity of the synthetic dataset to the real dataset. To measure the generalizability of the synthetic datasets, it is customary (e.g., in~\cite{zhao2021ctab, sivakumar2023generativemtd, virtualdatalab, mostlyai}) to measure the distance of each synthetic sample to its closest record in the real dataset. The objective is not to have too many very low values (identical or very similar records), as it indicates memorization of the training set. The normalized $l_2$ distance between two records $a$ and $b$ over $L$ SNPs is defined as:
\begin{equation}
    d_{l_2}(a,b) = \sqrt{\frac{1}{4L}\sum_{i=1}^L(s_{a,i} - s_{b,i})^2}\nonumber
\end{equation}
where $s_i\in\{0,1,2\}$ is the SNP score at locus $i$ and the distance is scaled such that it has a range of $[0,1]$. So we have $\text{DCR}(a) = \min_b d_{l_2}(a,b)$. 

%%%%%%%%%%%%%%%%%%%%%%%%%%%%%%%%%%%%%%%%%%%%%%%%%%%%%%%%%%%%%%%%%%%%%%%
%%%%%%%%%%%%%%%%%%%%%%%%%%%%%%%%%%%%%%%%%%%%%%%%%%%%%%%%%%%%%%%%%%%%%%%
\subsection{Baseline}
\label{sec:baseline}
As a baseline, we select a local differential privacy (LDP) approach, as it provides the most comparable differential privacy framework to our proposed pipeline and is commonly used as a baseline in DP research for this type of dataset~\citep[e.g.,][]{jiang2022reproducibility, yilmaz2022genomic}. Our method generates a synthetic dataset that has the original SNP sequence length, aligning with the output of an LDP mechanism. Specifically, in an LDP framework, each feature of every record is perturbed to introduce uncertainty, thereby ensuring a quantifiable degree of deniability for individual contributions. In Appendix~\ref{sec:app-baseline}, we provide a brief overview of LDP and describe the specific mechanism used in our paper, that is, the \textit{generalized randomized response (GRR)}.

\begin{figure*}[ht!]
    \centering
    \begin{minipage}[b]{0.33\textwidth}
        \centering
        \includegraphics[width=\textwidth]{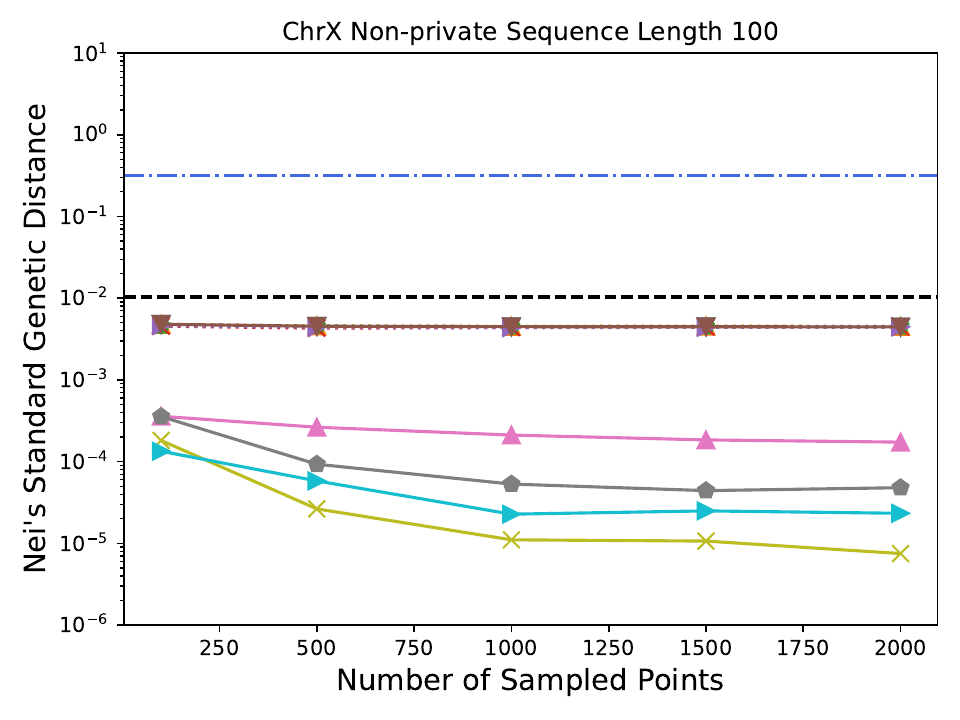}
    \end{minipage}%
    \begin{minipage}[b]{0.33\textwidth}
        \centering
        \includegraphics[width=\textwidth]{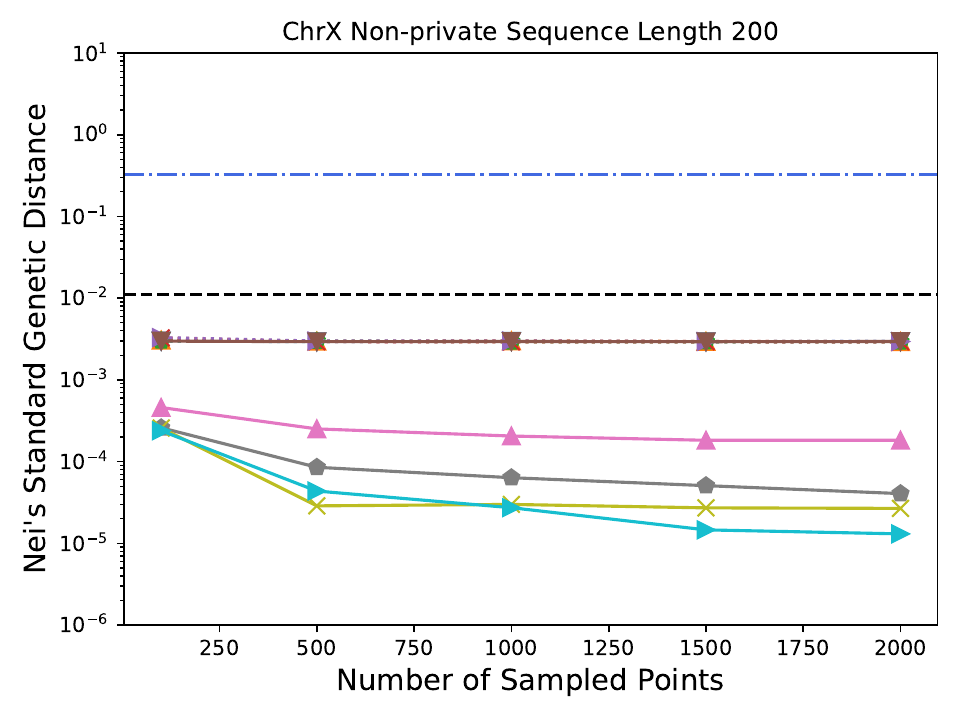}
    \end{minipage}%
    \begin{minipage}[b]{0.33\textwidth}
        \centering
        \includegraphics[width=\textwidth]{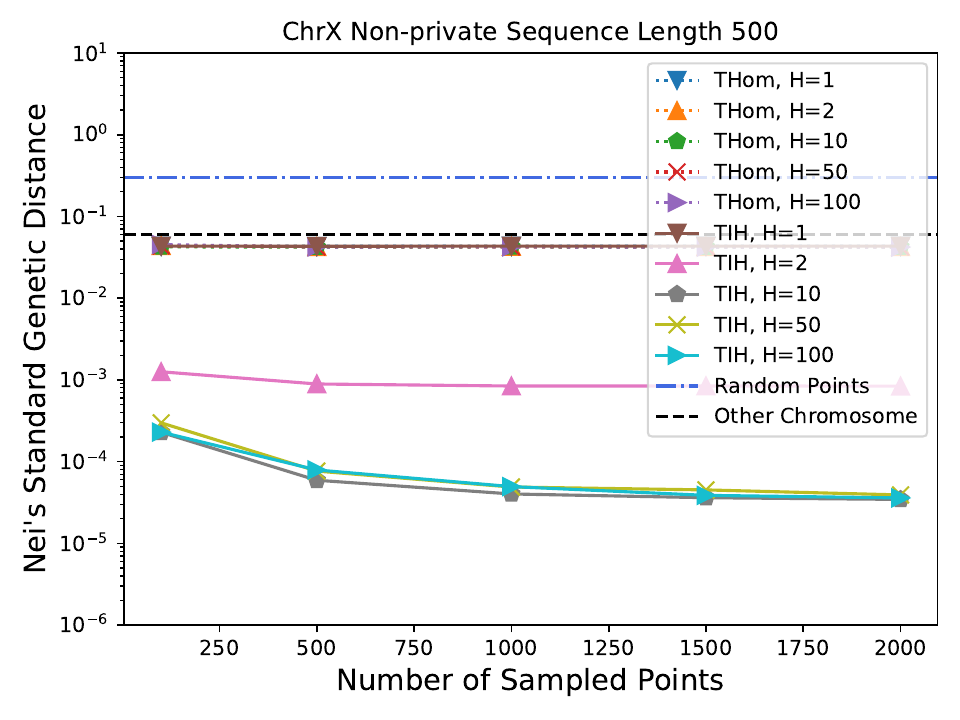}
    \end{minipage}
    
    \caption{Nei's genetic distance between the training data (chromosome X) and synthetic dataset for the time-homogeneous (THom) and time-inhomogeneous (TIH) models with different number of hidden state $H$.}
    \label{fig:distances_Neis_chrx}
\end{figure*}

\begin{figure*}[ht!]
    \centering
    \hspace{-0.4cm}
    \begin{minipage}[b]{0.33\textwidth}
        \centering
        \includegraphics[width=\textwidth]{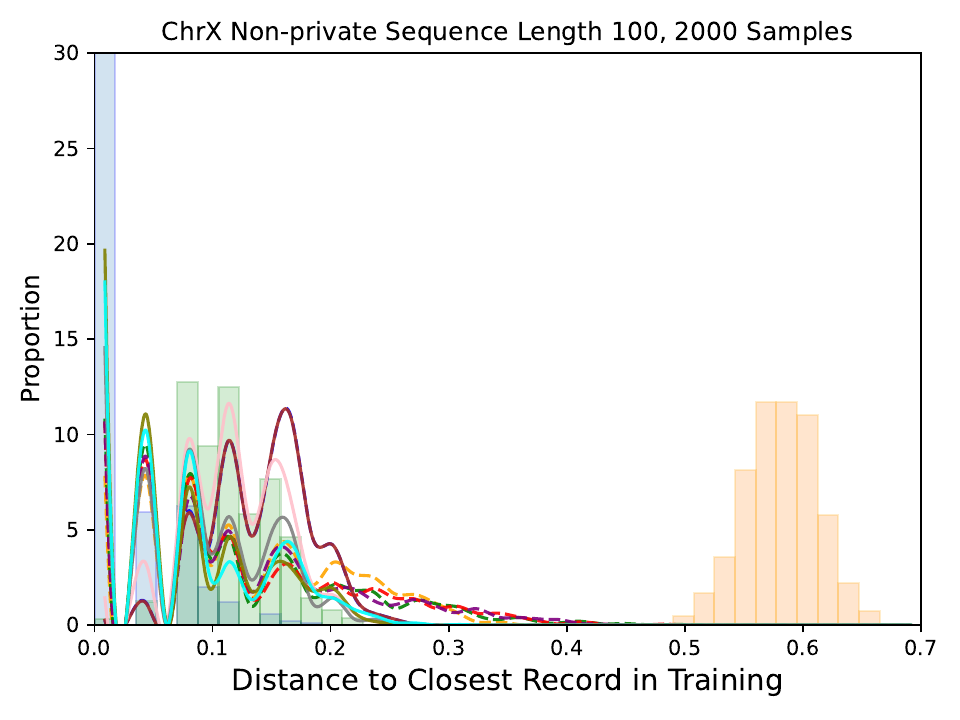}
    \end{minipage}%
    \hspace{-0.30cm}
    \begin{minipage}[b]{0.33\textwidth}
        \centering
        \includegraphics[width=\textwidth]{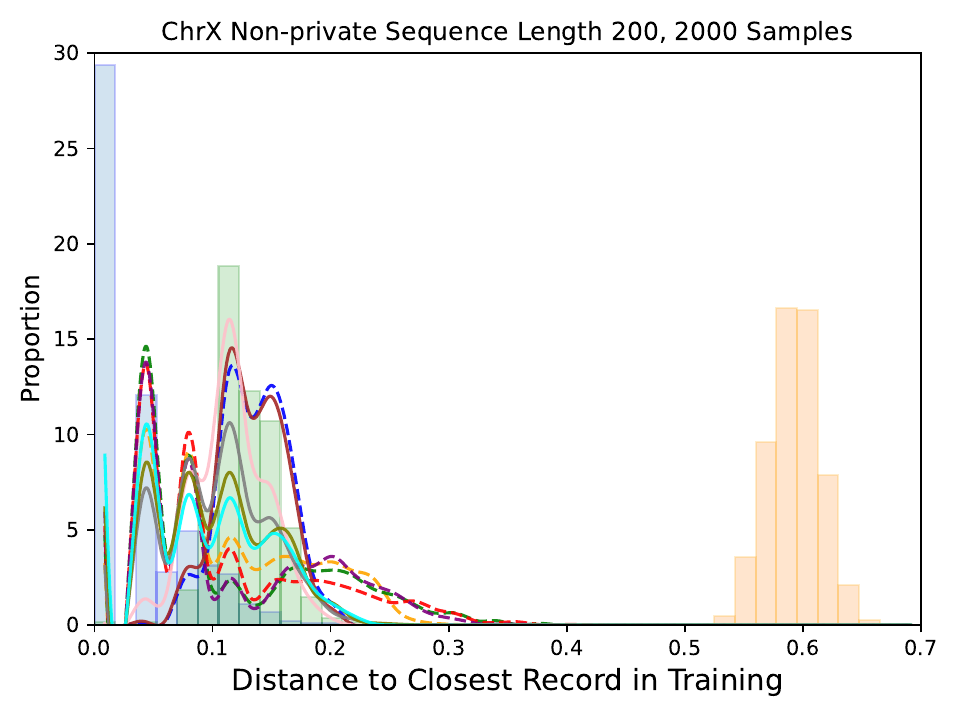}
    \end{minipage}%
    \hspace{-0.30cm}
    \begin{minipage}[b]{0.33\textwidth}
        \centering
        \includegraphics[width=\textwidth]{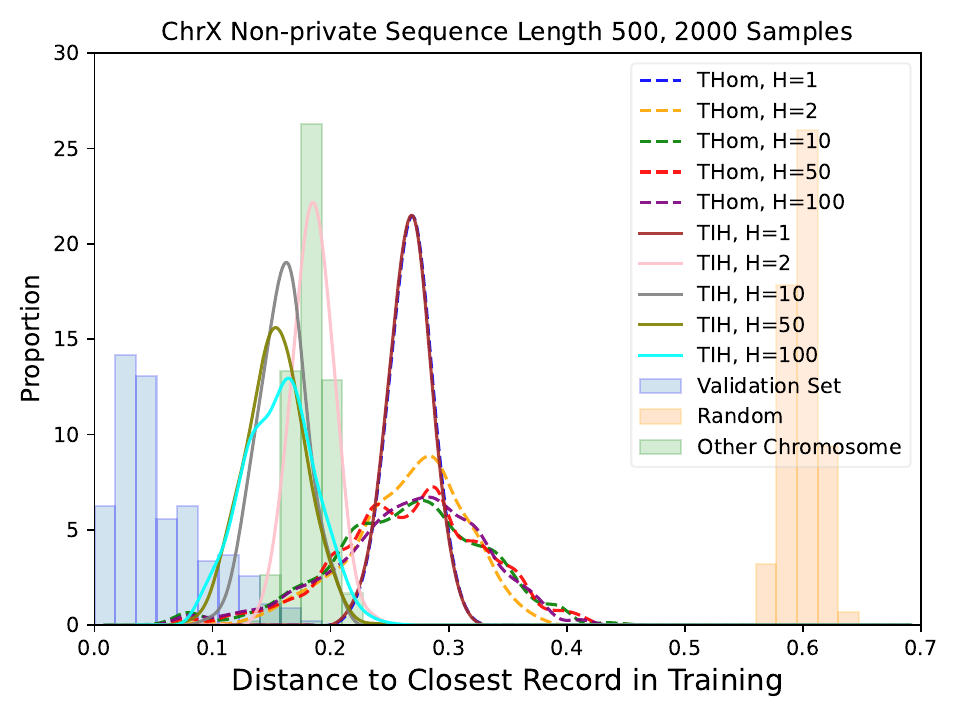}
    \end{minipage}
        
    \caption{Histograms of distances to the closest record in training (chromosome X) for the time-homogeneous (THom) and time-inhomogeneous (TIH) models and different number of hidden states $H$.}\label{fig:histograms-ns2000}
\end{figure*}
\subsection{Non-private Experiments}

We first conduct experiments without applying differential privacy to establish the baseline performance of the HMMs.

\myparagraph{Setup} We conduct our experiments on segments of the first consecutive SNPs from Chromosomes X and 22 with sequence lengths $L \in \{100, 200, 500\}$. The datasets are first shuffled and divided into 5 equal parts. Four parts (2036 points) are used for training, while the remaining part is reserved as a hold-out validation set.

The HMMs are trained over 20 epochs, with 3 observable outcomes $O \in \{0, 1, 2\}$, corresponding to the SNP values. We train the HMMs with varying capacities for the number of hidden states, $H \in \{1, 2, 10, 50, 100\}$. Following a preliminary hyperparameter sweep, we fixed the learning rate at $0.015$, which yielded the best validation performance across most models. We standardized the training epochs and optimization settings across configurations to be able to study the effect of the number of hidden states $H$ better. After training, each model is used to generate synthetic datasets of size $N \in \{100, 500, 1000, 1500, 2000\}$. 

For comparison, we employ two baselines. First, we generate 2000 random sequences of the same length as the original SNP segments, where each SNP value is sampled uniformly at random, i.e., $\Pr(\text{SNP}=0) = \Pr(\text{SNP}=1) = \Pr(\text{SNP}=2) = 1/3$ at each locus. Second, we compute the evaluation metrics for the first consecutive SNPs from another chromosome with the same sequence length as the training dataset. We chose chromosome 21 for this purpose. 

\myparagraph{Distance measures} Figure~\ref{fig:distances_Neis_chrx} presents the performance evaluation of our time-homogeneous (THom) and time-inhomogeneous (TIH) models on chromosome X based on Nei's genetic distance for various numbers of hidden states ($H$) and different numbers of generated synthetic samples. Results for the other two distance metrics and chromosome 22 are provided in Appendix~\ref{sec:app-exp-nodp}.

The first observation is that the performance of the THom model remains constant regardless of the number of sampled points or model capacity ($H$), staying close to the genetic distance observed for the other chromosome across all sequence lengths. In contrast, the TIH model's performance improves with an increasing number of samples and hidden states, achieving very low values (close to $10^{-5}$), indicating a strong resemblance to the training dataset. Note that the range of Nei's genetic distance is $[0, \inf]$. We discuss the interpretation of values for Nei's distance in Appendix~\ref{sec:app-neis}.

For $H=1$, THom and TIH are effectively equivalent and with a single hidden state, both reduce to estimating the average emission distribution over the sequence, leading to indistinguishable performance. For $H=2$, TIH remains too limited to capture the dependencies in the data, an effect that is especially pronounced for the longest sequences ($L=500$). For sequences of length $L=500$, increasing the number of hidden states beyond $H=10$ (e.g., $H\in\{50,100\}$) does not improve any of our distance measures, despite a reduction in validation negative log-likelihood. Under a matched training epoch budget and identical optimization settings, the additional capacity does not translate into better alignment with the long-horizon distributional statistics captured by these metrics; in our setting, $H=10$ is sufficient for $L=500$.

\myparagraph{Histograms of $l_2$ distance to the closest record in training} We present the results for histograms of distances between each synthetic point and its closest neighbor in the training set for chromosome X in Figure~\ref{fig:histograms-ns2000}, considering $N = 2000$ samples. For comparison, we also include histograms of distances to the training set for the hold-out validation set, another chromosome (chromosome 21), and randomly generated points. To enhance clarity, we use cubic splines (degree 3) to connect the midpoints of the histograms for synthetic samples generated by the THom and TIH models, with the number of hidden states denoted as $H$. The histograms for the training chromosome 22 can be found in Appendix~\ref{sec:app-exp-nodp}.

For all sequence lengths, the histograms show that THom models exhibit a longer right tail compared to TIH models, indicating the THom model's difficulty in generating synthetic points similar to the training dataset. This discrepancy becomes more pronounced as the sequence length increases. At length $L = 500$, the peaks of the two models (TIH and THom) become distinctly separated, with the mean distances for samples from the THom model shifting closer to those of random points. 

Additionally, both TIH and THom models exhibit identical behavior for $H = 1$. For TIH with $H = 2$, we observe a heavier right tail, particularly at length $L = 500$, where its peak shifts to the right. However, for higher numbers of hidden states, no significant differences or improvements are observed between the models.

%%%%%%%%%%%%%%%%%%%%%%%%%%%%%%%%%%%%%%%%%%%%%%%%%%%%%%%%%%%%%%%%%%%
%%%%%%%%%%%%%%%%%%%%%%%%%%%%%%%%%%%%%%%%%%%%%%%%%%%%%%%%%%%%%%%%%%%
%%%%%%%%%%%%%%%%%%%%%%%%%%%%%%%%%%%%%%%%%%%%%
\begin{figure*}[ht!]
    \centering
    \hfill
    \begin{minipage}[b]{0.33\textwidth}
        \centering
        \includegraphics[width=\textwidth]{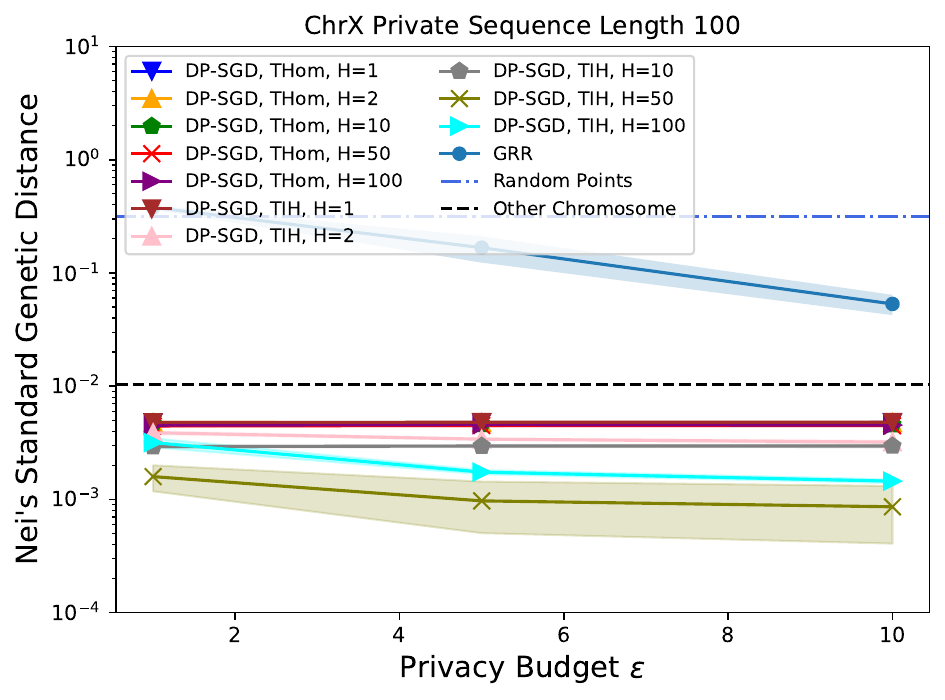}
    \end{minipage}%
    \begin{minipage}[b]{0.33\textwidth}
        \centering
        \includegraphics[width=\textwidth]{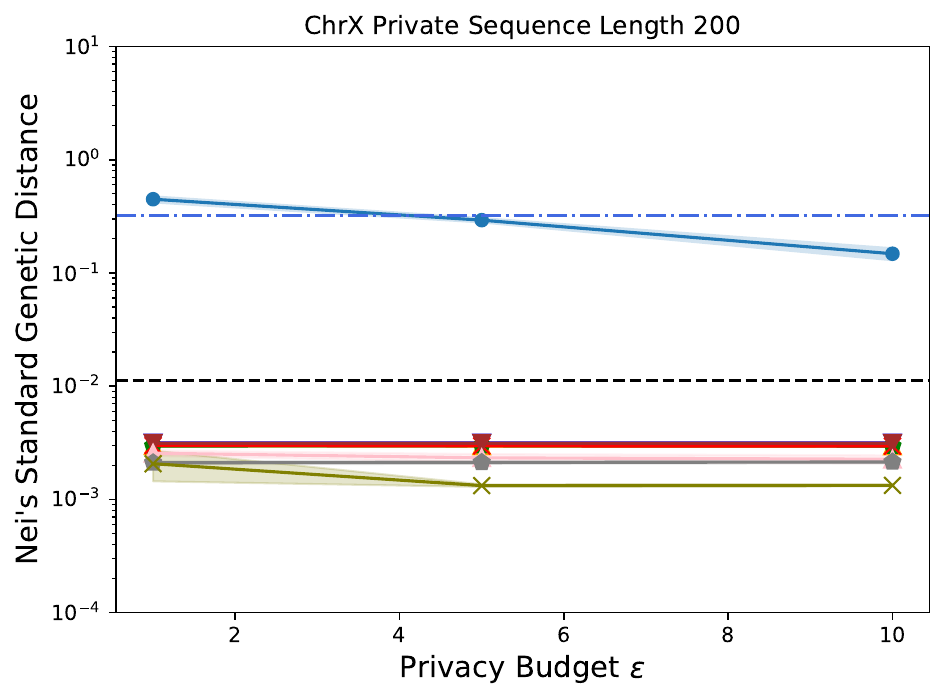}
    \end{minipage}%
    \begin{minipage}[b]{0.33\textwidth}
        \centering
        \includegraphics[width=\textwidth]{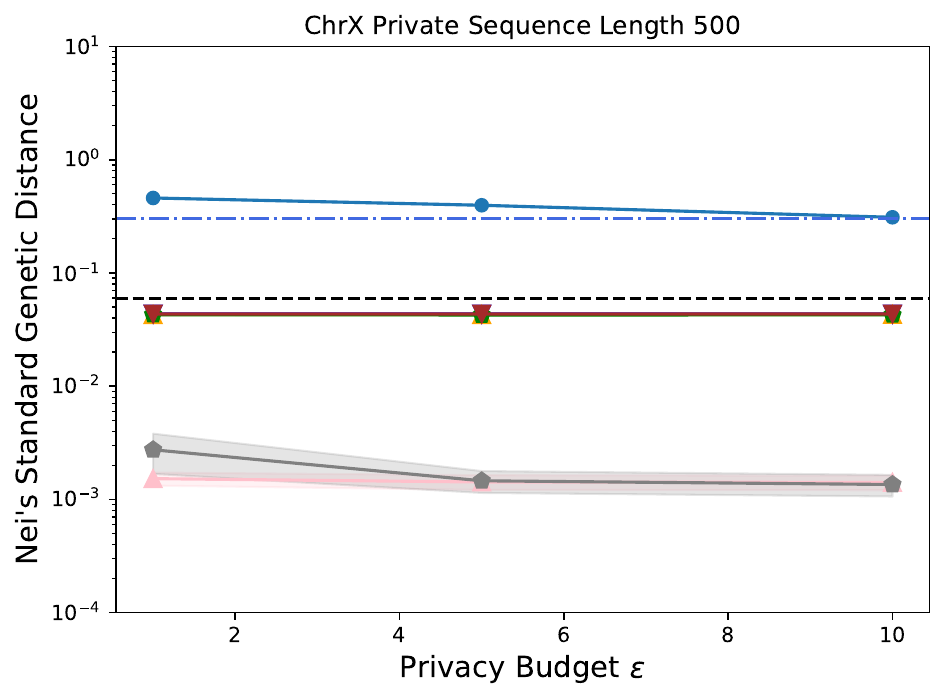}
    \end{minipage}
    
    \caption{Nei's genetic distance between the training chromosome X and synthetic dataset for the time-homogeneous (THom) and time-inhomogeneous (TIH) models with different number of hidden state $H$. The baseline of GRR mechanism is also shown in blue. The shaded areas show the standard deviation over three random runs.}
    \label{fig:distances_Neis_chrx_dplog}
\end{figure*}

\begin{figure*}[ht!]
    \centering
    \includegraphics[width=0.9\textwidth]{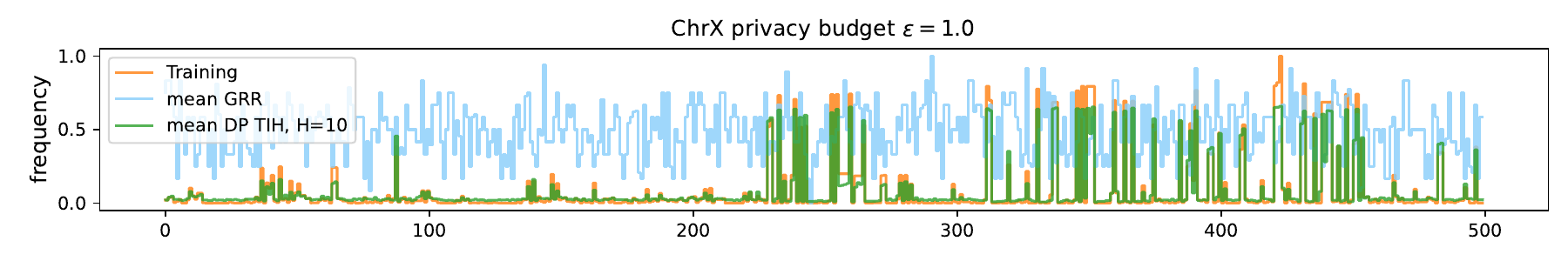}
    \includegraphics[width=0.9\textwidth]{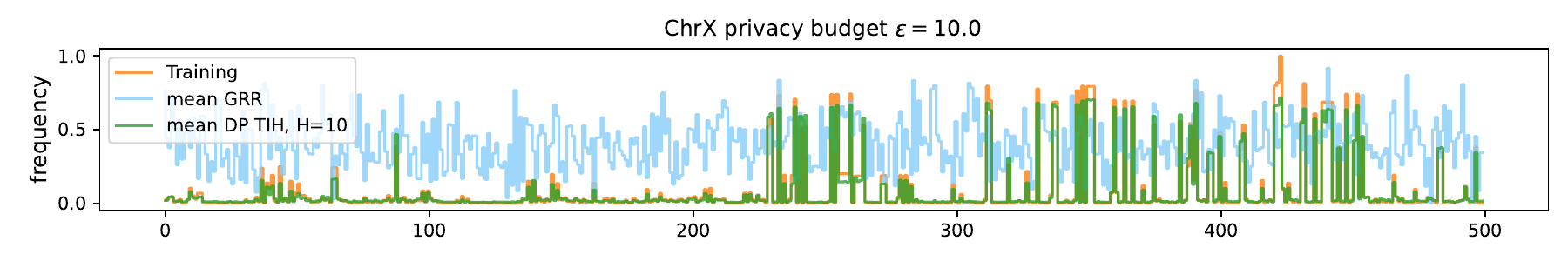}
    \caption{Minor allele frequencies from the real training dataset vs the generated samples from the DP-trained time-inhomogeneous HMM vs the GRR baseline, for SNP sequence length of 500.}
    \label{fig:all-minor-allele-frequencies}
\end{figure*}

\subsection{Differentially-Private HMMs}
We now proceed to the experiments addressing the primary objective of this paper: evaluating whether synthetic datasets sampled from DP-trained models can effectively replicate the statistical properties of the real training dataset.

\myparagraph{Setup} For our DP experiments, we set $\varepsilon \in \{1, 5, 10\}$, spanning a range from strong formal privacy guarantees to more practical privacy levels, consistent with prior work~\cite{ponomareva2023dp}. For the privacy parameter $\delta$, a common guideline is $\delta\in[1/N^{2},\,1/N]$ for $N$ data points~\cite{ponomareva2023dp}. With $N=2000$, we set $\delta=10^{-4}$, satisfying $\delta<1/N$.

Training is performed using a batch size $B = 8$, learning rate $\eta = 0.015$, and $T = 20$ training epochs, matching the configuration of the non-private models. We conducted a phase of preliminary experiments for different values of clipping norms $C \in \{0.1, 1, 5, 10\}$ and selected $C = 1$, as it achieves the highest validation log-likelihood across most model and $H$ settings.

Each experiment is run with three different random seeds, and the mean and standard deviation across runs are reported. This applies to both our DP-SGD method and the generalized randomized response (GRR) baseline. To ensure a fair comparison with the GRR mechanism, we generate 2000 samples from HMMs. Since we define the frequencies to be between 0 and 1, we clip the denoised frequency estimates obtained from the GRR mechanism to lie within this range, ensuring biologically plausible outputs.

\myparagraph{Distance measures} Figure~\ref{fig:distances_Neis_chrx_dplog} presents the results for three different SNP sequence lengths for training chromosome X. The means of Nei's distances are indicated by markers, while the shaded regions represent the standard deviation across three runs. The results for the Euclidean and Manhattan distances as well as the other training chromosome can be found in Appendix~\ref{sec:app-exp-dp}.

Experiments were conducted on a single NVIDIA TITAN RTX GPU with approximately 24 GB of available memory. Due to memory constraints during DP-SGD training in PyTorch, the time-inhomogeneous models with $L = 200, H = 100$ and $L = 500, H \in \{50, 100\}$ exceeded available GPU capacity. Consequently, no results are reported for these configurations. The average training times are also reported in Appendix~\ref{sec:app-exp-dp} and, as expected, the training time increases almost linearly with the sequence length $L$.

For all sequence lengths and $\varepsilon$ values, the GRR mechanism exhibits the lowest utility, performing worse than all DP-trained models and even the other chromosome baselines. As previously observed, the time-homogeneous models do not benefit from a higher number of hidden states or increased $\varepsilon$ (weaker privacy guarantees).

In contrast, the DP-trained TIH models achieve better performance across all lengths, with a clear superiority especially at length $500$. This is especially pronounced in the results for training chromosome 22.

% For $L = 100$, the most complex TIH model ($H = 100$) underperforms relative to lower-capacity variants. This may be because the differential privacy noise has a more detrimental impact on models with a larger number of parameters. 

% Consistent with our observations in the non-private setting, the TIH models achieve lower Nei's distance at sequence length 500. This improvement can be attributed to the richer genomic context, which enables the model to better capture transition dynamics and emission patterns.

\myparagraph{Minor allele frequencies} Figure~\ref{fig:all-minor-allele-frequencies} presents the minor allele frequencies at each SNP locus for the first 500 SNPs of chromosome X. Frequencies are shown for 2000 samples generated by DP-trained TIH models with hidden states $H = 10 $ averaged over three random runs. For GRR, we also plot the debiased frequencies averaged over 3 random runs. The results for $H=2$ as well as chromosome 22 can be found in Appendix~\ref{sec:app-exp-dp}.

The GRR baseline fails to recover meaningful allele frequency patterns, instead producing outputs that resemble random noise. In contrast, the TIH model exhibits a more structured behavior. Under strong privacy constraints (small $\varepsilon$), it tends to reproduce a smoothed, averaged version of the signal, dampening both peaks and low values. As the privacy budget increases, the synthetic allele frequencies generated by the TIH model progressively converge toward those of the real dataset, reflecting a closer alignment with the true distribution.

%%%%%%%%%%%%%%%%%%%%%%%%%%%%%%%%%%%%%%%%%%%%%%%%%%%%%%%%%%%%%%%%%%%%%%%%%%%%%
%%%%%%%%%%%%%%%%%%%%%%%%%%%%%%%%%%%%%%%%%%%%%%%%%%%%%%%%%%%%%%%%%%%%%%%%%%%%
\begin{figure}[ht!]
    \includegraphics[width=0.35\textwidth]{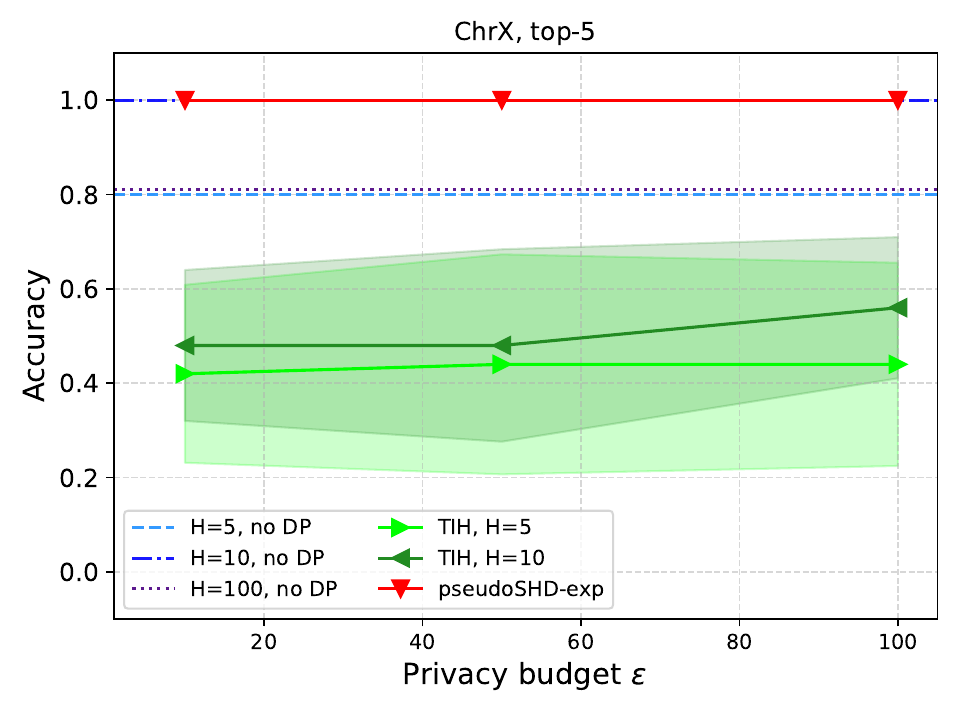}
    \includegraphics[width=0.35\textwidth]{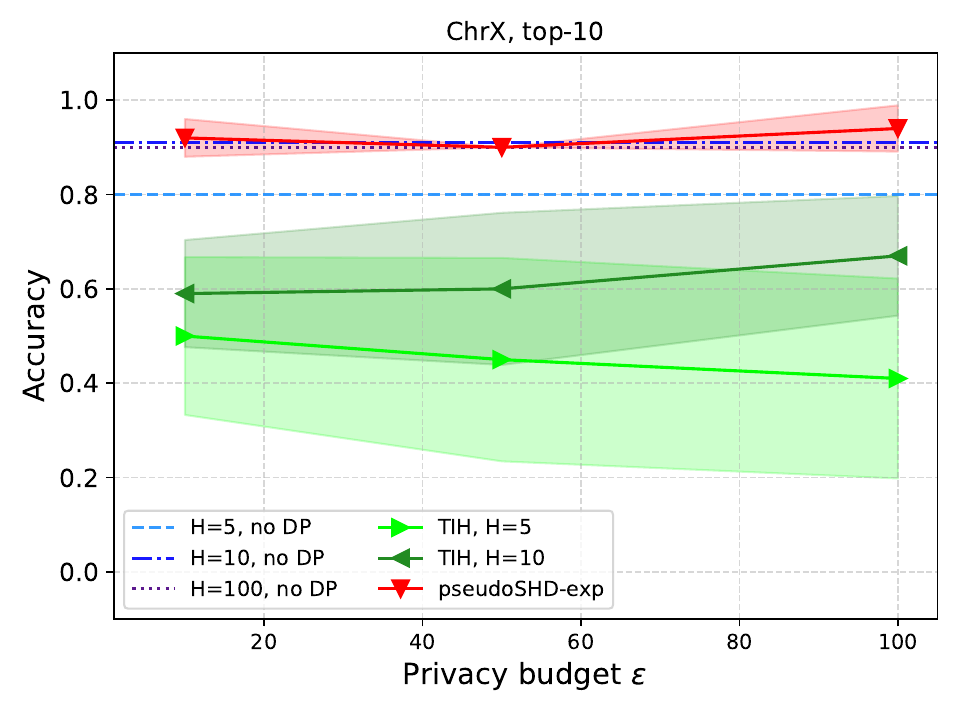}
    \caption{Averaged accuracies of returning the top-$k$ associated SNPs between case and control group. the shaded area shows the standard deviation over random runs of the DP methods}
    \label{fig:chrx-topk}
\end{figure}

\subsection{GWAS Downstream Task}
\label{sec:gwas-downstream}
A central downstream task in GWAS is the identification of SNPs that are statistically associated with a phenotype. To quantify such associations, the $\chi^2$ test of independence is commonly employed. This test evaluates the extent to which the observed genotype (SNP) frequencies differ between case and control groups, relative to the expected frequencies under the null hypothesis of no association. In this work, we focus on the allelic test statistic. 

Consider a biallelic SNP encoded by $\{0, 1, 2\}$, denoting the number of minor alleles carried by an individual. Let $s_0$, $s_1$, and $s_2$ be the counts of individuals in the control group (of size $S$) with genotypes $0$, $1$, and $2$, respectively. Analogously, let $r_0$, $r_1$, and $r_2$ denote the corresponding counts in the case group (of size $R$). Denote by $n_0$, $n_1$, and $n_2$ the total number of individuals (cases and controls combined) with genotypes $0$, $1$, and $2$, respectively. These genotype counts can be mapped to the number of minor alleles in cases and controls, as summarized in this table:

{\small
\begin{table}[h]
\centering
% \caption{Allelic contingency table per SNP}
\label{tab:allele-contingency}
\begin{tabular}{|c|c|c|c|}
\hline
\textbf{Allele} & \textbf{Cases} & \textbf{Controls} & \textbf{Row total}\\
\hline
Minor & $r_1+2r_2$ & $s_1+2s_2$& $n_1+2n_2$\\
Major & $2r_0+r_1$ & $2s_0+s_1$ & $2n_0+n_1$\\
\hline
\textbf{Column total} & $2R$ & $2S$ &$2N$\\
\hline
\end{tabular}
\end{table}
}

The allelic test statistic for this contingency table is given by: 
\begin{equation}
    \chi^2 = \frac{2N[(2r_0+r_1)S - (2s_0+s_1)R]^2}{RS(2n_0+n_1)(n_1+2n_2)}\nonumber
    \label{eq:ca-test-stat}
\end{equation}

For each SNP in the dataset, this statistic is computed, and the SNPs exhibiting the strongest associations with the phenotype are subsequently selected.

\myparagraph{Phenotype simulation} Due to privacy concerns, access to labeled or phenotyped genomic datasets is severely restricted. Even many publicly available resources that previously included phenotypic annotations have been removed from circulation; a notable example is the OpenSNP~\cite{OpenSNPClosure2023} project. In the absence of phenotype information, we simulate case-control labels using the 1000 Genomes dataset.

To construct these synthetic phenotypes, we first randomly select a SNP locus  $i$ such that its minor allele frequency $f^m_i$ satisfies  $0.1 < f^m_i < 0.9$. This threshold ensures that the locus exhibits sufficient variability across individuals, avoiding cases where the SNP is nearly monomorphic (i.e. all individuals have the same allele at that specific locus). Individuals with genotype 2 at this locus are assigned to the case group, while the remainder are designated as controls. To balance the class sizes, we apply a post-processing step: if the case group is overrepresented, a subset of individuals from the control group is randomly selected and reassigned as cases, yielding an approximately balanced case-control split, and vice versa. 

\myparagraph{Setup} We conduct our experiments using sequence length of $L = 500$ and employ time-inhomogeneous HMMs. For each of the case and control groups, we randomly shuffle the data and partition it into five subsets, using four for training and one as a hold-out validation set. Two separate TIH-HMMs are trained: one exclusively on the case training set, and the other on the control training set. \textit{\textbf{This setup reflects a typical potential use case of our proposed pipeline, in which a data holder, such as a clinical institution, may train an HMM on the SNP sequences of a specific cohort (e.g., individuals with a particular disease) and release the model to enable exploratory analyses by external researchers.}}

Given that the training data for each model is limited to approximately 1000 individuals, a reduction in performance is anticipated. So we allow for a higher privacy budget and evaluate our approach using $\varepsilon \in \{10, 50, 100\}$ and 10 random seeds. Other hyperparameters of the DP-SGD training are kept the same as in the previous section.

\myparagraph{Evaluation}  
For each experiment, we generate 2000 samples from the TIH-HMM trained on the case group and another 2000 samples from the model trained on the control group. We then perform a $\chi^2$ test between these two synthetic datasets and identify the top-$k$ associated SNPs based on their $p$-values. To evaluate the fidelity of the synthetic data in recovering meaningful genetic signals, we define the accuracy as:
$$
\text{Acc}(k) = \frac{|\{\text{SNP}_k^*\} \cap \{\text{SNP}'_k\}|}{k}
$$
where $\{\text{SNP}_k^*\}$ denotes the set of top-$k$ SNPs identified using the real case/control datasets, and $\{\text{SNP}'_k\}$ represents the corresponding top-$k$ SNPs obtained from the synthetic sequences generated by the trained models.  

\myparagraph{SOTA Baseline} 
To assess the performance of our model, we compare against a state-of-the-art DP method specifically designed to return the top-$k$ most strongly associated SNPs in GWAS. This approach employs the exponential mechanism to select SNPs based on the \textit{shortest Hamming distance (SHD)} score~\cite{johnson2013privacy}. In essence, the SHD measures the minimum number of modifications to the dataset required to flip a SNP from significant to non-significant or vice versa.

Computing the exact SHD scores, however, is computationally expensive. For this reason, we adopt the approximate and highly efficient variant proposed by~\cite{yamamoto2023joint}, which we refer to as \textit{pseudoSHD-exp}. Two important aspects of these methods should be noted: firstly, bounded DP is used in the definition of pseudoSHD-exp. Secondly, the privacy budget is allocated exclusively to the $k$ SNPs selected by the exponential mechanism. This stands in contrast to our method, which ensures that the entire signal is privatized.

\myparagraph{Results} Figure~\ref{fig:chrx-topk} reports the top-5 and top-10 accuracies on chromosome~X, where non-private TIH baselines with $H \in \{10, 50, 100\}$ are also included for comparison. The shaded regions denote the standard deviation across random runs of the DP mechanism. Results for top-1, top-3, and top-$k$ accuracies on chromosome~22 are provided in Appendix~\ref{sec:app-downstream}.  

For chromosome~X, the non-private baselines achieve consistently strong performance. Notably, the TIH model with $H=10$ outperforms or matches the more complex variant with $H=100$ across all settings. This may be due to the limited training budget of 20 rounds, which constrains the larger model's optimization, or because the broader representations learned with $H=100$ are less aligned with the specific task of SNP association under this data regime.

The DP-trained models also demonstrate clear improvements over random chance (expected accuracies of $0.1$ and $0.2$ for top-5 and top-10, respectively). Among these, TIH with $H=10$ surpasses the smaller $H=5$ model, which generally struggles to improve even under higher privacy budgets. This indicates that the $H=5$ configuration lacks sufficient capacity to capture the signal at the level of precision required to generate reliable top-$k$ SNPs.  

Interestingly, for TIH with $H=5$ accuracy does not increase monotonically with the privacy budget. We attribute this to several interacting factors. First, DP-SGD noise provides implicit regularization; at larger $\varepsilon$ the reduced noise can lead to overfitting of the smaller model to cohort-specific artifacts, degrading downstream GWAS ranking despite improved allele-frequency fit. Second, our training objective (matching MAF/sequence statistics) is only a proxy for association recovery; improvements in the proxy need not translate to better top-$k$ SNP identification. Third, the interaction between gradient clipping and the optimizer is nonlinear in the noise scale, so fixed hyperparameters (learning rate, clipping threshold, epochs) are not jointly optimal across $\varepsilon$, and for our experiments we use the optimal parameters for lower $\varepsilon$ regime of $[1, 10]$. Finally, top-$k$ accuracy can fluctuate when multiple SNPs have nearly identical test statistics, since small sampling differences in the synthetic cohorts may change their order. This sensitivity to near-ties and sampling variability can lead to non-monotonic trends across privacy budgets. This last point is extensively discussed in Appendix~\ref{sec:app-downstream}

Nevertheless, SOTA pseudoSHD method consistently outperforms our DP-trained models. The performance gap is particularly evident for the top-1 SNP, as well as in scenarios where the $p$-values (or equivalently, test statistics) of the top-$k$ and top-$(k+1)$ SNPs are nearly indistinguishable. As discussed extensively in Appendix~\ref{sec:app-downstream}, this outcome is expected: the probability of pseudoSHD selecting the top-$k$ SNP is directly proportional to the original test statistic, whereas our locus-dependent HMM is designed to model the global signal distribution. Consequently, the performance of our method deteriorates when consecutive associated SNPs differ only marginally in their test statistics.  

We therefore consider SOTA approaches to be complementary rather than competing baselines, as they address fundamentally different problem formulations. Exponential mechanism optimizes the recovery of specific top-ranked SNPs, while our DP-trained HMM targets the reconstruction of broader association patterns.  

%%%%%%%%%%%%%%%%%%%%%%%%%%%%%%%%%%%%%%%%%%%%%%%%
%%%%%%%%%%%%%%%%%%%%%%%%%%%%%%%%%%%%%%%%%%%%%%%%%

\subsection{Pairwise Correlation of SNPs}  
A widely used approach for analyzing correlation structures among SNPs is the computation of pairwise linkage disequilibrium (LD) using the $r^2$ statistic~\cite{rogers2009linkage}. The $r^2$ value ranges from 0 (no LD) to 1 (perfect correlation), thus providing a quantitative measure of the strength of association between SNP pairs. Characterizing LD patterns plays a crucial role as a preprocessing step in genome-wide association studies, particularly for tasks such as SNP imputation. Since SNP datasets often contain missing values, these are typically inferred (imputed) using HMMs trained on a complete reference panel~\cite{li2003modeling}. The HMM leverages SNP-SNP correlations to perform this imputation. It is important to note, however, that in this setting, imputation is usually carried out on allele sequences (two per individual), whereas in our models we instead operate on alternate count representations of SNPs (genotypes).

\myparagraph{Performance measures} 
To evaluate how well our models preserve LD patterns, we introduce two primary metrics: the \emph{Best-Tag Shift Score} (BTSS) and the \emph{Exact Match Rate}. 

Consider an $L \times L$ matrix of pairwise LD correlations $r_{ij}^2$ for a sequence of length $L$. For each SNP $i$, we denote the strongest tag SNP as
    $R_{i}^* = \max_j r_{ij}^2, 
    \quad J_i^* = \arg\max_j r_{ij}^2$ ,
where $R_i^*$ is the maximum correlation and $J_i^*$ is the corresponding SNP index. Analogously, let $\hat{R}_i^*$ and $\hat{J}_i^*$ denote the same quantities obtained from a synthetic or alternative dataset. 

The per-SNP BTSS is then defined as
\begin{align}
\small{%
    \text{BTSS}_i = \exp\!\left(-\frac{|J_i^* - \hat{J}_i^*|}{\lambda}\right) 
    \bigl(1 - |R_i^* - \hat{R}_i^*|\bigr),\nonumber
}
\end{align}
where $\lambda > 0$ is a decay parameter that controls the tolerance for positional shifts, which we set to $2$. A perfect match of tag SNP position and strength yields $\text{BTSS}_i = 1$, while large discrepancies in either position or $r^2$ drive the score toward $0$. The overall BTSS is obtained by averaging $\text{BTSS}_i$ across all SNPs.

As a complementary measure, we define the $\text{Exact Match Rate}= \frac{1}{L} \sum_{i=1}^{L} \mathbf{1}\!\bigl(J_i^* = \hat{J}_i^*\bigr)$, which quantifies the fraction of SNPs for which the tag SNPs coincide exactly.

% we define the \emph{Exact Match Rate}, which quantifies the fraction of SNPs for which the tag SNPs coincide exactly:
% \begin{align}
%     \text{Exact Match Rate} = \frac{1}{L} \sum_{i=1}^{L} \mathbf{1}\!\bigl(J_i^* = \hat{J}_i^*\bigr).\nonumber
% \end{align}

\begin{figure*}[ht!]
    \centering
    \includegraphics[width=0.29\textwidth]{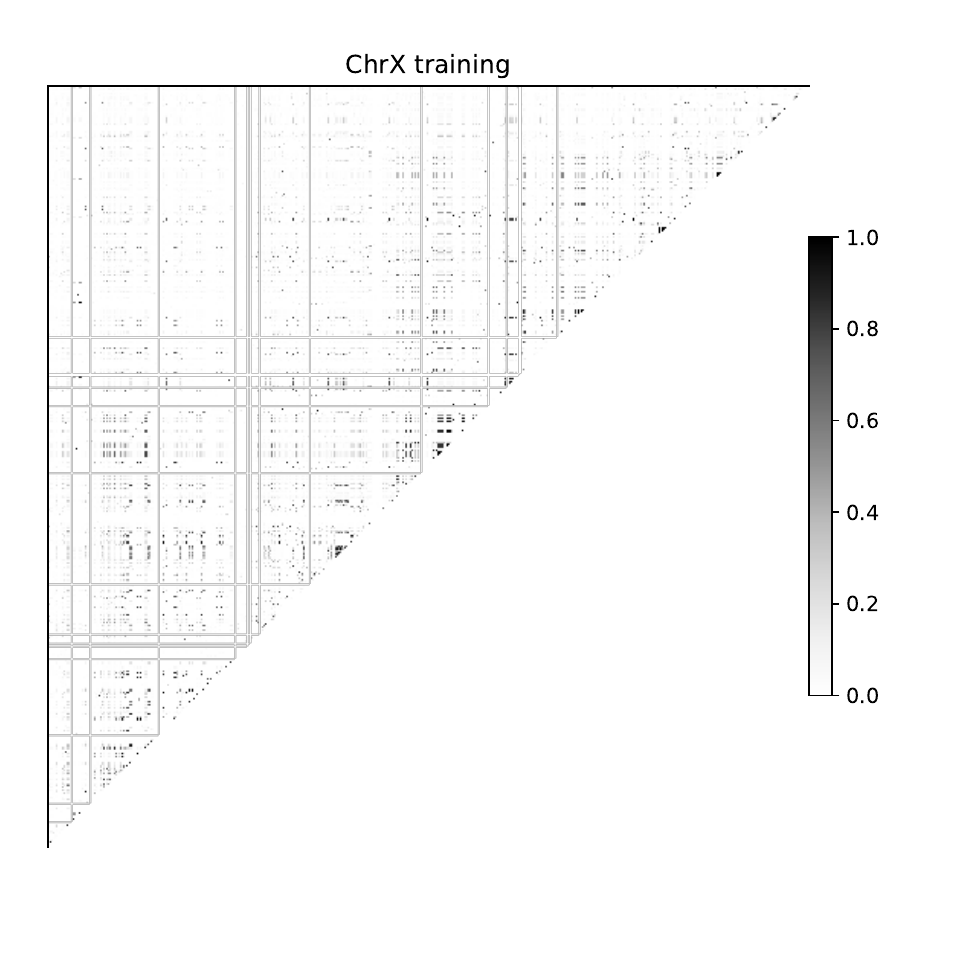}
    \includegraphics[width=0.29\textwidth]{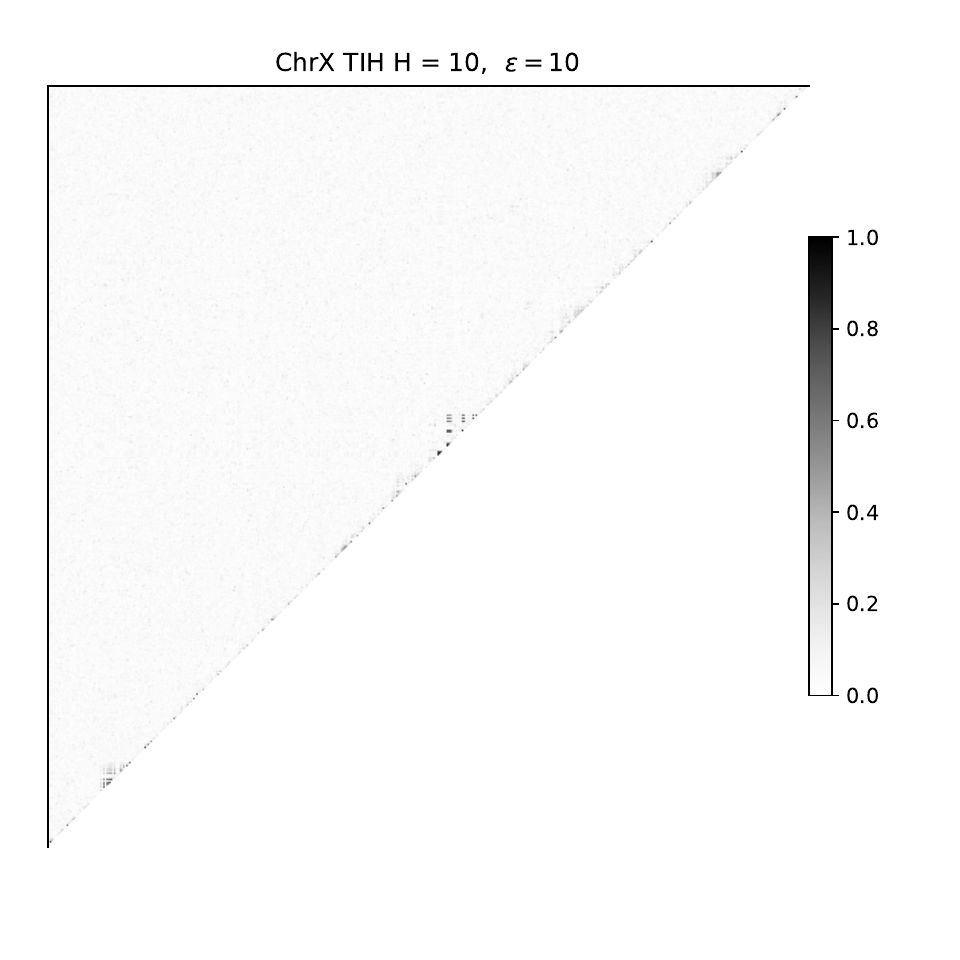}
    \includegraphics[width=0.29\textwidth]{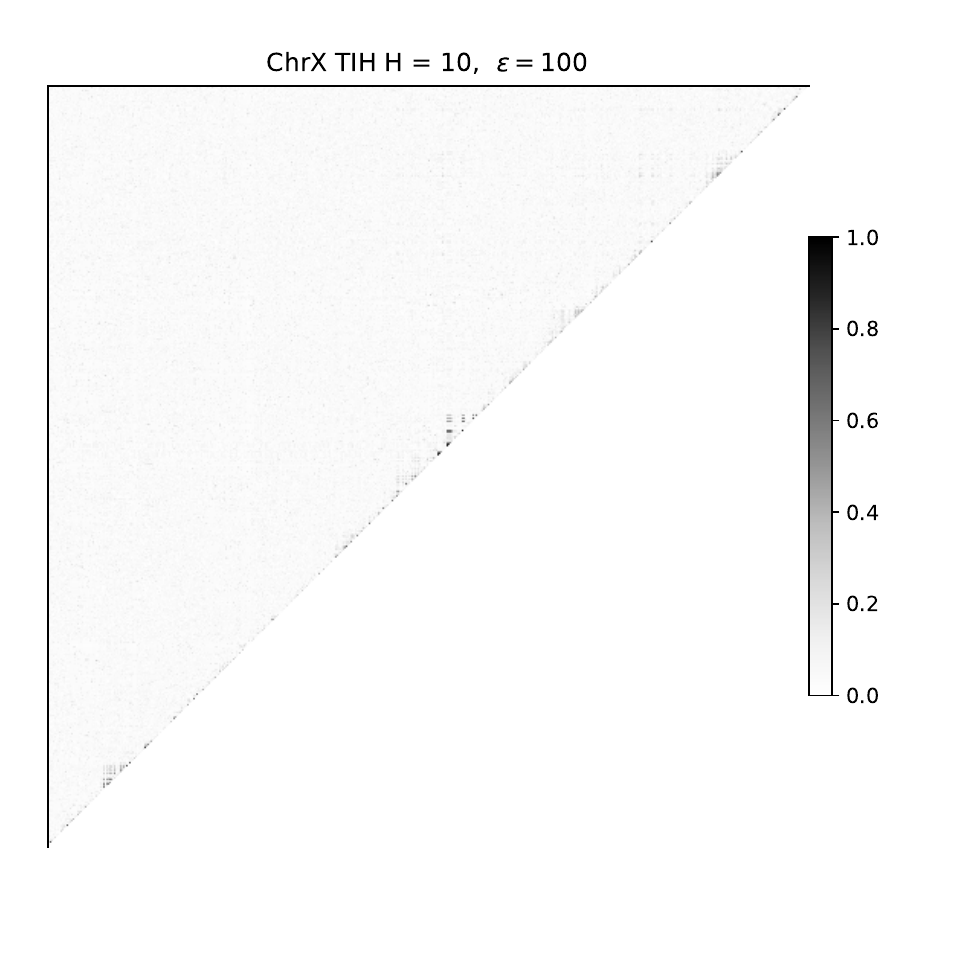}
    \includegraphics[width=0.29\textwidth]{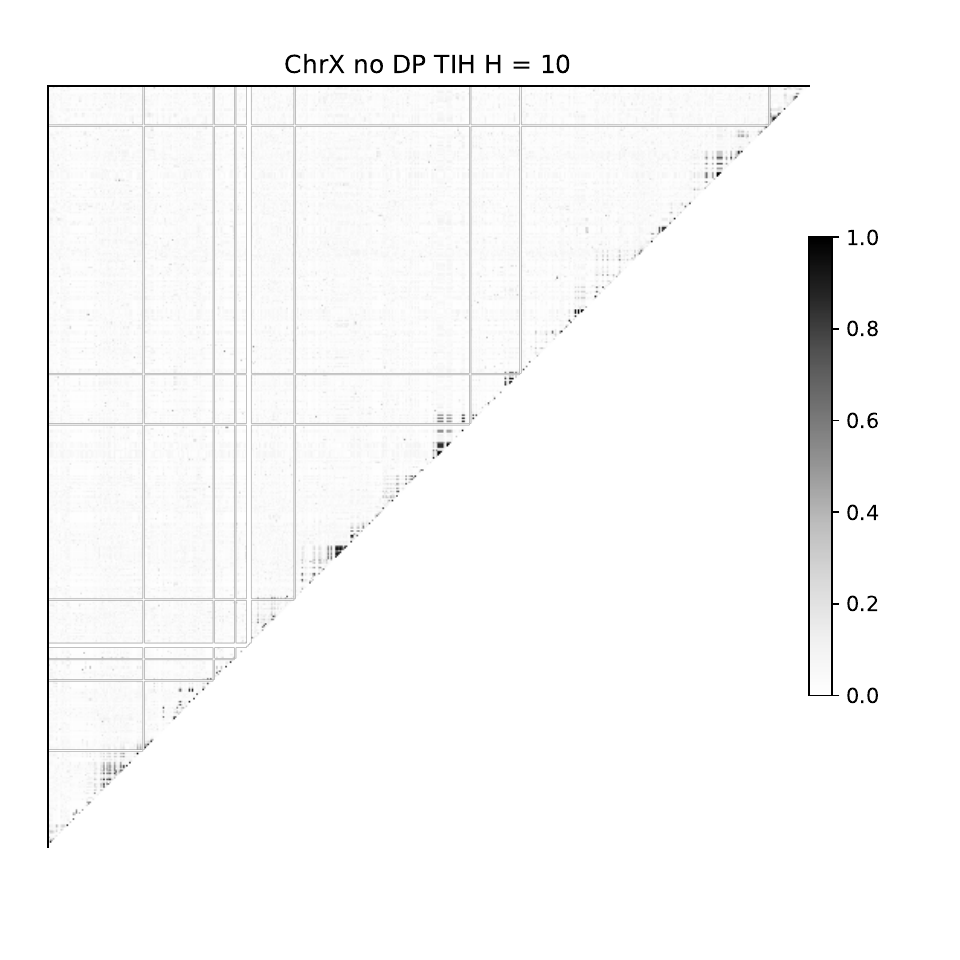}
    \includegraphics[width=0.29\textwidth]{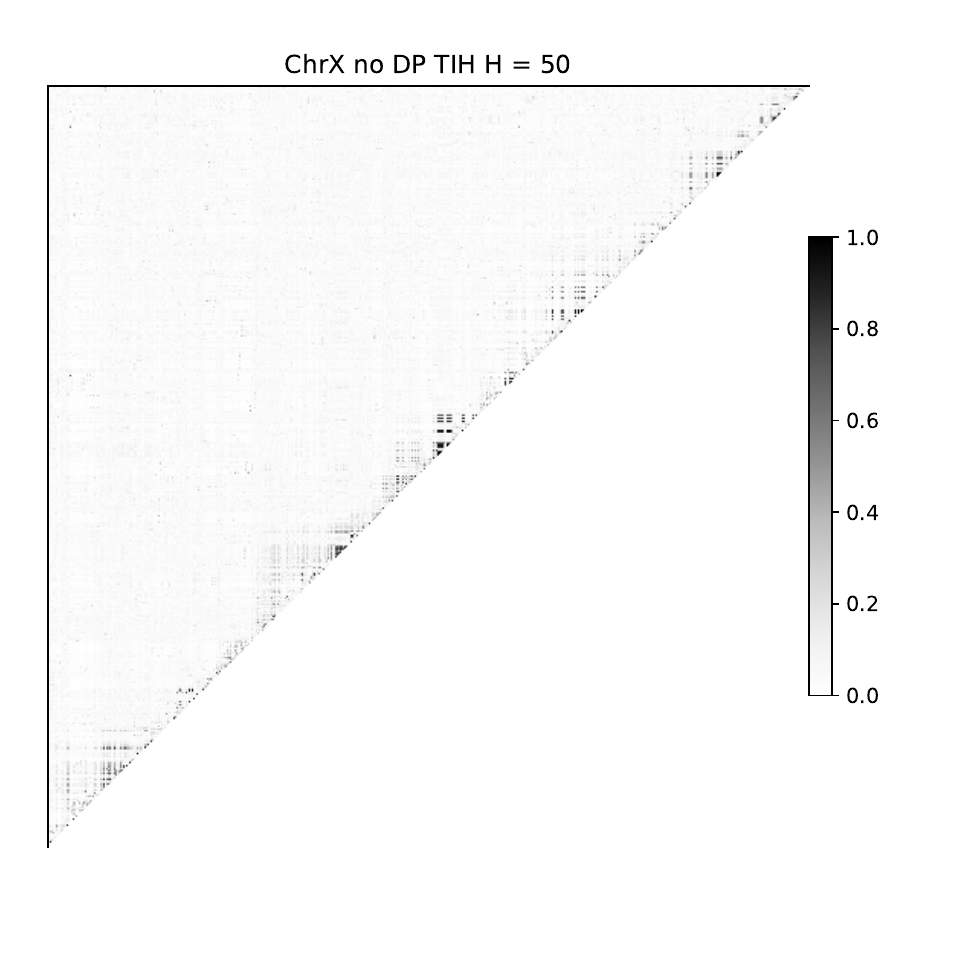}
    \includegraphics[width=0.29\textwidth]{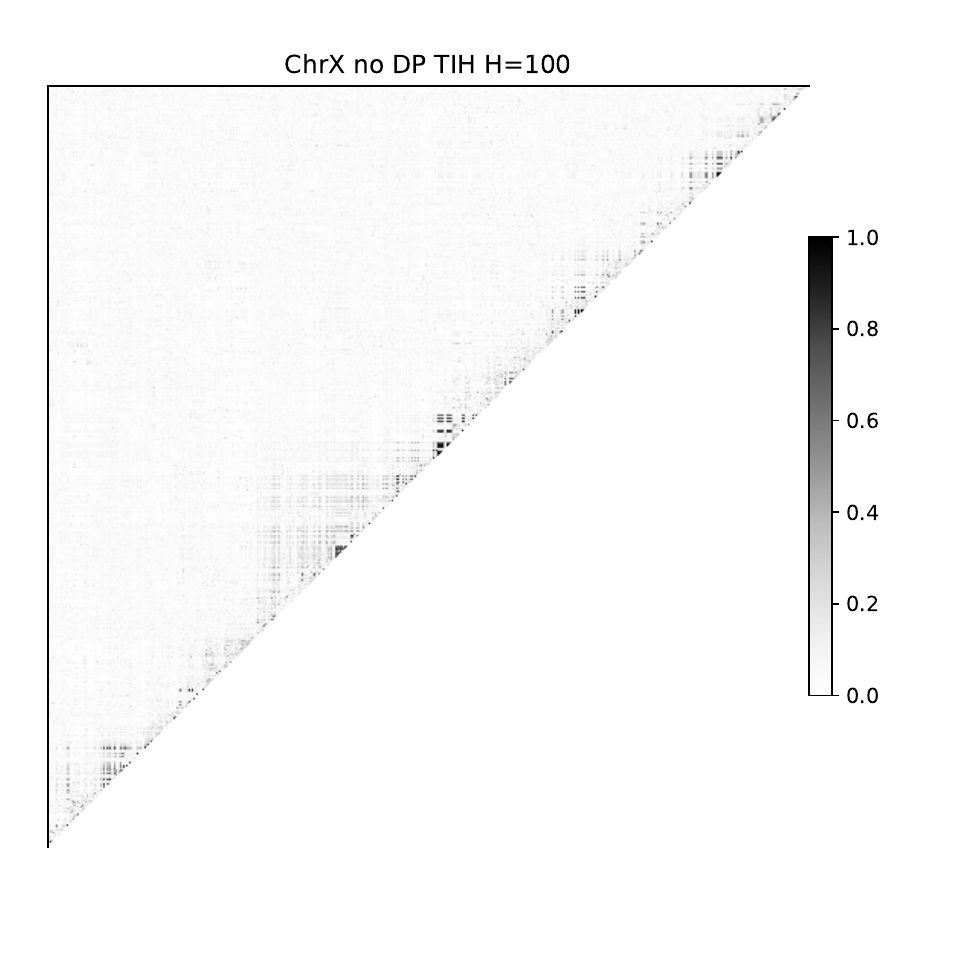}
    \includegraphics[width=0.29\textwidth]{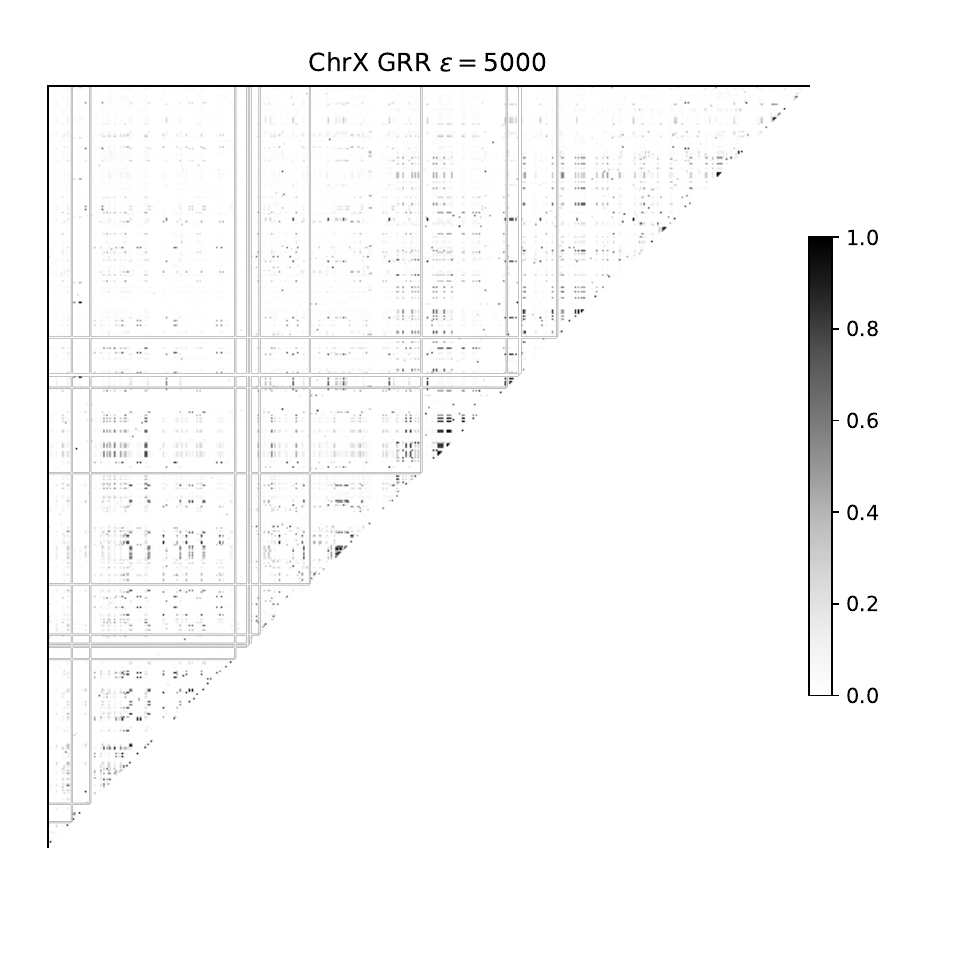}
    \includegraphics[width=0.29\textwidth]{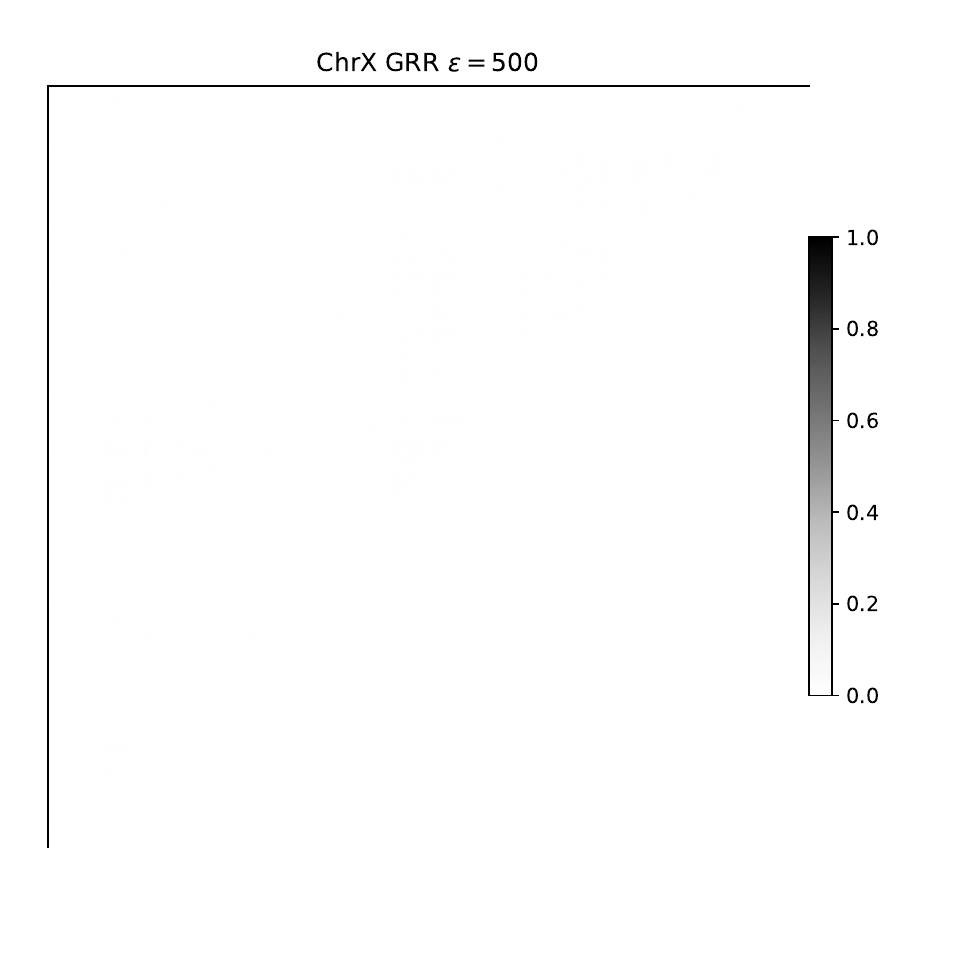}
    \includegraphics[width=0.29\textwidth]{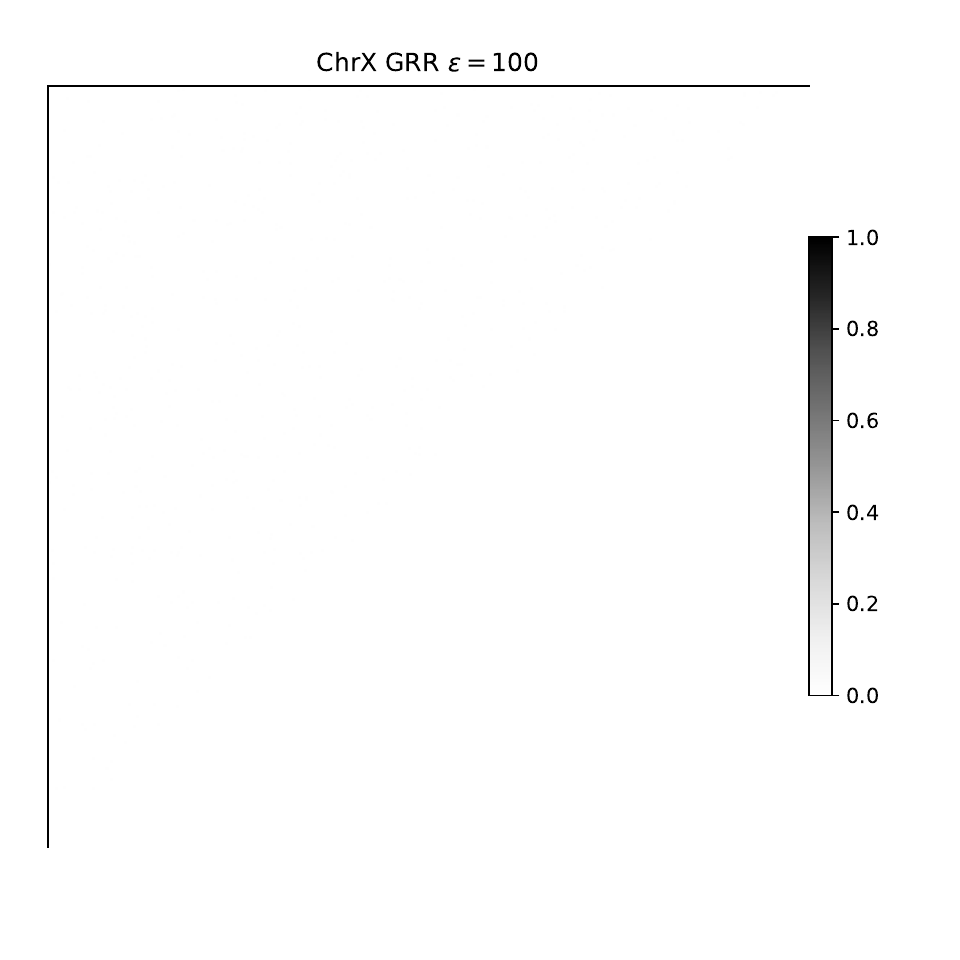}
    \caption{Pairwise LD correlations of the first 500 SNPs for chromosome X.}
    \label{fig:main-chrX-ld-plots}
\end{figure*}

\myparagraph{Results} Table~\ref{tab:main-chrx-btss-exact} presents the results for chromosome~X. Reported values correspond to the mean and standard deviation of the DP mechanism, averaged over three independent random runs. As a baseline, we again include results obtained using GRR. The corresponding LD panels are also shown in Figure~\ref{fig:main-chrX-ld-plots}. For the DP mechanisms, we plot the results from a single random seed. To improve the visual clarity of the correlation heatmaps, we scale each cell by $(r^2)^{0.4}$ for TIH model results. Results for chromosome~22 are provided in Appendix~\ref{sec:app-ld-correlations}. 

A key observation is that for the BTSS and Exact Match metrics, the non-private TIH model substantially outperforms GRR, even at a high privacy budget of $\varepsilon = 500$. This trend persists for the DP-trained TIH models, with GRR only surpassing TIH performance at a significantly larger privacy budget. 

Interestingly, larger TIH models ($H \in \{50, 100\}$) underperform compared to the smaller TIH model ($H=10$). Examination of the LD panels reveals that while the larger models are capable of recovering longer-range correlations, this comes at the expense of accurately capturing sharp local peaks. We hypothesize that extending the training of the non-private TIH models for additional epochs could improve their performance. Alternatively, incorporating higher-order HMM structures (i.e., allowing transitions not only to adjacent states but also to more distant ones) may further enhance the performance of the smaller TIH model with $H=10$. 

Overall, these findings highlight that non-private TIH models are still able to capture key correlation patterns, despite being trained on a dataset of a different type.

\begin{table}[h!]
\centering
\caption{Comparison of BTSS and Exact Match for different mechanisms for training chromosome X.}
\small{%
\begin{tabular}{lcc}
\hline
 & BTSS & Exact Match \\
\hline
GRR $\varepsilon=100$ &  $0.20\pm 0.01$&  $0.13 \pm 0.01$\\
GRR $\varepsilon=500$ &  $0.23\pm 0.01$&  $0.25 \pm 0.01$\\
GRR $\varepsilon=5000$ &  $0.97\pm 0.01$&  $0.96 \pm 0.01$\\
TIH $H=10,$ no DP & $0.67\pm 0.00$ & $0.61 \pm 0.00$ \\
TIH $H=50,$ no DP & $0.54 \pm 0.00$ & $0.46 \pm 0.00$ \\
TIH $H=100,$ no DP & $0.50 \pm 0.00$ & $0.45 \pm 0.00$ \\
TIH $H=10, \varepsilon=10$ & $0.34 \pm 0.01$ & $0.31\pm 0.01$ \\
TIH $H=10, \varepsilon=100$ & $0.35\pm 0.01$ & $0.31 \pm 0.02$ \\
\hline
\end{tabular}
}
\label{tab:main-chrx-btss-exact}
\end{table}

\scalebox{0.9}{
\hspace{-0.5cm}\begin{tcolorbox}[width=1.1\linewidth, sharp corners=all, colback=white!95!black]
Final takeaways:
\begin{enumerate}
\item For the given dataset size, sequence length, and training budget, $H=10$ consistently achieves the best overall performance across tasks. In contrast, $H=100$ underperforms at this budget, yet captures additional information such as long-range correlations. This suggests that $H=100$ may benefit from more epochs, though at the cost of a higher privacy budget due to repeated DP-SGD steps.  
\item Our TIHMM approach yields robust results across multiple metrics and downstream tasks. While not surpassing task-specific state-of-the-art models, our models trained solely on genotyping values, generalizes effectively and delivers competitive performance without task-specific optimization.  
\end{enumerate}
\end{tcolorbox}
}

%% file: body/related-work.tex
\section{Related Work}
\label{sec:related-work}
The vast number of SNPs, reaching over 107,000 on chromosome X alone in the 1000 Genomes Project, and their complex correlations induced by linkage disequilibrium present significant challenges in designing differentially private algorithms for genomic datasets. Prior work on privacy-preserving genome-wide association studies can be broadly categorized into two main research directions:

\textbf{1) DP-protected release of GWAS statistics.}  
Early approaches primarily focused on releasing summary statistics such as $p$-values of the top-$k$ most associated SNPs. These methods typically release only a few SNPs (e.g., 2–5), striking a balance between utility and privacy while avoiding the challenges posed by long, correlated SNP sequences. Fienberg et al.~\cite{fienberg2011privacy} and Uhler et al.~\cite{uhlerop2013privacy} introduced differentially private mechanisms for releasing averaged minor allele frequencies, $\chi^2$ statistics, and SNP $p$-values. Johnson and Shmatikov~\cite{johnson2013privacy} applied the exponential mechanism to protect a variety of GWAS-derived statistics, including the number and location of significant SNPs, correlation blocks, and pairwise correlations. Tramer et al.~\cite{tramer2015differential} proposed relaxations of differential privacy to improve the utility of $\chi^2$ statistics under varying adversarial assumptions.

\textbf{2) Privacy-aware SNP (subset) release using auxiliary information.}  
A complementary line of research focuses on releasing selected subsets of SNPs deemed safe, either by relaxing DP definitions or employing alternative privacy notions. These methods often utilize auxiliary information, such as public SNP correlation structures, to guide SNP selection. Humbert et al.~\cite{humbert2014reconciling} formulated the SNP release problem as a non-linear optimization task, selecting SNP subsets that maximize utility under privacy constraints; they release up to 50 SNPs in their experiments. Yilmaz et al.~\cite{yilmaz2020preserving} proposed the concept of $\varepsilon$-indirect differential privacy, where sharing decisions are based on an attacker’s auxiliary knowledge, rather than on noise addition. In their experiments, approximately 100 SNPs per individual are released. Deznabi et al.~\cite{deznabi2017inference} extended belief propagation attacks~\cite{humbert2013addressing} by incorporating SNP correlations, kinship, and phenotype data. As a defense, they proposed a belief-limiting mechanism that defines privacy in terms of bounding the adversary’s belief update; this approach enables the release of up to 900 SNPs from a dataset of 1000 SNPs. 

Yilmaz et al.~\cite{yilmaz2022genomic} introduced $T$-dependent Local Differential Privacy (LDP), which relaxes traditional LDP by requiring indistinguishability only among SNP values that are statistically plausible, i.e., those with sufficiently high posterior probability given previously released SNPs. By eliminating implausible genotypes and redistributing the probability mass accordingly, their method enhances utility while ensuring privacy, allowing for the full release of 1000 SNPs. Jiang et al.~\cite{jiang2022reproducibility} proposed a two-stage framework in which SNPs are first binarized and then perturbed via a Bernoulli XOR mechanism~\cite{ji2021differentially}. A post-processing step uses optimal transport to adjust the perturbed dataset according to publicly available minor allele frequencies, enabling the release of up to approximately 28,000 SNPs.

\textbf{Our work.}  
Distinct from prior work, our method does not aim to release DP statistics or select a privacy-compliant subset of SNPs. Instead, we focus on generating a synthetic dataset that can support exploratory genomic studies. Our approach operates independently of any auxiliary datasets or public SNP correlation information. It neither obfuscates nor selectively omits SNPs; rather, it releases full sequences of SNPs in a chosen genomic region. We demonstrate that, using only a single GPU with 24\,GB of memory, our method can release synthetic genomic sequences spanning up to 500 consecutive SNPs.

%%%%%%%%%%%%%%%%%%%%%%%%%%%%%%%%%%%%%%%%%%%%%%%%%%%%%%%%%%%%%%%%%%%%%%%%%%%%%%%%%%%

%% file: body/appendix.tex
\section{On the Value of Nei's Standard Distance}
\label{sec:app-neis}

Although studies directly reporting Nei's genetic distance on genome-wide SNP datasets are scarce, there are closely related works on alternative marker types that provide useful numerical baselines. Hu et al.~\cite{hu2017high} computed Nei's standard genetic distance between populations from the 1000 Genomes Project using copy number variation (CNV) loci across the whole genome. They reported values as low as $0.001$ between very closely related East Asian populations (CHB--CHD), while Yoruba versus Han Chinese comparisons reached up to $0.0241$, with mean values of $0.0029$ within Africa, $0.0085$ within non-Africans, and $0.0174$ between African and non-African populations. Similarly, Zhao et al.~\cite{zhao2022genetic} analyzed insertion--deletion (InDel) polymorphisms across autosomes in the same reference panels, calculating Nei’s genetic distance from genome-wide panels. They observed values in the range $0.0009$--$0.0033$ between Han Chinese and other East Asian populations, and as high as $0.0269$--$0.0555$ with African populations. While these measures are not derived from SNP datasets, they nevertheless provide a frame of reference: distances on the order of $10^{-3}$ characterize very close populations, whereas values above $10^{-2}$ reflect continental-scale divergence.

\section{Baseline}
\label{sec:app-baseline}
As a baseline, we select a local differential privacy (LDP) approach, as it provides the most comparable differential privacy framework to our proposed pipeline and is commonly used as a baseline in DP research for GWAS datasets~\citep[e.g.,][]{jiang2022reproducibility, yilmaz2022genomic}. Our method generates a synthetic dataset that has the original SNP sequence length, aligning with the output of an LDP mechanism. Specifically, in an LDP framework, each feature of every record is perturbed to introduce uncertainty, thereby ensuring a quantifiable degree of deniability for individual contributions. 

Here, we provide a brief overview of LDP and describe the specific mechanism used in our paper: generalized randomized response (GRR).

Local Differential Privacy is a privacy framework where individuals perturb their data locally before sharing it, ensuring that the raw data is never exposed. 
\begin{definition}[Local differential privacy (LDP)~\cite{evfimievski2003limiting, kasiviswanathan2011can}]
\label{definition:LDP}
A randomized mechanism $\mathcal{A}$ satisfies $\varepsilon$-LDP if for any two input values $x, x'$ and any output $y$, the following holds:
\begin{equation}
\forall T \subseteq Range(\mathcal{A}): \Pr[\mathcal{A}(x) \in T] \leq e^\varepsilon \Pr[\mathcal{A}(x') \in T],
\end{equation} 
\end{definition}
where $\epsilon \geq 0$ is the privacy parameter. Local differential privacy allows sharing of data points with an untrusted party, and the privacy of the individuals is protected by achieving indistinguishability from other possible data points. 

\subsubsection{Generalized Randomized Response}
The most well-known mechanism to ensure local differential privacy is the generalized randomized response (GRR). As shown in~\cite{wang2017locally}, when the size of the domain $d$ is small and we have $d < 3e^{\varepsilon}+2$, the generalized randomized response with the direct encoding scheme returns the most optimal result:
\begin{definition}[Direct Encoding GRR]
\label{definition:grr}
Given a domain of possible values $\mathcal{V}=\{v_1, v_2, ...,v_k\}$ and an input $v\in\mathcal{V}$, GRR perturbs $v$ into another value $v'\in\mathcal{V}$ such that:  
\begin{equation}
\Pr[\mathcal{A}(v) = v'] = \begin{cases}
    p = \frac{e^{\varepsilon}}{e^{\varepsilon}+d-1}& \text{if}\ v=v'\\
    q=\frac{1}{e^{\varepsilon}+d-1} &\text{if}\ v\neq v'
\end{cases}
\end{equation}
\end{definition}
Note that the size of the domain for our problem is $d=|\mathcal{V}=\{0, 1, 2\}|=3$, which is the 3 possible values of SNPs. The unbiased frequency $\tilde{f_v}$ can be estimated from the noisy frequency $f'_v$ as $\tilde{f_v} = \frac{f'_v-q}{p-q}$.

\myparagraph{Discussion} As discussed in Section~\ref{sec:bg-snps}, SNPs exhibit correlation with one another, with no defined limit for correlation length in genome sequences. Evidence suggests long-range linkage disequilibrium ($>250k$ nucleobases)\cite{koch2013long}, and no universal rules exist regarding correlation patterns. Consequently, the privacy budget $\varepsilon$ of the GRR mechanism theoretically scales with the sequence length $L$~\cite{chen2014correlated, zhang2022correlated}. To ensure a fair comparison between GRR and our HMM trained with a given $\varepsilon$, the GRR mechanism must use a privacy budget of $\varepsilon/L$ per SNP locus.

Another important consideration is the difference in privacy guarantees between the two approaches. The GRR mechanism satisfies pure $\varepsilon$-differential privacy (DP), whereas DP-SGD ensures $(\varepsilon, \delta)$-DP. This discrepancy complicates direct comparisons between the two methods. However, to the best of our knowledge, no alternative DP mechanism exists that would serve as a more suitable baseline for a fair comparison to our method.

%%%%%%%%%%%%%%%%%%%%%%%%%%%%%%%%%%%%%%%%%%%%%%%%%%%%%%%%%%%%%%%%%%%%%%%

\section{Non-private Experiments}
\label{sec:app-exp-nodp}

\begin{figure*}[ht!]
    \centering
    \hfill
    \hspace{-0.5cm}
    \begin{minipage}[b]{0.34\textwidth}
        \centering
        \includegraphics[width=\textwidth]{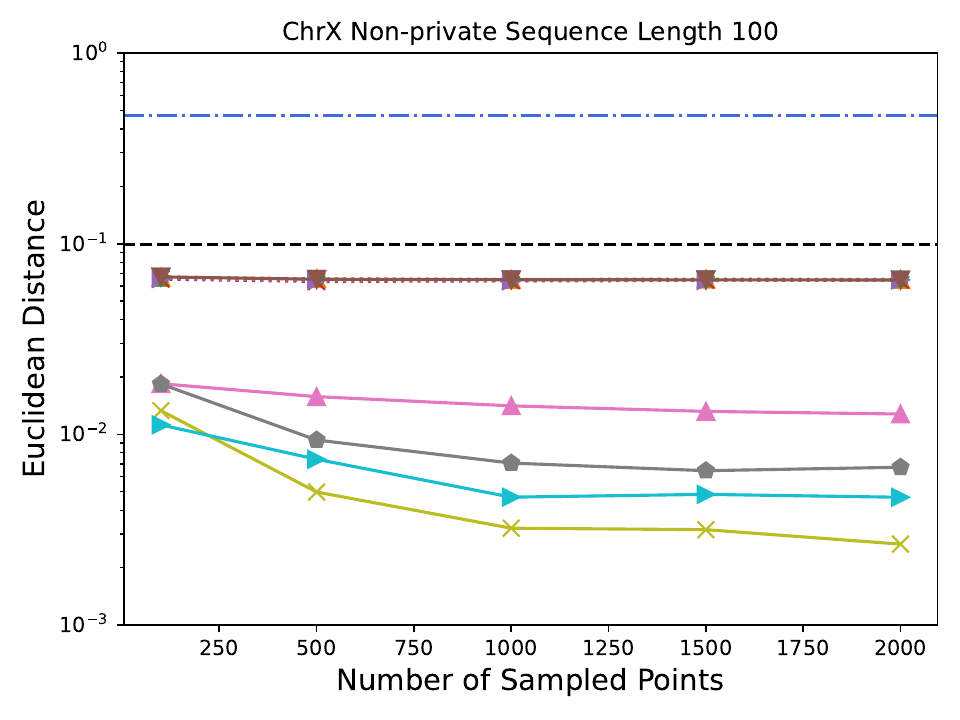}
    \end{minipage}%
    \hspace{-0.2cm}
    \begin{minipage}[b]{0.34\textwidth}
        \centering
        \includegraphics[width=\textwidth]{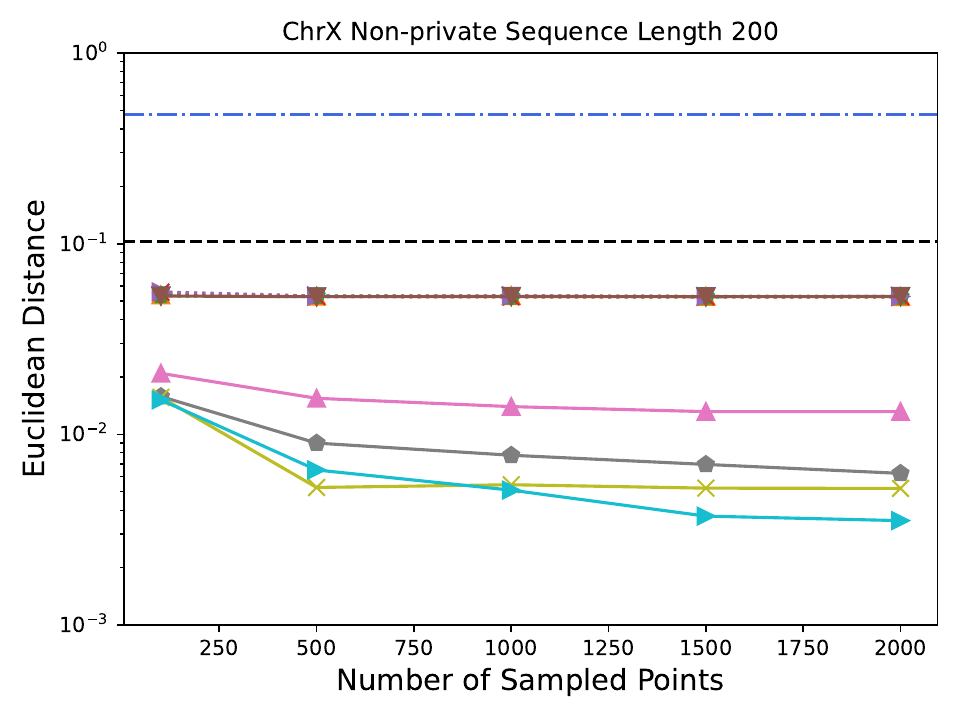}
    \end{minipage}%
    \hspace{-0.2cm}
    \begin{minipage}[b]{0.34\textwidth}
        \centering
        \includegraphics[width=\textwidth]{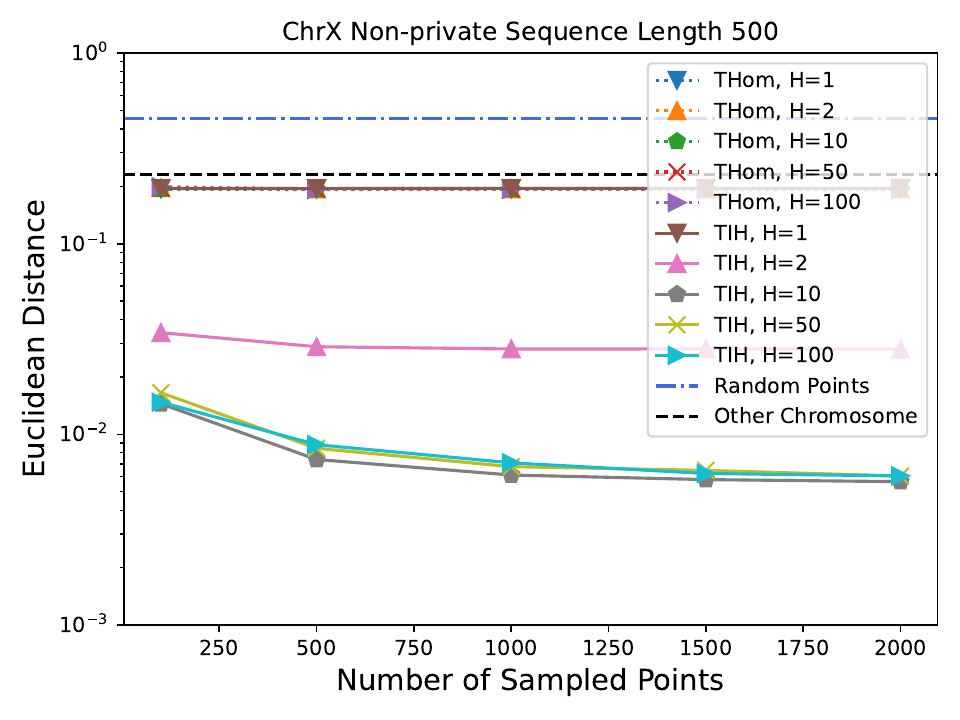}
    \end{minipage}
    
    \hfill
    \hspace{-0.5cm}
    \begin{minipage}[b]{0.34\textwidth}
        \centering
        \includegraphics[width=\textwidth]{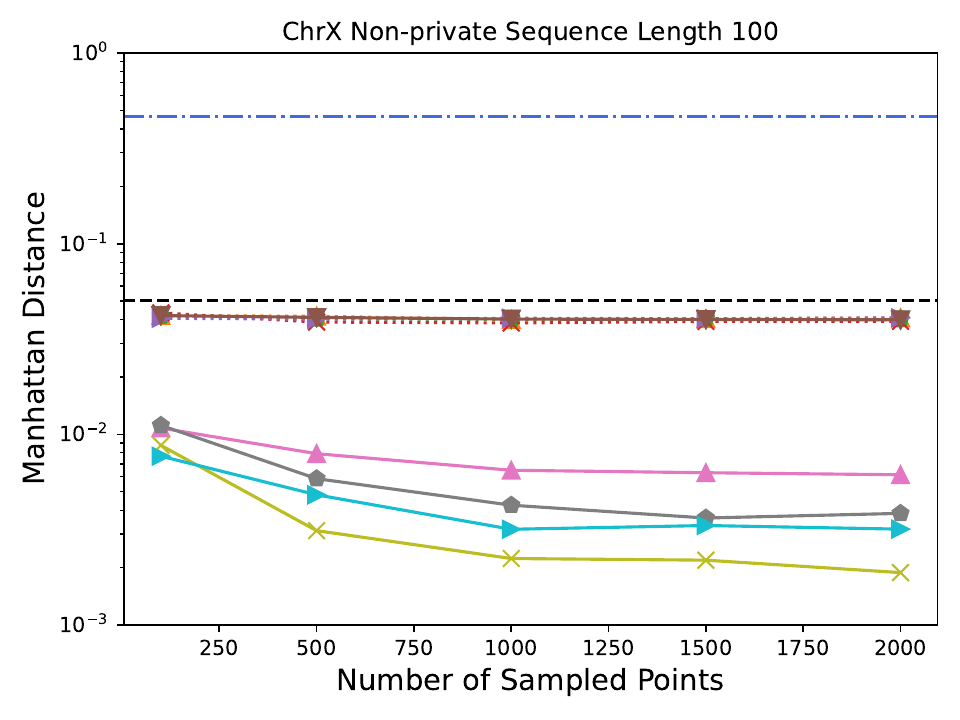}
    \end{minipage}%
    \hspace{-0.2cm}
    \begin{minipage}[b]{0.34\textwidth}
        \centering
        \includegraphics[width=\textwidth]{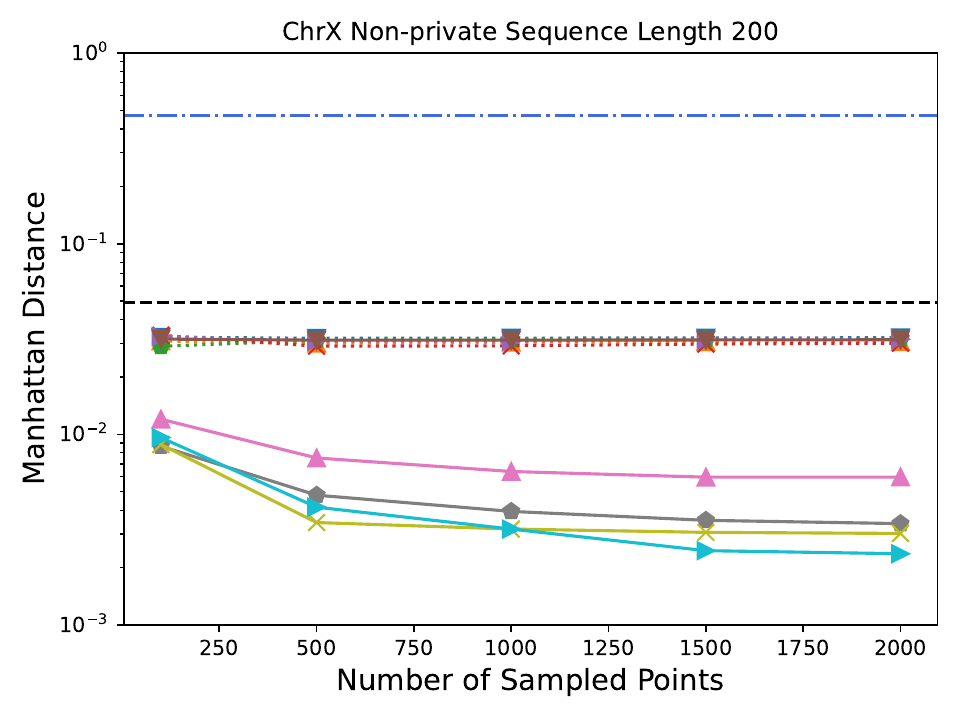}
    \end{minipage}%
    \hspace{-0.2cm}
    \begin{minipage}[b]{0.34\textwidth}
        \centering
        \includegraphics[width=\textwidth]{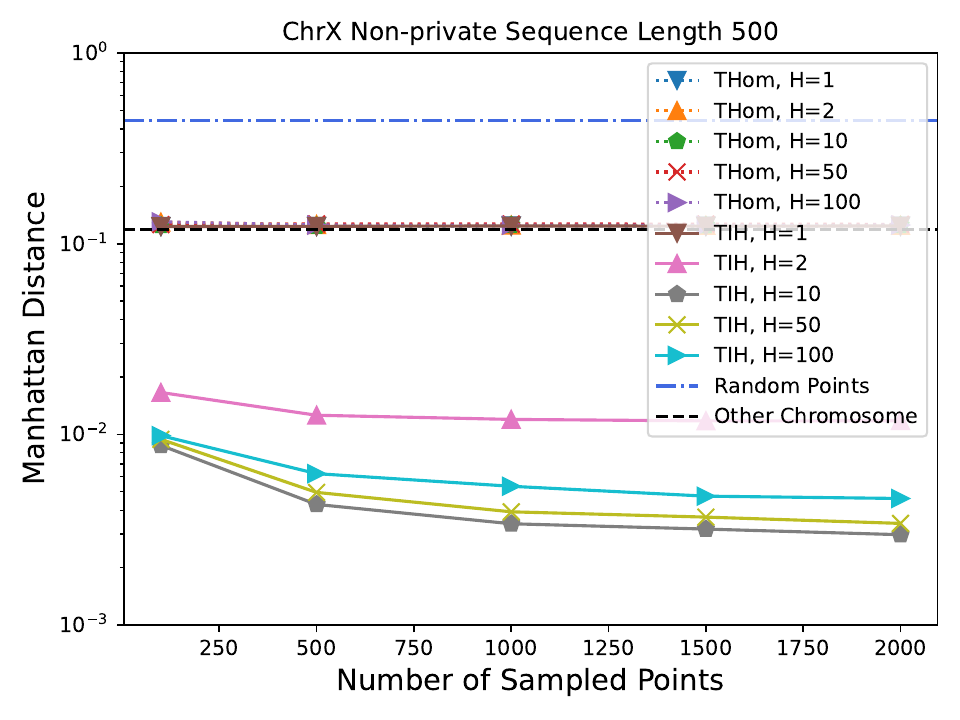}
    \end{minipage}
    
    \hfill
    \hspace{-0.5cm}
    \begin{minipage}[b]{0.34\textwidth}
        \centering
        \includegraphics[width=\textwidth]{figs/nodp-log-Nei_chrX_L100.pdf}
    \end{minipage}%
    \hspace{-0.2cm}
    \begin{minipage}[b]{0.34\textwidth}
        \centering
        \includegraphics[width=\textwidth]{figs/nodp-log-Nei_chrX_L200.pdf}
    \end{minipage}%
    \hspace{-0.2cm}
    \begin{minipage}[b]{0.34\textwidth}
        \centering
        \includegraphics[width=\textwidth]{figs/nodp-log-Nei_chrX_L500.pdf}
    \end{minipage}
    % \hfill
    % \hspace{-0.5cm}
    % \begin{minipage}[b]{0.34\textwidth}
    %     \centering
    %     \includegraphics[width=\textwidth]{figs/nodp-log-Wasserstein_chrX_L100.pdf}
    % \end{minipage}%
    % \hspace{-0.2cm}
    % \begin{minipage}[b]{0.34\textwidth}
    %     \centering
    %     \includegraphics[width=\textwidth]{figs/nodp-log-Wasserstein_chrX_L200.pdf}
    % \end{minipage}%
    % \hspace{-0.2cm}
    % \begin{minipage}[b]{0.34\textwidth}
    %     \centering
    %     \includegraphics[width=\textwidth]{figs/nodp-log-Wasserstein_chrX_L500.pdf}
    % \end{minipage}
    
    \caption{All distance measures for non-private training with chromosome X.}
    \label{fig:all-distances_chrx-nodp}
\end{figure*}

\begin{figure*}[ht!]
    \centering
    \hfill
    \hspace{-0.5cm}
    \begin{minipage}[b]{0.34\textwidth}
        \centering
        \includegraphics[width=\textwidth]{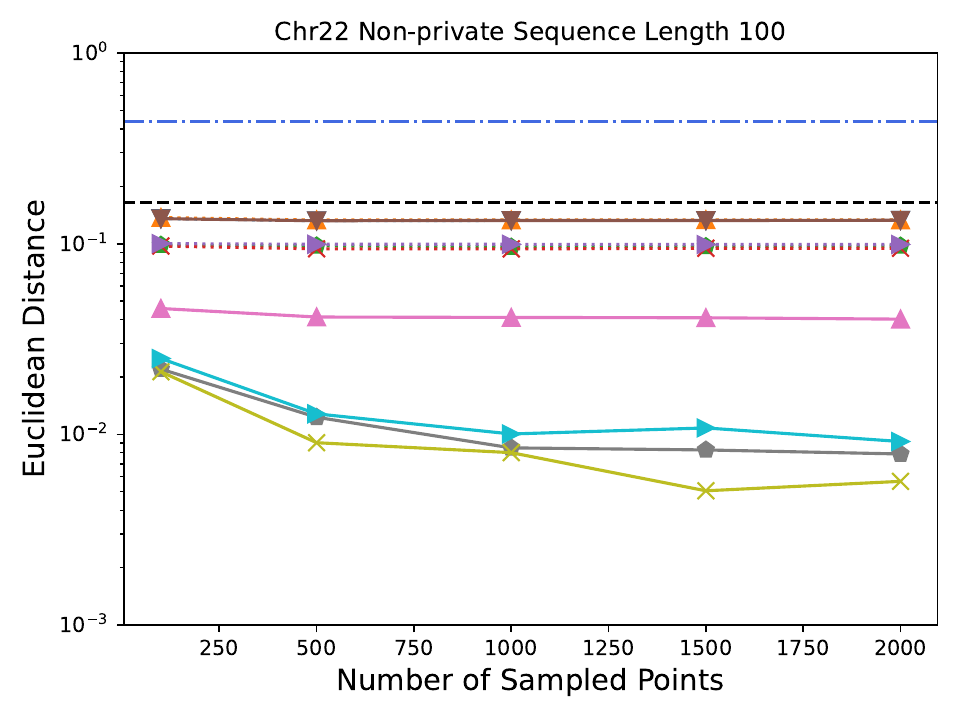}
    \end{minipage}%
    \hspace{-0.2cm}
    \begin{minipage}[b]{0.34\textwidth}
        \centering
        \includegraphics[width=\textwidth]{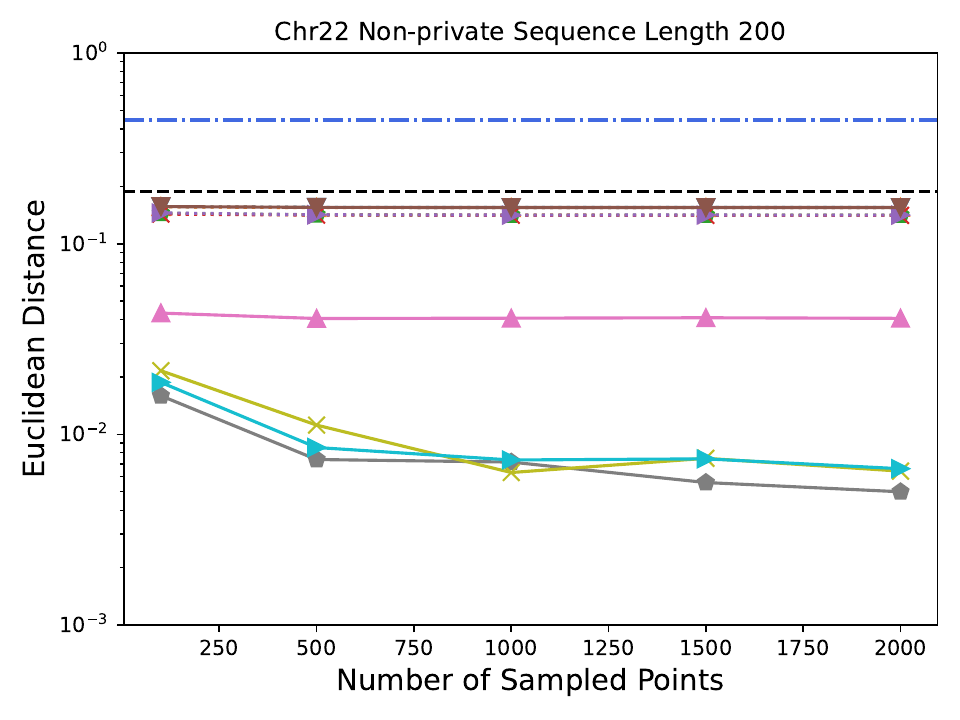}
    \end{minipage}%
    \hspace{-0.2cm}
    \begin{minipage}[b]{0.34\textwidth}
        \centering
        \includegraphics[width=\textwidth]{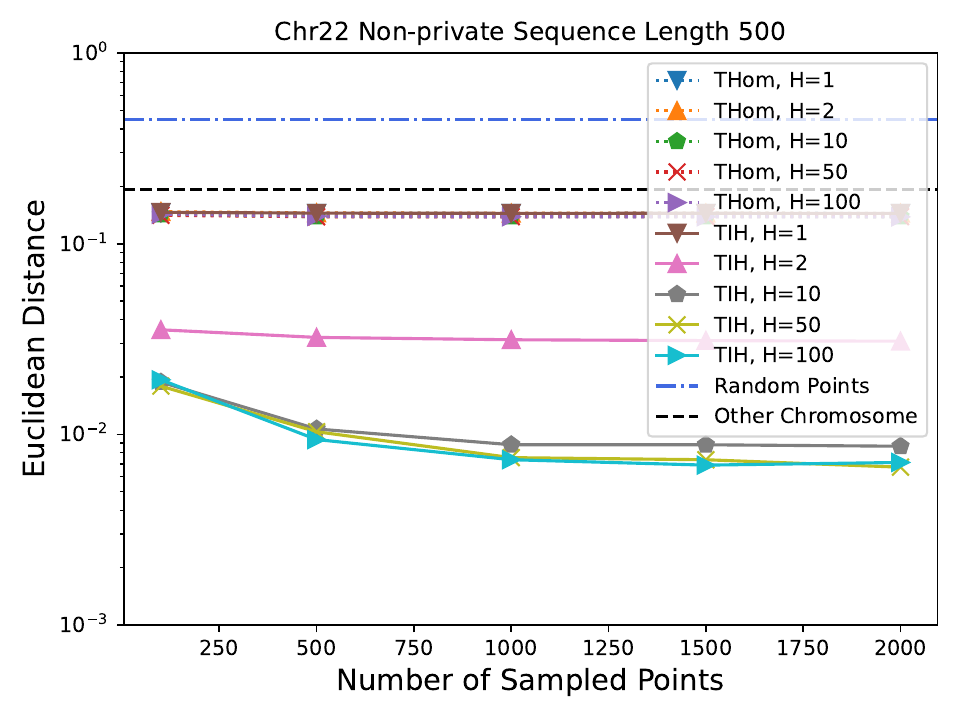}
    \end{minipage}
    
    \hfill
    \hspace{-0.5cm}
    \begin{minipage}[b]{0.34\textwidth}
        \centering
        \includegraphics[width=\textwidth]{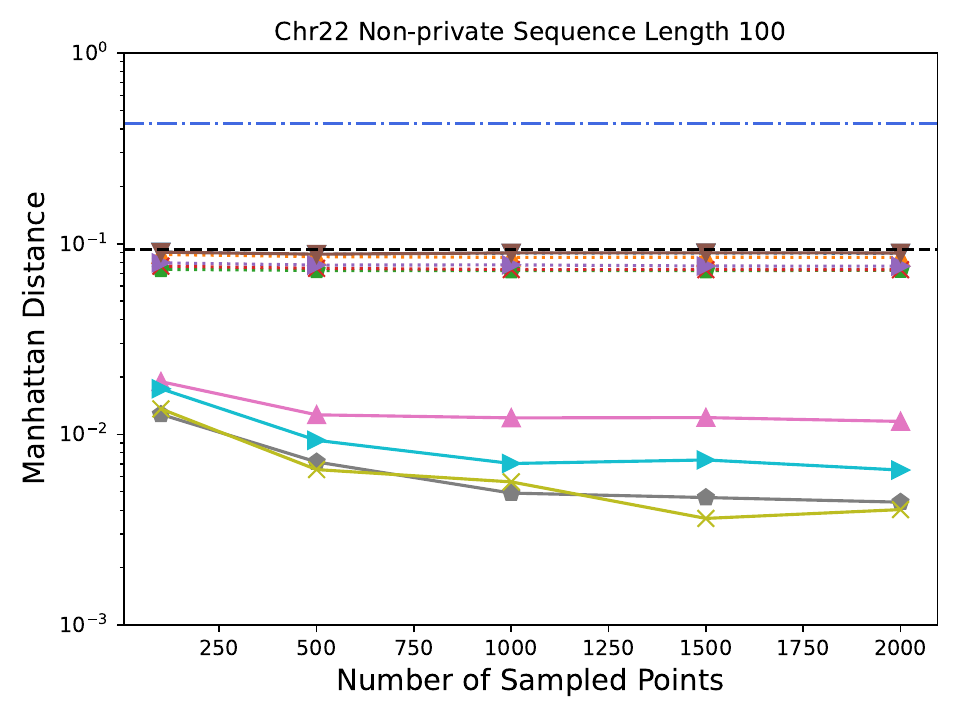}
    \end{minipage}%
    \hspace{-0.2cm}
    \begin{minipage}[b]{0.34\textwidth}
        \centering
        \includegraphics[width=\textwidth]{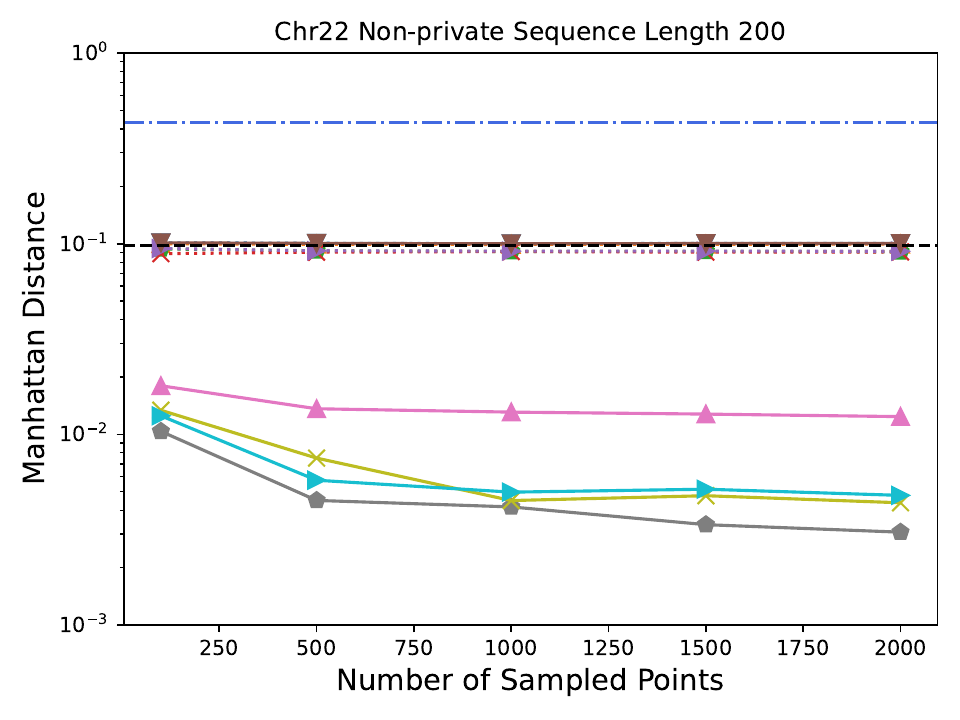}
    \end{minipage}%
    \hspace{-0.2cm}
    \begin{minipage}[b]{0.34\textwidth}
        \centering
        \includegraphics[width=\textwidth]{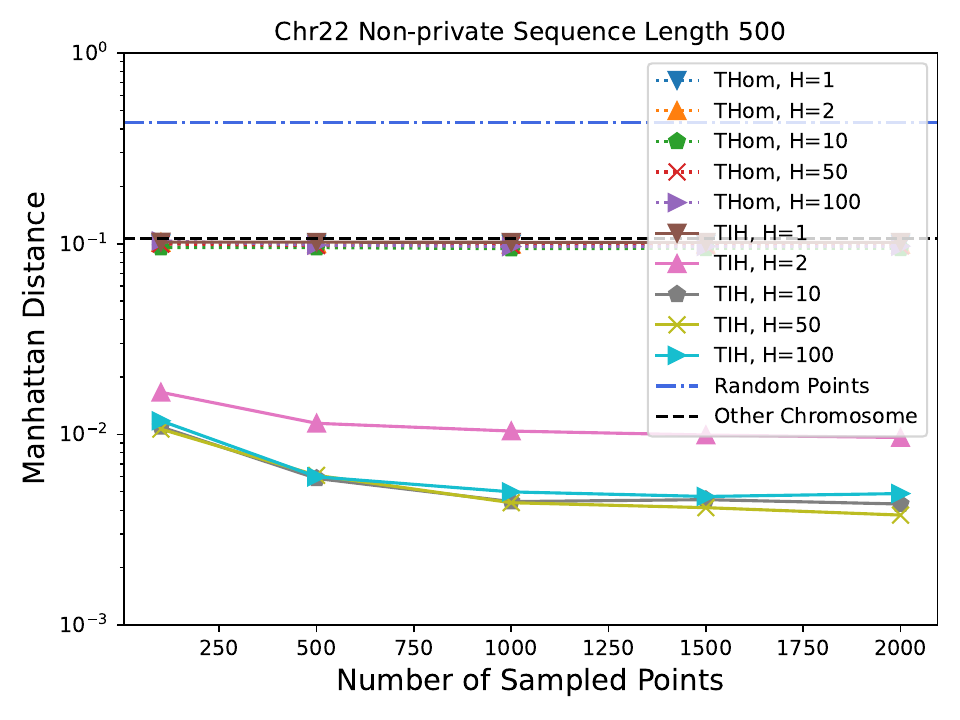}
    \end{minipage}
    
    \hfill
    \hspace{-0.5cm}
    \begin{minipage}[b]{0.34\textwidth}
        \centering
        \includegraphics[width=\textwidth]{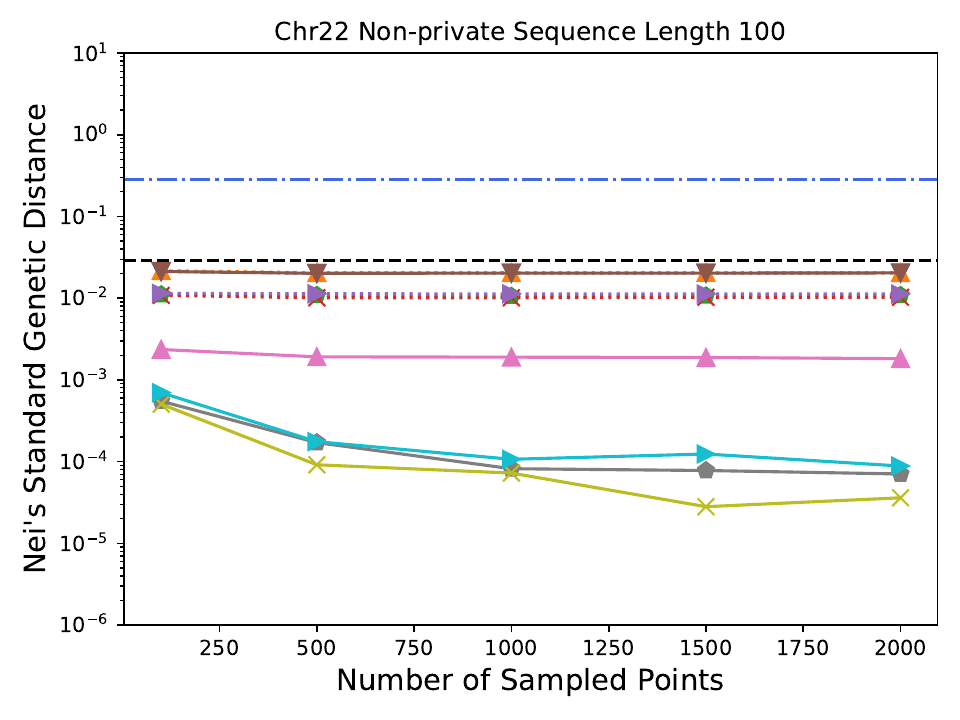}
    \end{minipage}%
    \hspace{-0.2cm}
    \begin{minipage}[b]{0.34\textwidth}
        \centering
        \includegraphics[width=\textwidth]{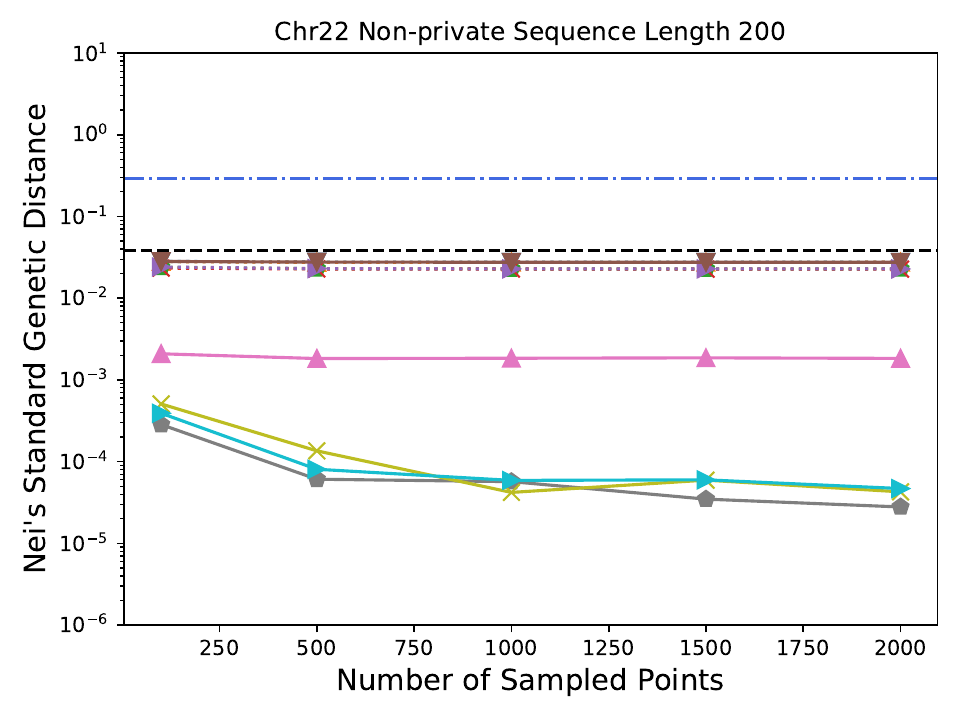}
    \end{minipage}%
    \hspace{-0.2cm}
    \begin{minipage}[b]{0.34\textwidth}
        \centering
        \includegraphics[width=\textwidth]{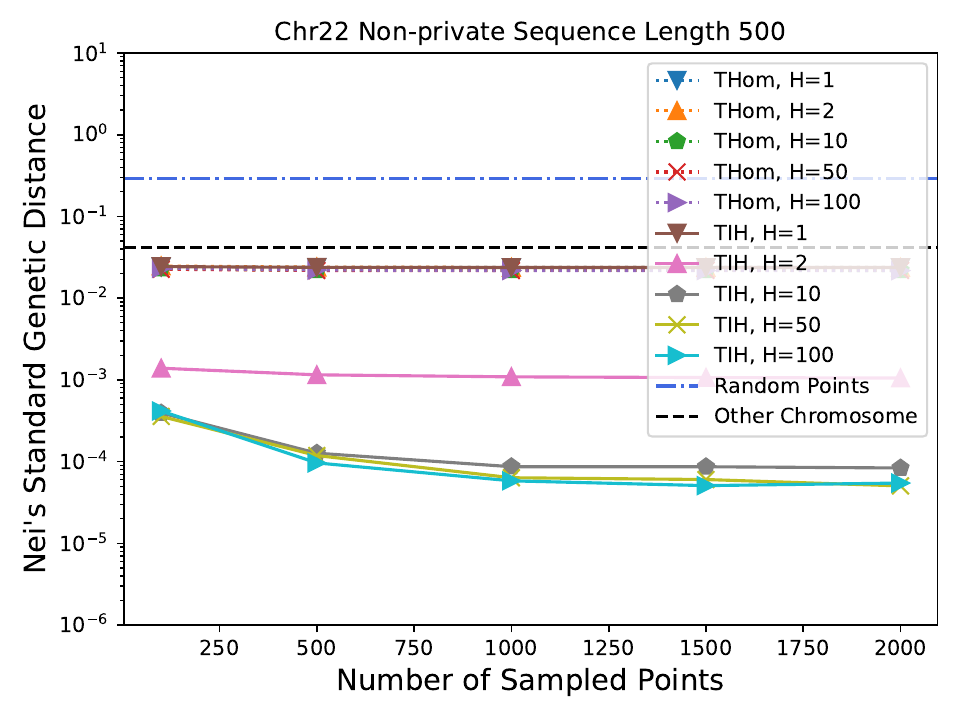}
    \end{minipage}
    % \hfill
    % \hspace{-0.5cm}
    % \begin{minipage}[b]{0.34\textwidth}
    %     \centering
    %     \includegraphics[width=\textwidth]{figs/nodp-log-Wasserstein_chr22_L100.pdf}
    % \end{minipage}%
    % \hspace{-0.2cm}
    % \begin{minipage}[b]{0.34\textwidth}
    %     \centering
    %     \includegraphics[width=\textwidth]{figs/nodp-log-Wasserstein_chr22_L200.pdf}
    % \end{minipage}%
    % \hspace{-0.2cm}
    % \begin{minipage}[b]{0.34\textwidth}
    %     \centering
    %     \includegraphics[width=\textwidth]{figs/nodp-log-Wasserstein_chr22_L500.pdf}
    % \end{minipage}
    
    \caption{All distance measures for non-private training with chromosome 22.}
    \label{fig:all-distances_chr22-nodp}
\end{figure*}

\myparagraph{Distance measures} Figure~\ref{fig:all-distances_chrx-nodp} and Figure~\ref{fig:all-distances_chr22-nodp} illustrate the utility of our models across different sequence lengths ($L \in \{100, 200, 500\}$) and sample sizes ($N \in \{100, 500, 1000, 1500, 2000\}$). 

The time-homogeneous (THom) models exhibit consistent behavior across all distance measures, showing no improvement with increasing model capacity ($H \in \{1, 2, 10, 50, 100\}$). In contrast, the time-inhomogeneous (TIH) models demonstrate a clear performance gain with increasing $H$, with the most significant improvement occurring after $H = 2$. TIH models consistently achieve low distances between generated and real data, with Nei's distances below $10^{-4}$ and Manhattan/Euclidean distances below $10^{-2}$ across all lengths and metrics for $N = 2000$.

\begin{figure*}[ht!]
    \centering
    \hspace{-0.cm}
    \begin{minipage}[b]{0.45\textwidth}
        \centering
        \includegraphics[width=\textwidth]{figs/nodp_hist_chrX_L100_ns2000.pdf}
    \end{minipage}%
    \hspace{-0.cm}
    \begin{minipage}[b]{0.45\textwidth}
        \centering
        \includegraphics[width=\textwidth]{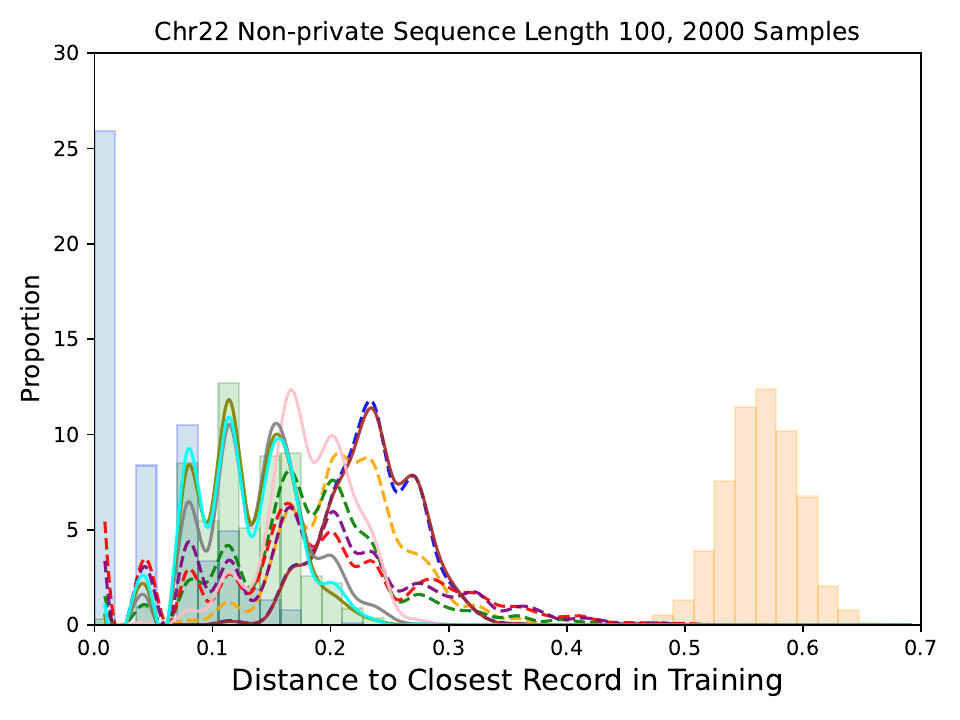}
    \end{minipage}%
    \hspace{-0.cm}
    \begin{minipage}[b]{0.45\textwidth}
        \centering
        \includegraphics[width=\textwidth]{figs/nodp_hist_chrX_L200_ns2000.pdf}
    \end{minipage}
    \hspace{-0.cm}
    \begin{minipage}[b]{0.45\textwidth}
        \centering
        \includegraphics[width=\textwidth]{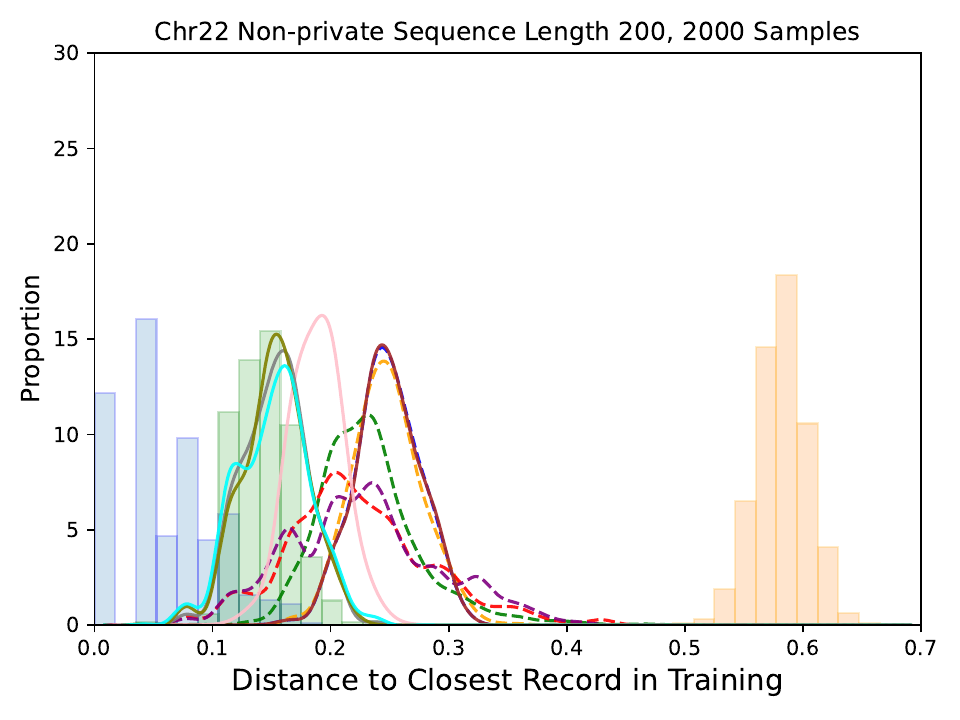}
    \end{minipage}%
    \hspace{-0.cm}
    \begin{minipage}[b]{0.45\textwidth}
        \centering
        \includegraphics[width=\textwidth]{figs/nodp_hist_chrX_L500_ns2000.pdf}
    \end{minipage}%
    \hspace{-0.cm}
    \begin{minipage}[b]{0.45\textwidth}
        \centering
        \includegraphics[width=\textwidth]{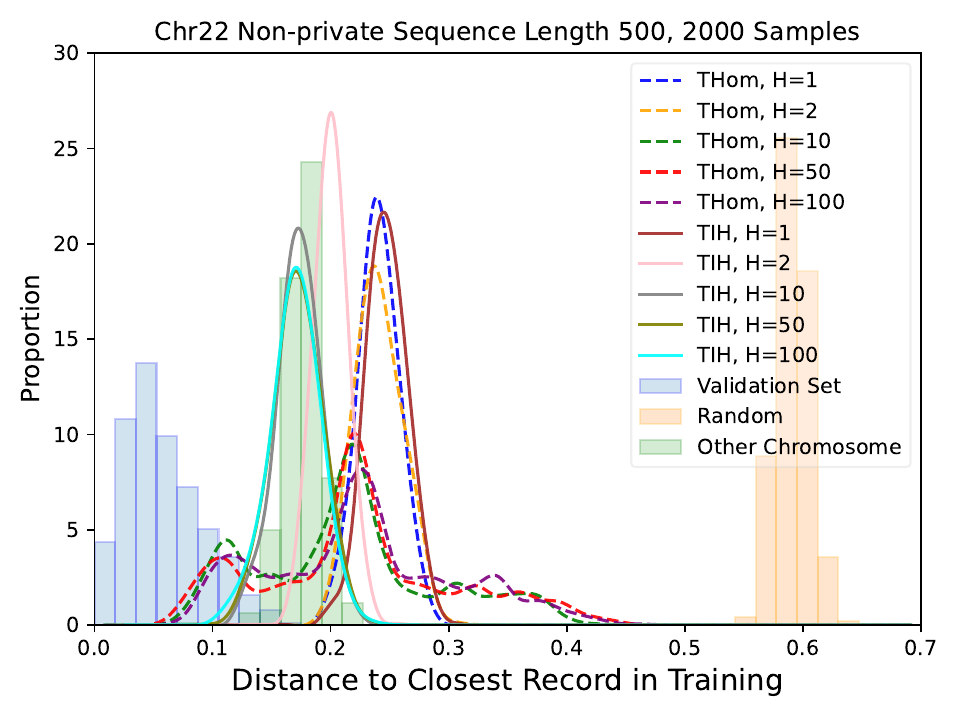}
    \end{minipage}
    \caption{Histograms of distances to the closest record in training for the time-homogeneous (THom) and time-inhomogeneous (TIH) models and different number of hidden states $H$, for chromosome 22 and chromosome X.}\label{fig:histograms-both-datasets-ns2000}
\end{figure*}

\myparagraph{Histograms of $l_2$ distance to the closest record in training} We present the results for histograms of distances between each synthetic point and its closest neighbor in the training set in Figure~\ref{fig:histograms-both-datasets-ns2000}, considering $N = 2000$ samples. For comparison, we also include histograms of distances to the training set for the hold-out validation set, another chromosome, and randomly generated points. To enhance clarity, we use cubic splines (degree 3) to connect the midpoints of the histograms for synthetic samples generated by the THom and TIH models, with the number of hidden states denoted as $H$.

For all sequence lengths, the histograms show that THom models exhibit a longer right tail compared to TIH models, indicating the THom model's difficulty in generating synthetic points similar to the training dataset. This discrepancy becomes more pronounced as the sequence length increases. At length $L = 500$, the peaks of the two models (TIH and THom) become distinctly separated, with the mean distances for samples from the THom model shifting closer to those of random points. 

Additionally, both TIH and THom models exhibit identical behavior for $H = 1$. For TIH with $H = 2$, we observe a heavier right tail, particularly at length $L = 500$, where its peak visibly shifts to the right. However, for higher numbers of hidden states, no significant differences or improvements are observed between the models.

%%%%%%%%%%%%%%%%%%%%%%%%%%%%%%%%%%%%%%%%%%%%%%%%%%%%%%%%%%%%%%%%%%%%%%%%
%%%%%%%%%%%%%%%%%%%%%%%%%%%%%%%%%%%%%%%%%%%%%%%%%%%%%%%%%%%%%%%%%%%%%%%

\section{Differentially Private Experiments}
\label{sec:app-exp-dp}

\begin{figure*}[ht!]
    \hfill
    \hspace{-0.9cm}
    \begin{minipage}[b]{0.34\textwidth}
        \centering
        \includegraphics[width=\textwidth]{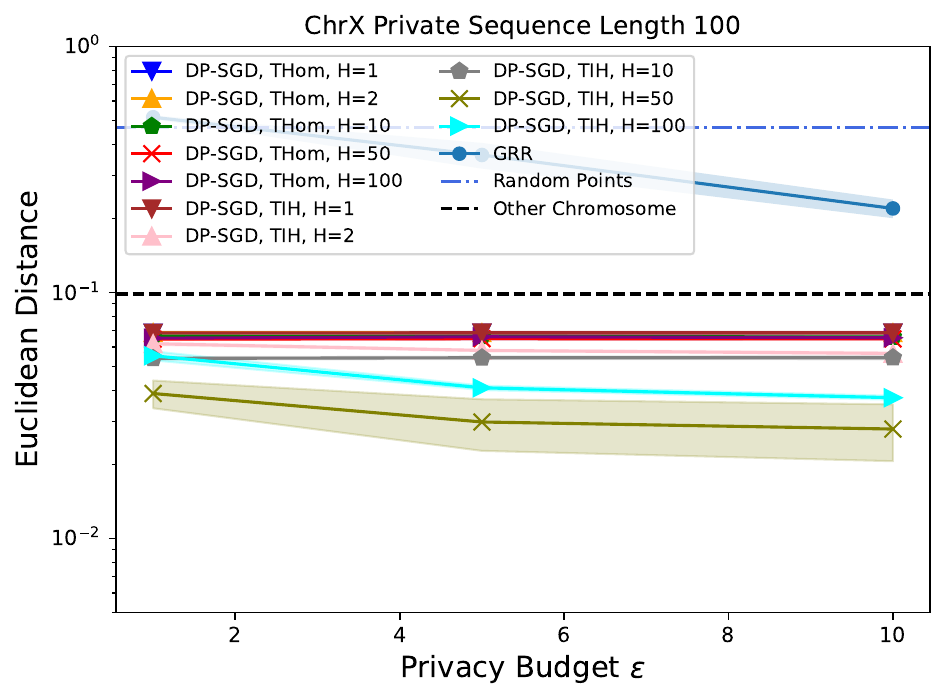}
    \end{minipage}%
    \hspace{-0.1cm}
    \begin{minipage}[b]{0.34\textwidth}
        \centering
        \includegraphics[width=\textwidth]{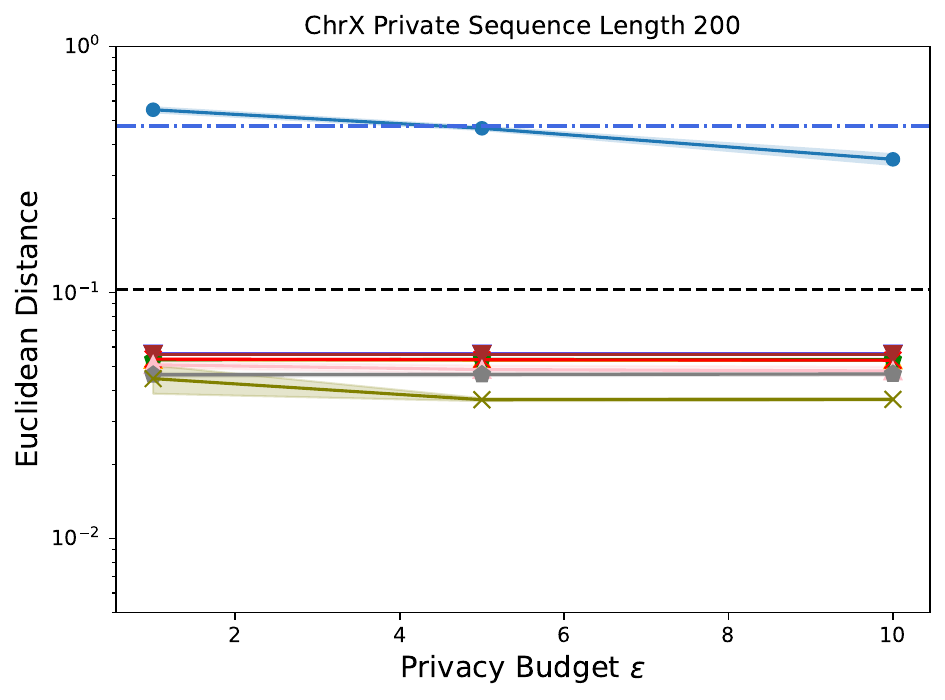}
    \end{minipage}%
    \hspace{-0.1cm}
    \begin{minipage}[b]{0.34\textwidth}
        \centering
        \includegraphics[width=\textwidth]{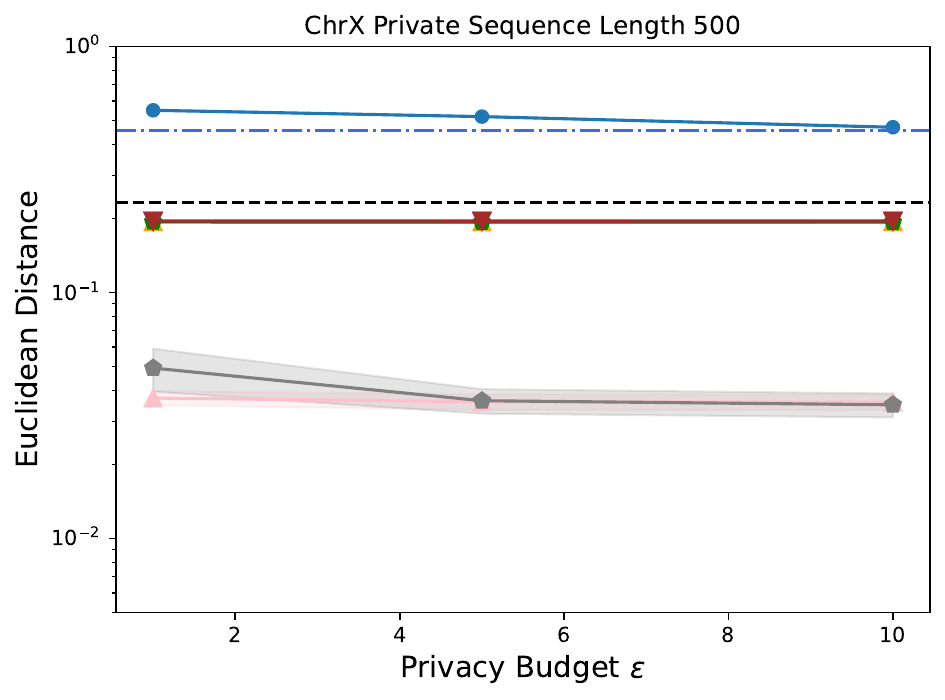}
    \end{minipage}
    
    \hfill
    \hspace{-0.9cm}
    \begin{minipage}[b]{0.34\textwidth}
        \centering
        \includegraphics[width=\textwidth]{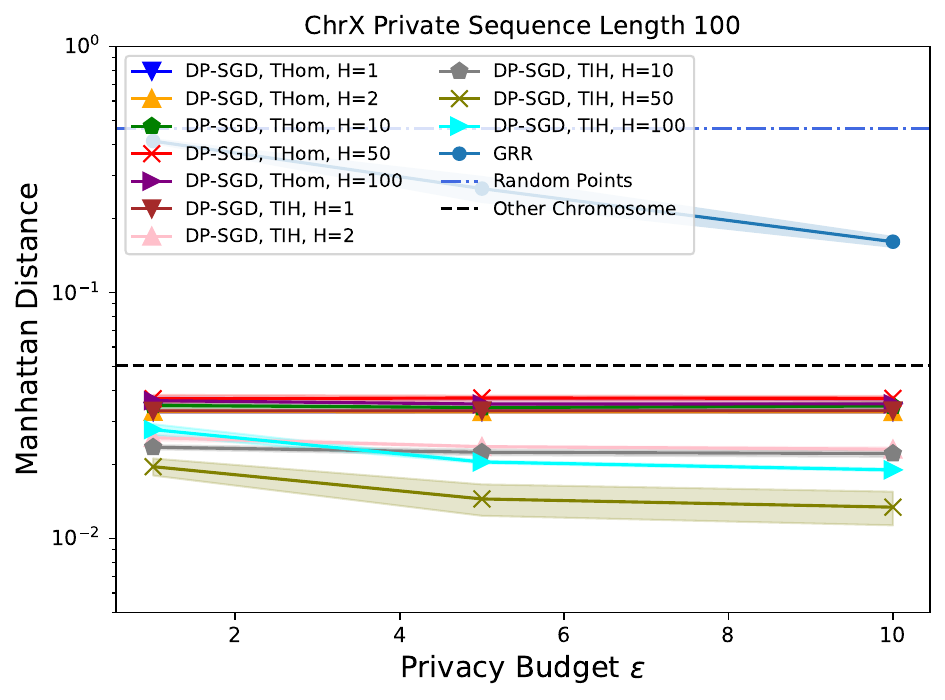}
    \end{minipage}%
    \hspace{-0.1cm}
    \begin{minipage}[b]{0.34\textwidth}
        \centering
        \includegraphics[width=\textwidth]{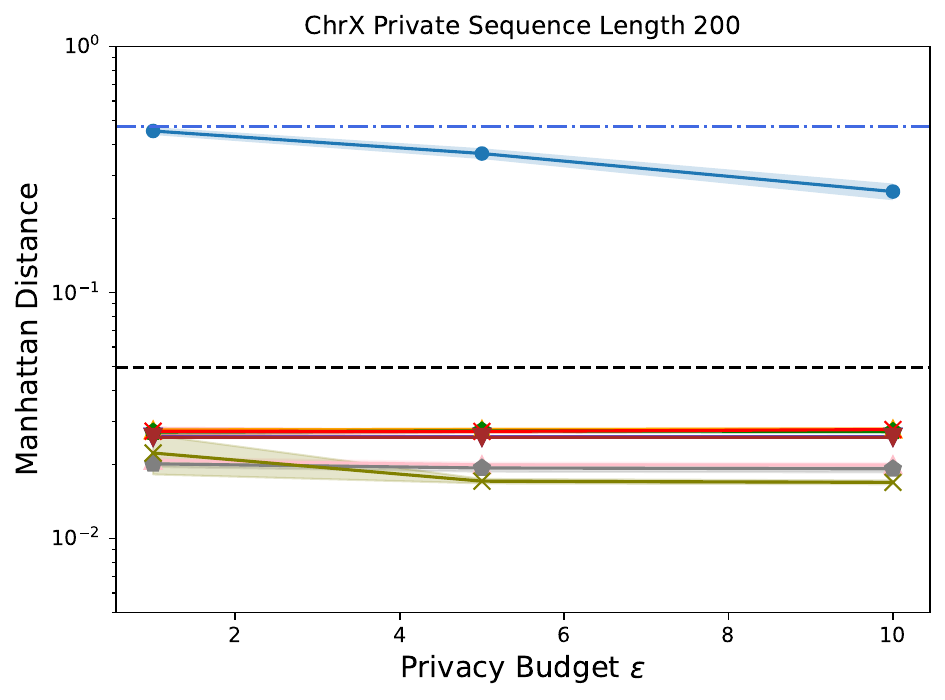}
    \end{minipage}%
    \hspace{-0.1cm}
    \begin{minipage}[b]{0.34\textwidth}
        \centering
        \includegraphics[width=\textwidth]{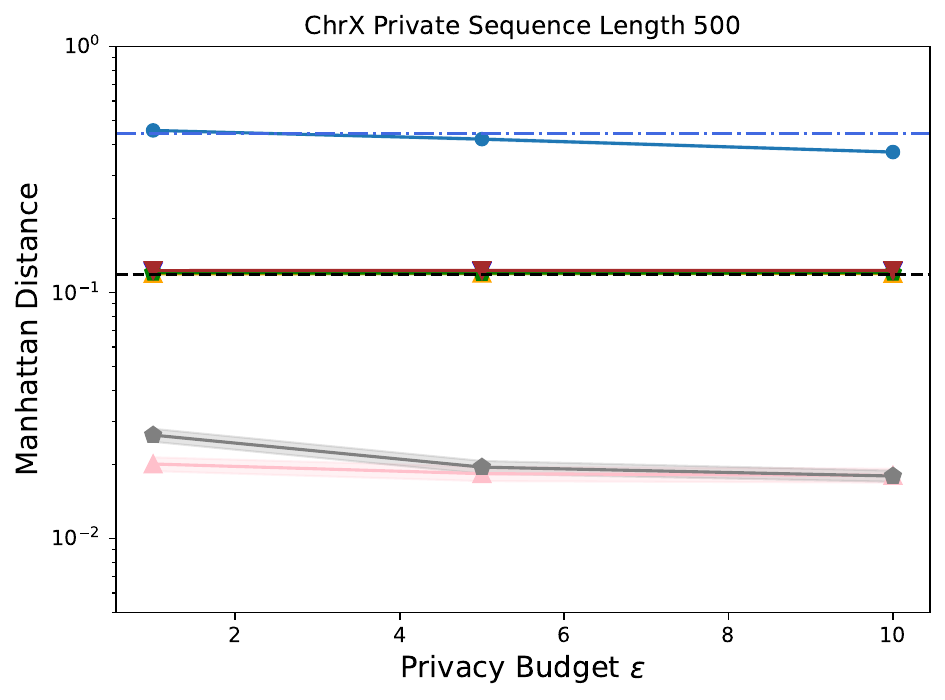}
    \end{minipage}
    
    \hfill
    \hspace{-0.9cm}
    \begin{minipage}[b]{0.34\textwidth}
        \centering
        \includegraphics[width=\textwidth]{figs/dp_Nei_chrX_L100.pdf}
    \end{minipage}%
    \hspace{-0.1cm}
    \begin{minipage}[b]{0.34\textwidth}
        \centering
        \includegraphics[width=\textwidth]{figs/dp_Nei_chrX_L200.pdf}
    \end{minipage}%
    \hspace{-0.1cm}
    \begin{minipage}[b]{0.34\textwidth}
        \centering
        \includegraphics[width=\textwidth]{figs/dp_Nei_chrX_L500.pdf}
    \end{minipage}
    
    \caption{All distance measures for DP-trained models. The GRR baseline is shown in blue. We set $\delta=10^{-4}$ for DP-SGD trained HMMs.}
    \label{fig:all-distances_chrx-dp}
\end{figure*}

\begin{figure*}[ht!]
    \hfill
    \hspace{-0.9cm}
    \begin{minipage}[b]{0.34\textwidth}
        \centering
        \includegraphics[width=\textwidth]{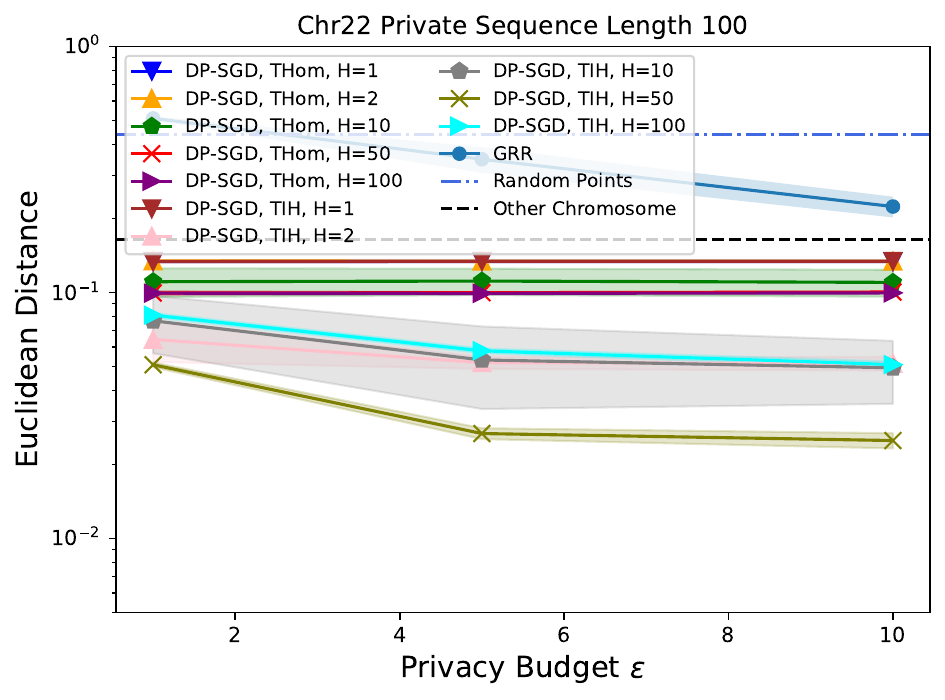}
    \end{minipage}%
    \hspace{-0.1cm}
    \begin{minipage}[b]{0.34\textwidth}
        \centering
        \includegraphics[width=\textwidth]{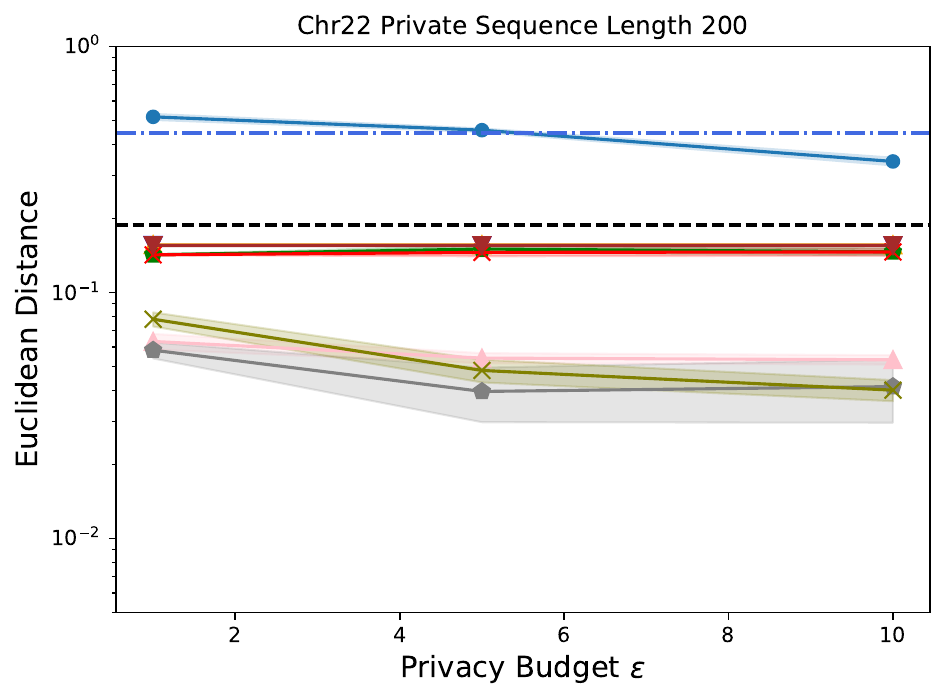}
    \end{minipage}%
    \hspace{-0.1cm}
    \begin{minipage}[b]{0.34\textwidth}
        \centering
        \includegraphics[width=\textwidth]{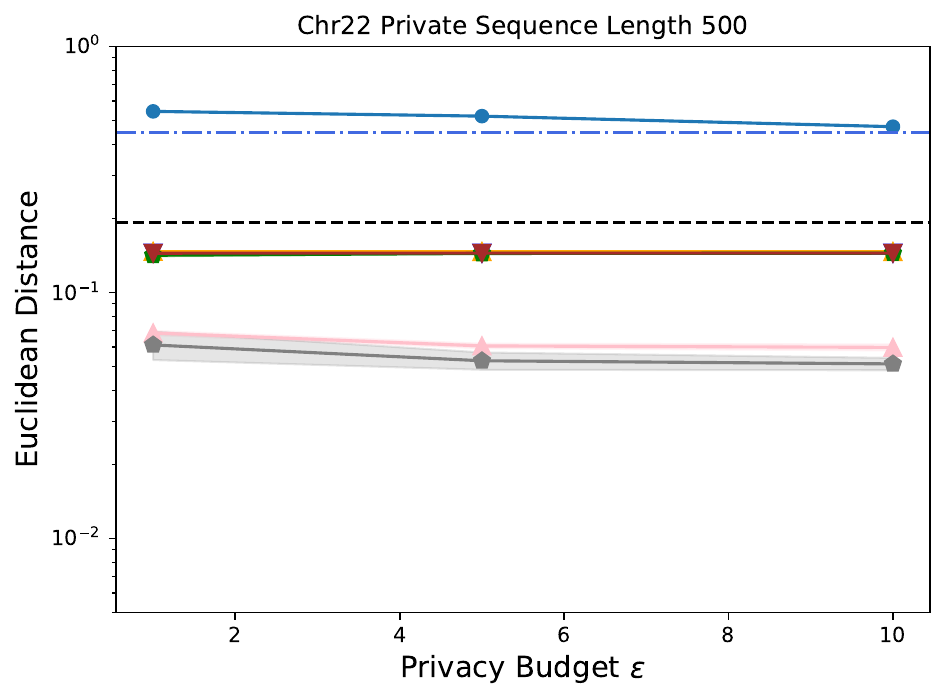}
    \end{minipage}
    
    \hfill
    \hspace{-0.9cm}
    \begin{minipage}[b]{0.34\textwidth}
        \centering
        \includegraphics[width=\textwidth]{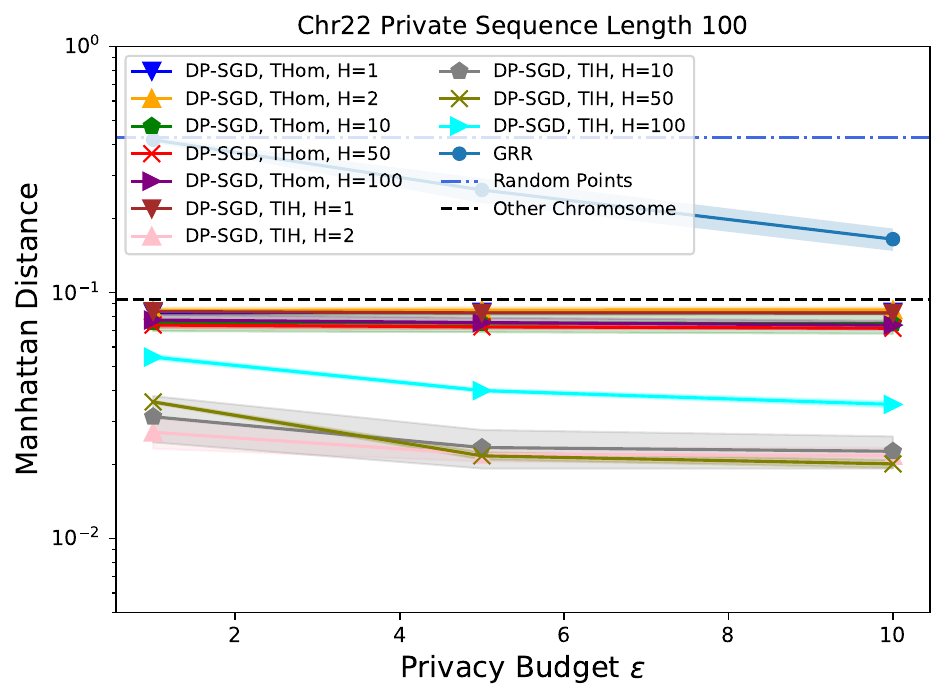}
    \end{minipage}%
    \hspace{-0.1cm}
    \begin{minipage}[b]{0.34\textwidth}
        \centering
        \includegraphics[width=\textwidth]{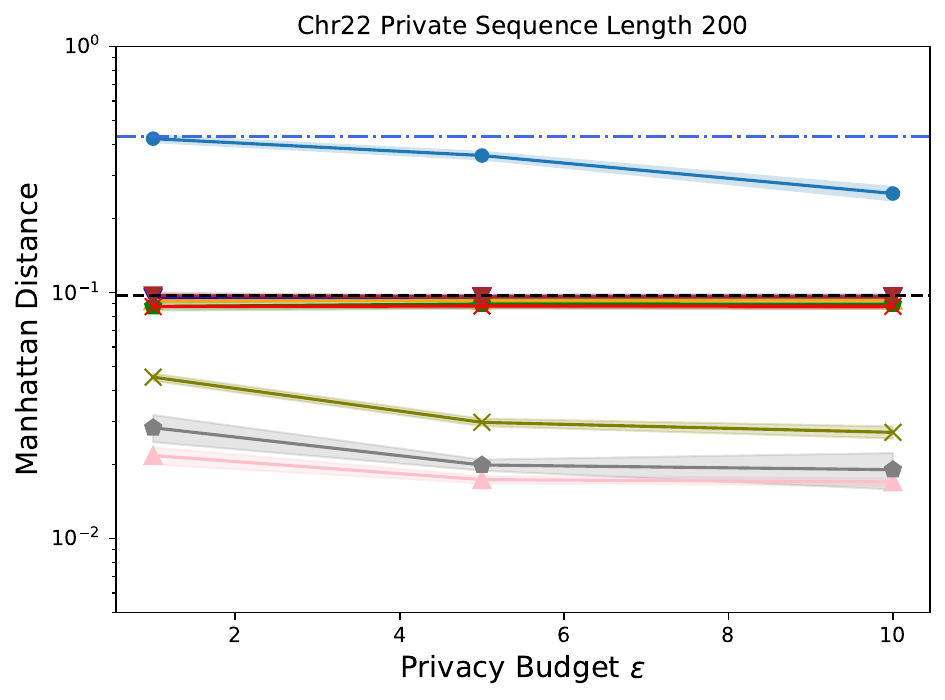}
    \end{minipage}%
    \hspace{-0.1cm}
    \begin{minipage}[b]{0.34\textwidth}
        \centering
        \includegraphics[width=\textwidth]{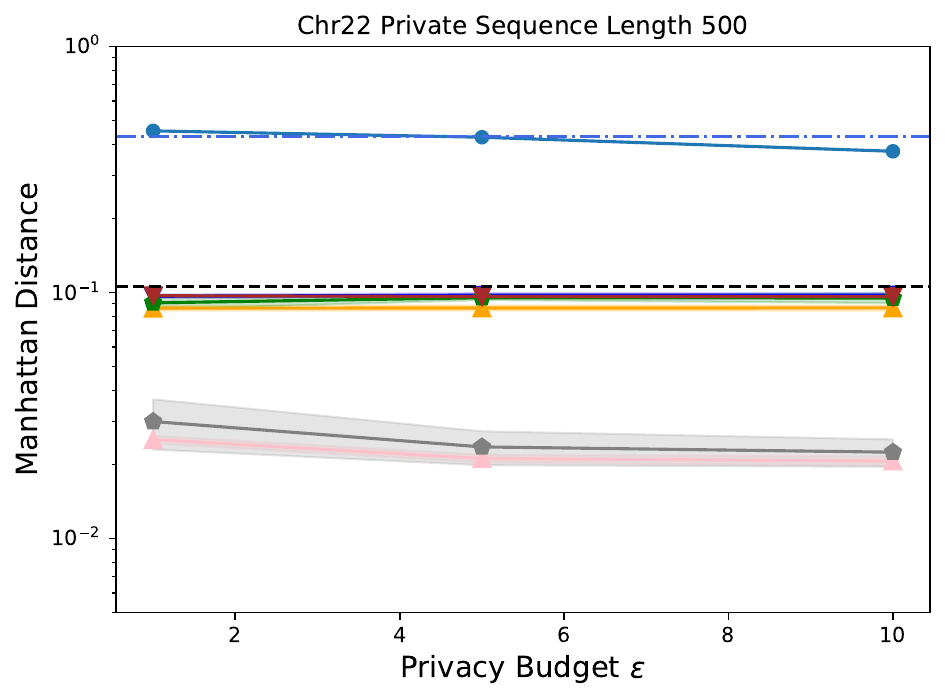}
    \end{minipage}
    
    \hfill
    \hspace{-0.9cm}
    \begin{minipage}[b]{0.34\textwidth}
        \centering
        \includegraphics[width=\textwidth]{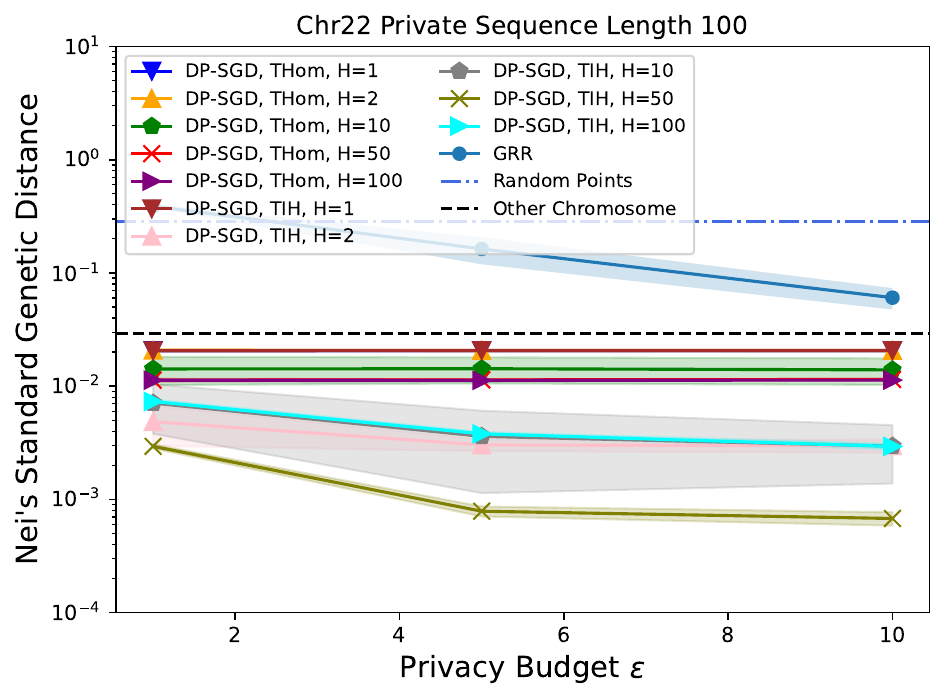}
    \end{minipage}%
    \hspace{-0.1cm}
    \begin{minipage}[b]{0.34\textwidth}
        \centering
        \includegraphics[width=\textwidth]{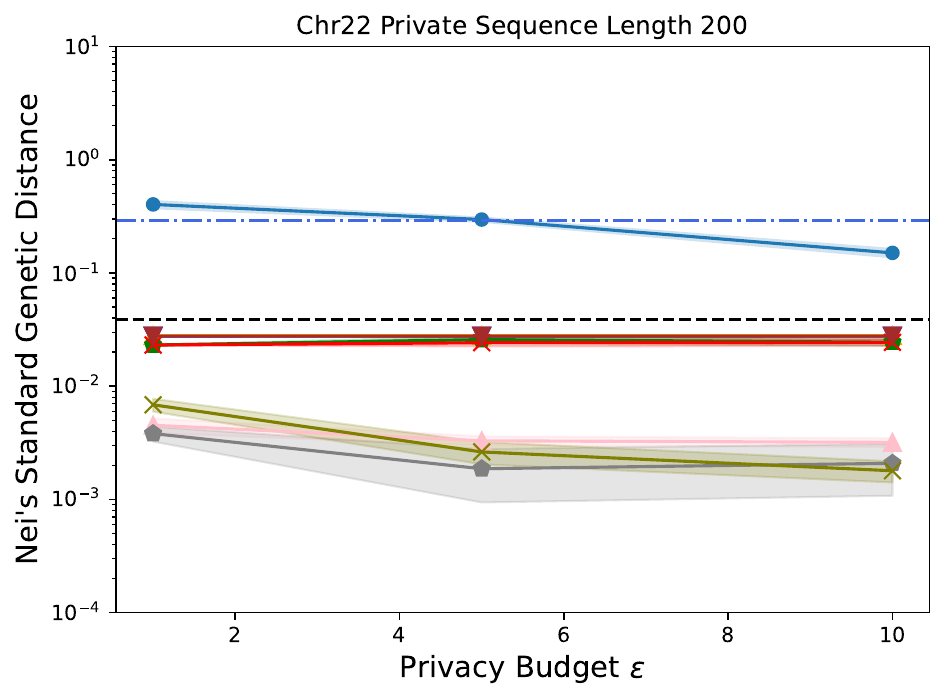}
    \end{minipage}%
    \hspace{-0.1cm}
    \begin{minipage}[b]{0.34\textwidth}
        \centering
        \includegraphics[width=\textwidth]{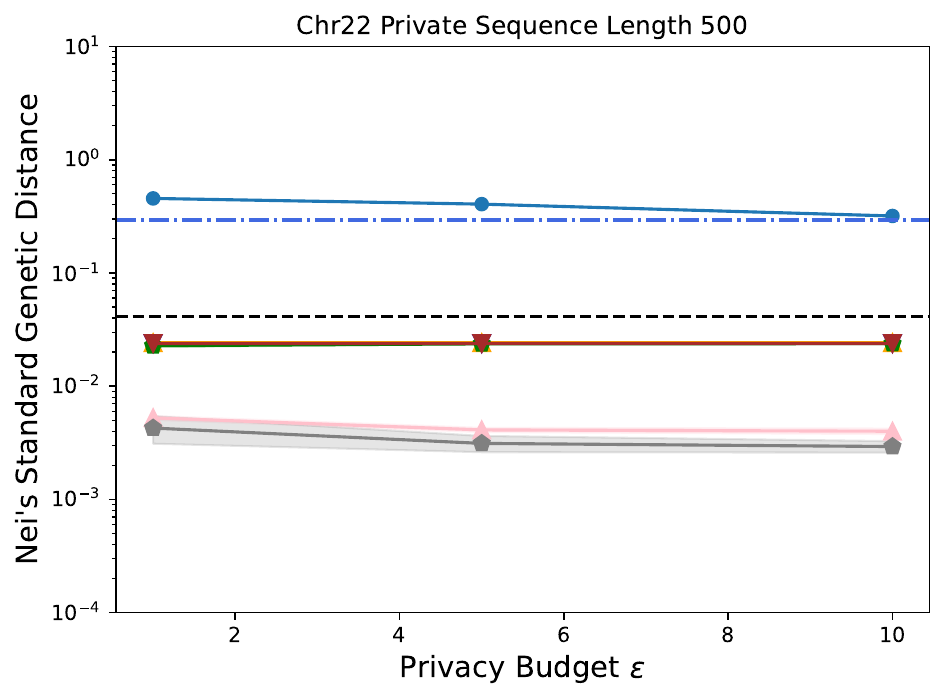}
    \end{minipage}
    
    \caption{All distance measures for DP-trained models. The GRR baseline is shown in blue. We set $\delta=10^{-4}$ for DP-SGD trained HMMs.}
    \label{fig:all-distances_chr22-dp}
\end{figure*}

\myparagraph{Distance measures} Figure~\ref{fig:all-distances_chrx-dp} and Figure~\ref{fig:all-distances_chr22-dp} present the distance measures for DP-trained HMMs, alongside the generalized randomized response (GRR) baseline (shown in blue). Markers indicate the mean of three DP experiment runs with different random seeds, with shaded regions representing standard deviations.

Across all metrics, the GRR baseline consistently underperforms relative to HMMs, demonstrating that applying theoretically correct local differential privacy renders the output of this mechanism ineffective for this privacy regime ($\varepsilon \in \{1, 5, 10\}$). The THom models again exhibit no sensitivity to varying privacy levels or hidden state capacities ($H$), particularly at $L=500$, where they fail to capture dataset structure at longer sequence lengths.

For $L=100$, the TIH model with $H=100$ underperforms compared to lower-capacity models. This aligns with expectations, as the DP-SGD noise disproportionately impacts more complex models, degrading performance.

\begin{figure*}
    \centering
    \includegraphics[width=\textwidth]{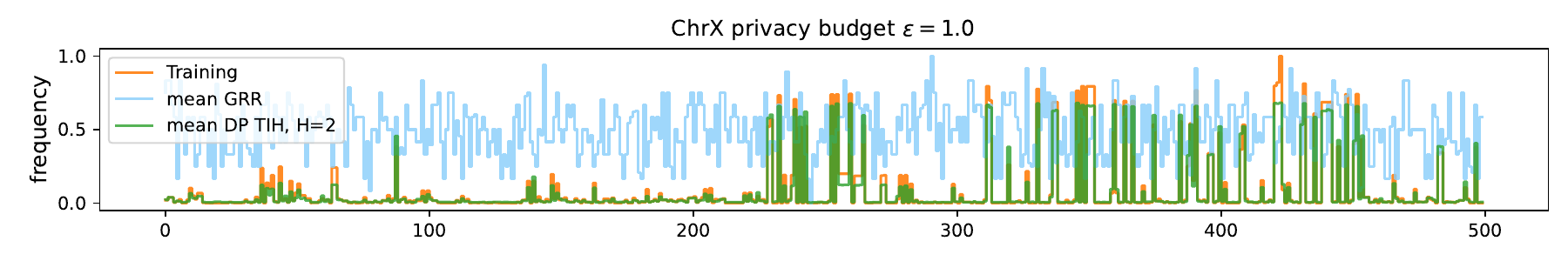}
    \includegraphics[width=\textwidth]{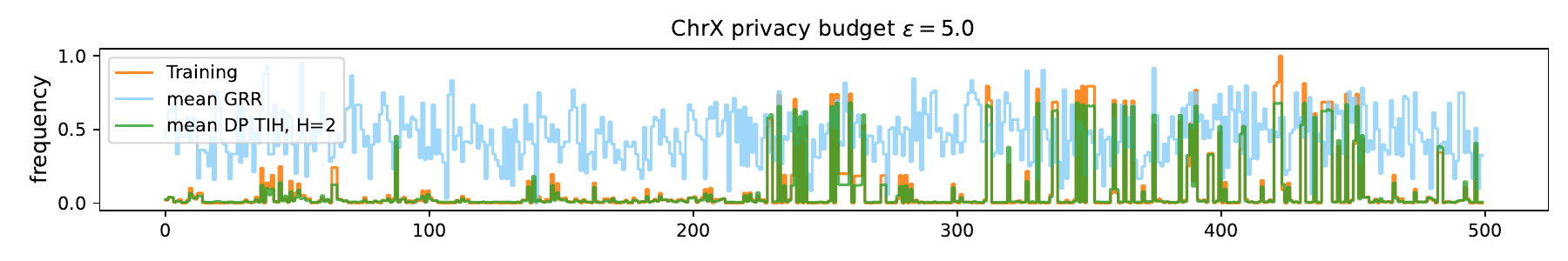}
    \includegraphics[width=\textwidth]{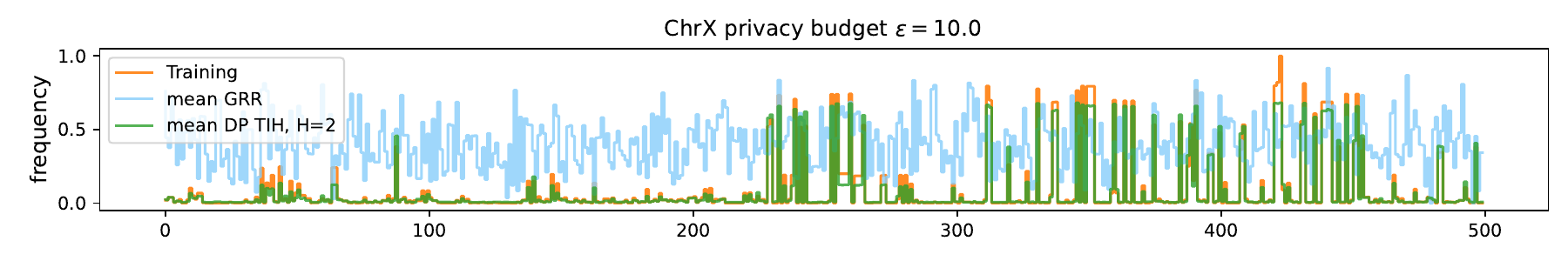}
    \includegraphics[width=\textwidth]{figs/frequencies_chrX_epsilon1.0.pdf}
    \includegraphics[width=\textwidth]{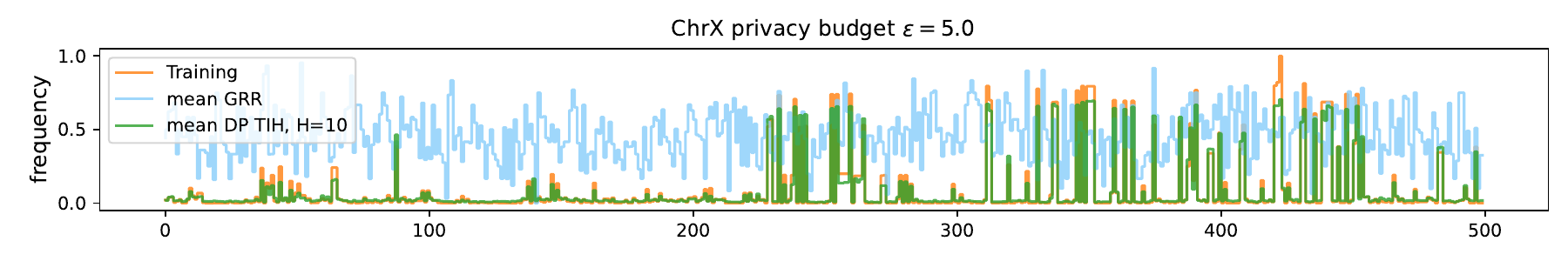}
    \includegraphics[width=\textwidth]{figs/frequencies_chrX_epsilon10.0.pdf}
    \caption{Minor allele frequencies from the real training dataset (chromosome X) vs the generated samples from the DP-trained time-inhomogeneous HMM vs the GRR baseline, for SNP sequence length of 500.}
    \label{fig:chrx-all-minor-allele-frequencies}
\end{figure*}

\begin{figure*}
    \centering
    \includegraphics[width=\textwidth]{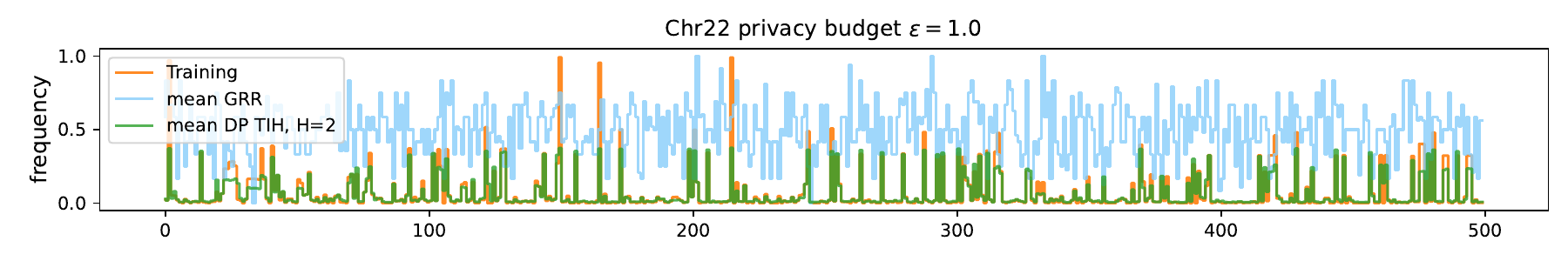}
    \includegraphics[width=\textwidth]{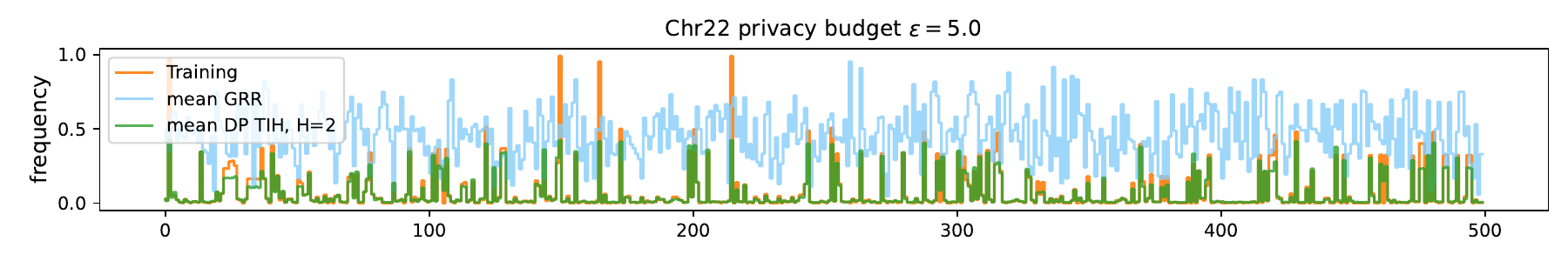}
    \includegraphics[width=\textwidth]{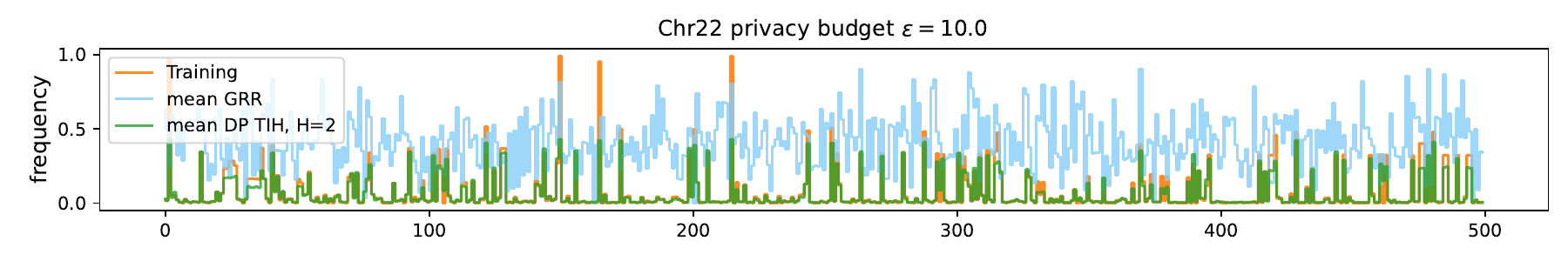}
    \includegraphics[width=\textwidth]{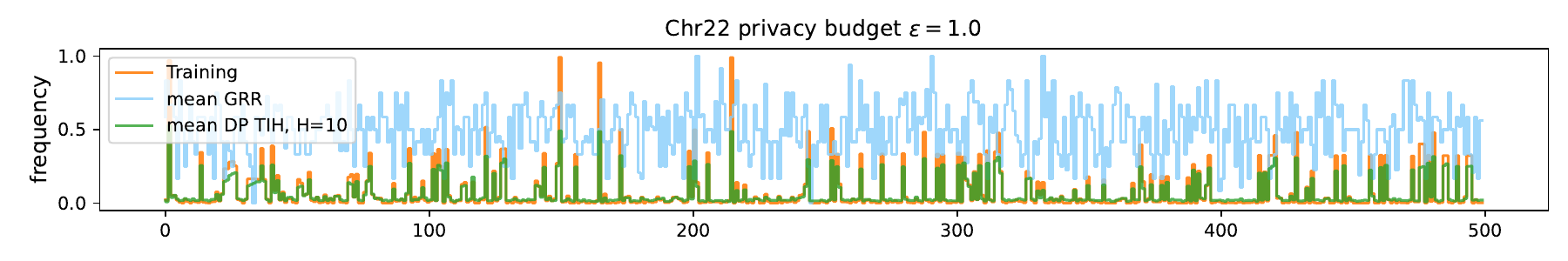}
    \includegraphics[width=\textwidth]{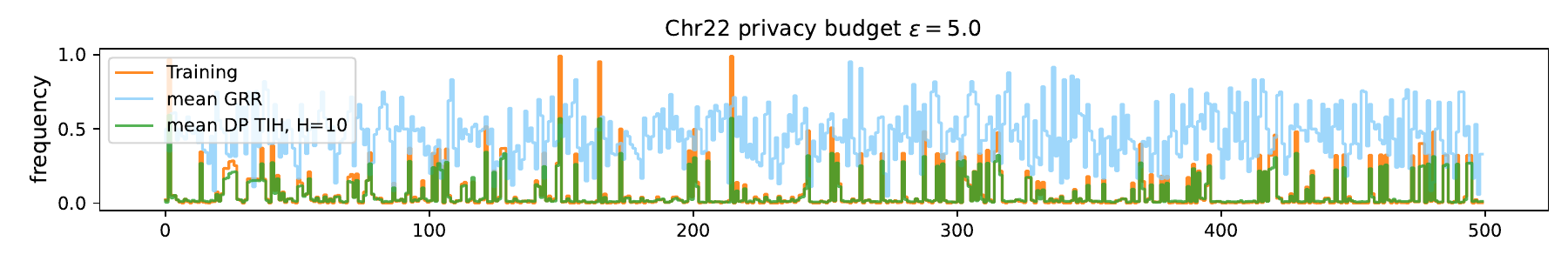}
    \includegraphics[width=\textwidth]{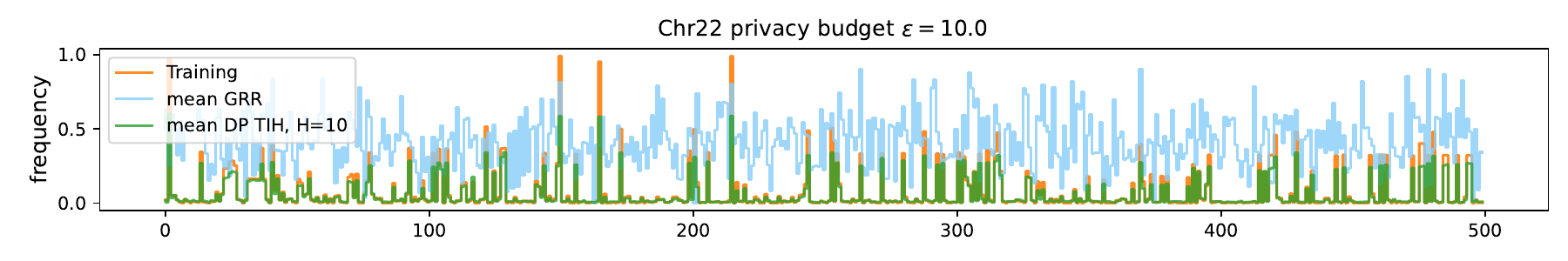}
    \caption{Minor allele frequencies from the real training dataset (chromosome 22) vs the generated samples from the DP-trained time-inhomogeneous HMM vs the GRR baseline, for SNP sequence length of 500.}
    \label{fig:chr22-all-minor-allele-frequencies}
\end{figure*}

\myparagraph{Minor allele frequencies} Figure~\ref{fig:chrx-all-minor-allele-frequencies} and Figure~\ref{fig:chr22-all-minor-allele-frequencies} present the minor allele frequencies at each SNP locus for the first 500 SNPs. For the GRR model, the reported results correspond to average allele frequencies computed over three random runs. For the TIH model, we similarly report averages across three random runs, each based on 2000 generated samples for $H=2$ and $H=10$.

The GRR baseline fails to produce meaningful results, with allele frequencies resembling random noise. Under stronger privacy constraints ($\varepsilon=1.0$), the TIH model exhibits an averaging effect: rather than reproducing sharp peaks and troughs in the frequency spectrum, the signal is smoothed toward intermediate values. This effect is particularly pronounced for TIH with $H=10$, as the DP mechanism has a stronger impact on the larger model compared to $H=2$. In contrast, at the weaker privacy setting ($\varepsilon=10$), the more complex model ($H=10$) demonstrates improved fidelity, capturing specific peaks more accurately than the smaller variant.

\myparagraph{Time complexity}
We also report the average times it takes to train the TIH via DP-SGD in Table~\ref{tab:time-complexity-chrx}. We use a single NVIDIA TITAN RTX GPU with 24GB of available memory. We see that for the longest sequence length $L=500$, we need less than 1 GPU hour to train our model.
{\small
\begin{table}[h]
    \centering
    \caption{Average times for training (chromosome X) of the TIH models over three random runs and three privacy levels, in seconds.}
    \begin{tabular}{|c|c|c|c|c|c|}
        \hline
        &H=1 & H=2 & H=10 & H=50 & H=100 \\ 
        \hline
        L=100& 270 & 558 & 475 & 706 & 1198 \\ 
        L=200& 533 & 1099 & 927 & 1511 & - \\ 
        L=500& 1338 & 2766 & 3171 & - & - \\ 
        \hline
    \end{tabular}
    \label{tab:time-complexity-chrx}
\end{table}
}

%%%%%%%%%%%%%%%%%%%%%%%%%%%%%%%%%%%%%%%%%%%%%%%%%%%%%%%
%%%%%%%%%%%%%%%%%%%%%%%%%%%%%%%%%%%%%%%%%%%%%%%%%%%%%%%%%%
%%%%%%%%%%%%%%%%%%%%%%%%%%%%%%%%%%%%%%%%%%%%%%%%%%%%%%%%
\section{GWAS Downstream Task}
\label{sec:app-downstream}

\begin{table}[ht!]
\caption{$p-$values for the top-$k$ SNPs for chromosome~X and chromosome~22.}
\label{tab:pvalues-chrX-chr22}
\centering
\begin{tabular}{c| c c}
\hline
$k$ & chrX & chr22 \\
\hline
1  & $6\times10^{-96}$ & $8\times10^{-74}$ \\
2  & $6\times10^{-74}$ & $9\times10^{-74}$ \\
3  & $2\times10^{-50}$ & $1\times10^{-73}$ \\
4  & $2\times10^{-48}$ & $3\times10^{-73}$ \\
5  & $4\times10^{-43}$ & $4\times10^{-73}$ \\
6  & $6\times10^{-41}$ & $2\times10^{-72}$ \\
7  & $3\times10^{-29}$ & $2\times10^{-72}$ \\
8  & $4\times10^{-24}$ & $3\times10^{-71}$ \\
9  & $5\times10^{-24}$ & $4\times10^{-71}$ \\
10 & $7\times10^{-22}$ & $4\times10^{-71}$ \\
11 & $3\times10^{-17}$ & $3\times10^{-67}$ \\
12 & $1\times10^{-18}$ & $5\times10^{-65}$ \\
13 & $1\times10^{-16}$ & $7\times10^{-59}$ \\
14 & $4\times10^{-16}$ & $8\times10^{-52}$ \\
15 & $4\times10^{-16}$ & $7\times10^{-49}$ \\
\hline
\end{tabular}
\end{table}

\begin{figure*}[ht!]
    \centering
    \hspace{-0.cm}
    \begin{minipage}[b]{0.45\textwidth}
        \centering
        \includegraphics[width=0.8\textwidth]{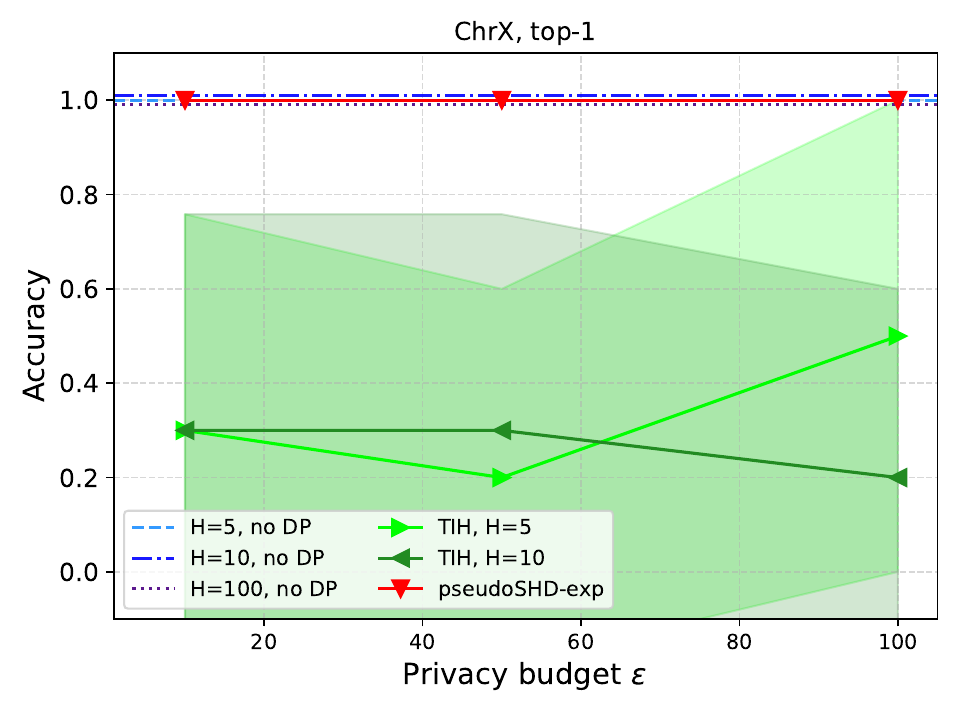}
    \end{minipage}%
    \hspace{-0.cm}
    \begin{minipage}[b]{0.45\textwidth}
        \centering
        \includegraphics[width=0.8\textwidth]{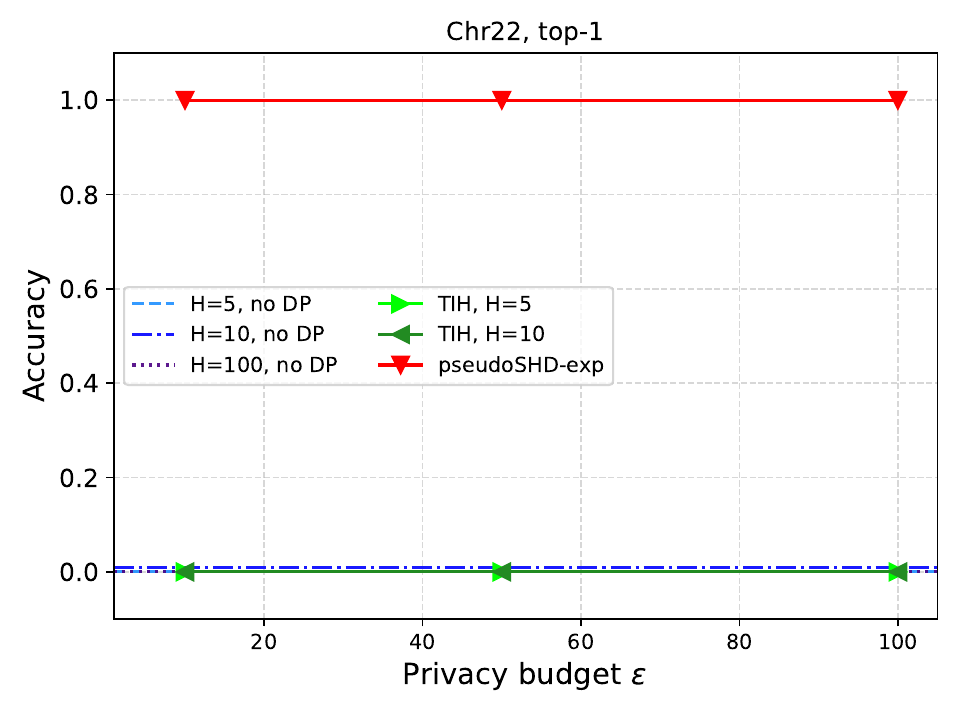}
    \end{minipage}%
    \hspace{-0.cm}
    \begin{minipage}[b]{0.45\textwidth}
        \centering
        \includegraphics[width=0.8\textwidth]{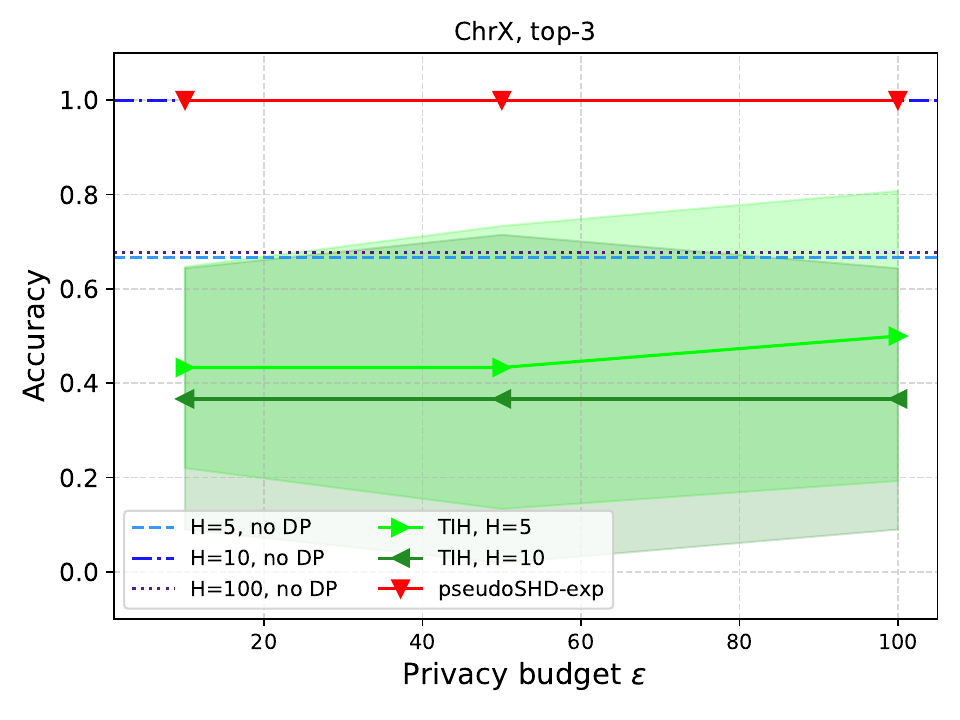}
    \end{minipage}
    \hspace{-0.cm}
    \begin{minipage}[b]{0.45\textwidth}
        \centering
        \includegraphics[width=0.8\textwidth]{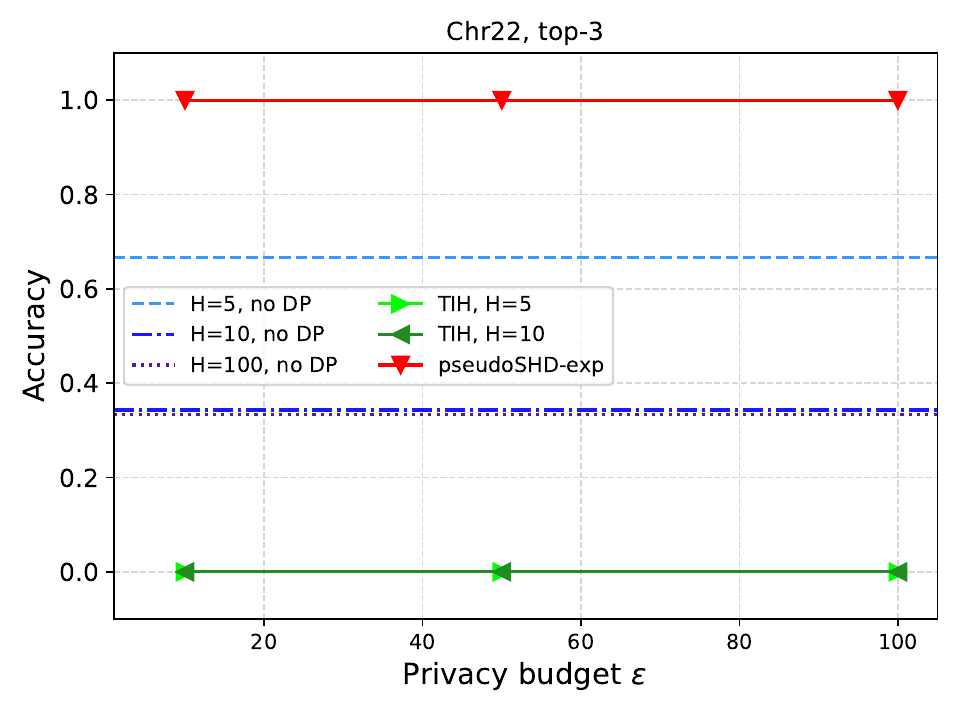}
    \end{minipage}%
    \hspace{-0.cm}
    \begin{minipage}[b]{0.45\textwidth}
        \centering
        \includegraphics[width=0.8\textwidth]{figs/chrX_top5.pdf}
    \end{minipage}%
    \hspace{-0.cm}
    \begin{minipage}[b]{0.45\textwidth}
        \centering
        \includegraphics[width=0.8\textwidth]{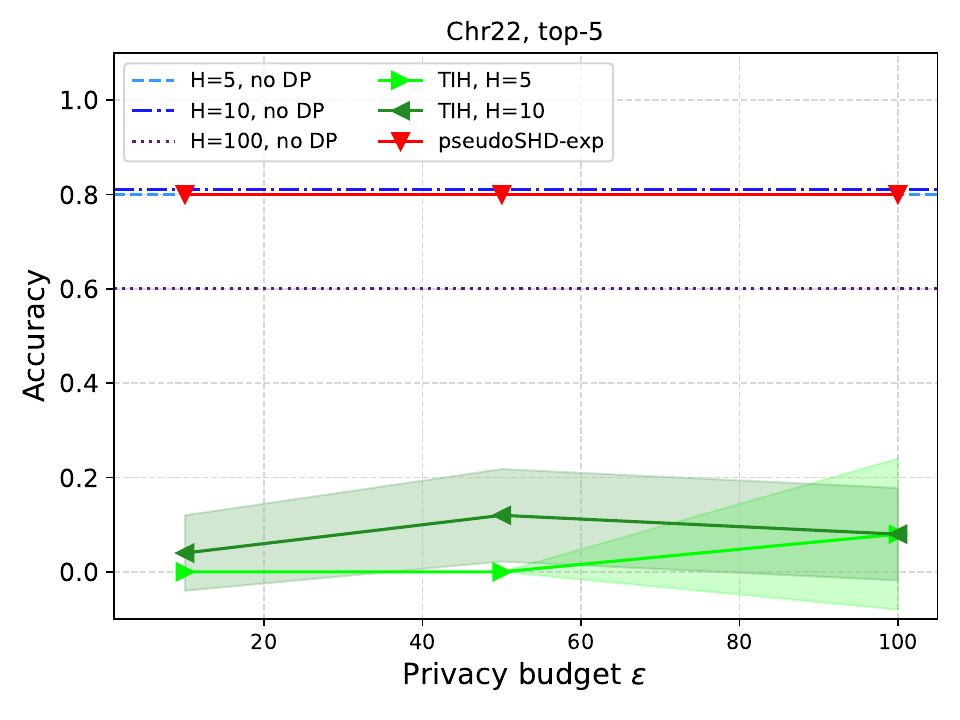}
    \end{minipage}
    \hspace{-0.cm}
    \begin{minipage}[b]{0.45\textwidth}
        \centering
        \includegraphics[width=0.8\textwidth]{figs/chrX_top10.pdf}
    \end{minipage}%
    \hspace{-0.cm}
    \begin{minipage}[b]{0.45\textwidth}
        \centering
        \includegraphics[width=0.8\textwidth]{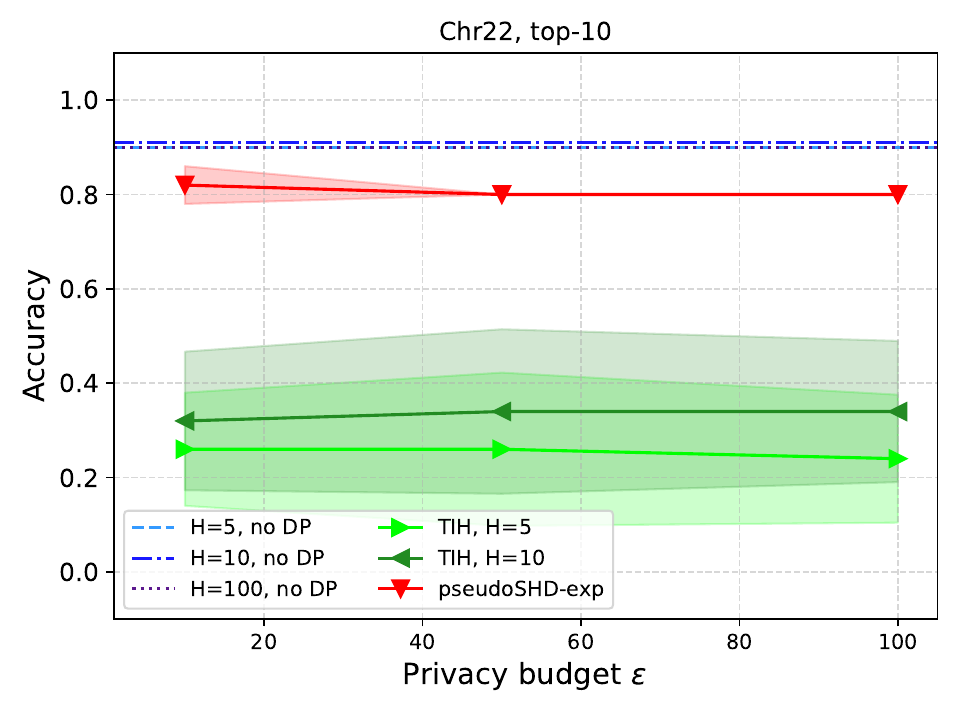}
    \end{minipage}
    \caption{Averaged accuracies of returning the top-$k$ associated SNPs between case and control group. the shaded area shows the standard deviation over random runs of the DP methods.}
    \label{fig:all-topk}
\end{figure*}

In this section, we present the complementary experimental results corresponding to Section~\ref{sec:gwas-downstream}. Figure~\ref{fig:all-topk} reports the accuracy of identifying the top-$k$ SNPs for $k \in \{1, 3, 5, 10\}$ across chromosomes~X and~22.  

Overall, the TIH model exhibits stronger performance on chromosome~X compared to chromosome~22. Specifically, all non-private TIH models fail to recover the top-1 SNP on chromosome~22, while the DP-trained TIH models show reduced performance relative to their counterparts trained on chromosome~X. These results indicate that chromosome~22 poses additional challenges, warranting further investigation.  

\begin{figure*}[ht!]
    \centering
    \hspace{-0.cm}
    \begin{minipage}[b]{0.5\textwidth}
        \centering
        \includegraphics[width=\textwidth]{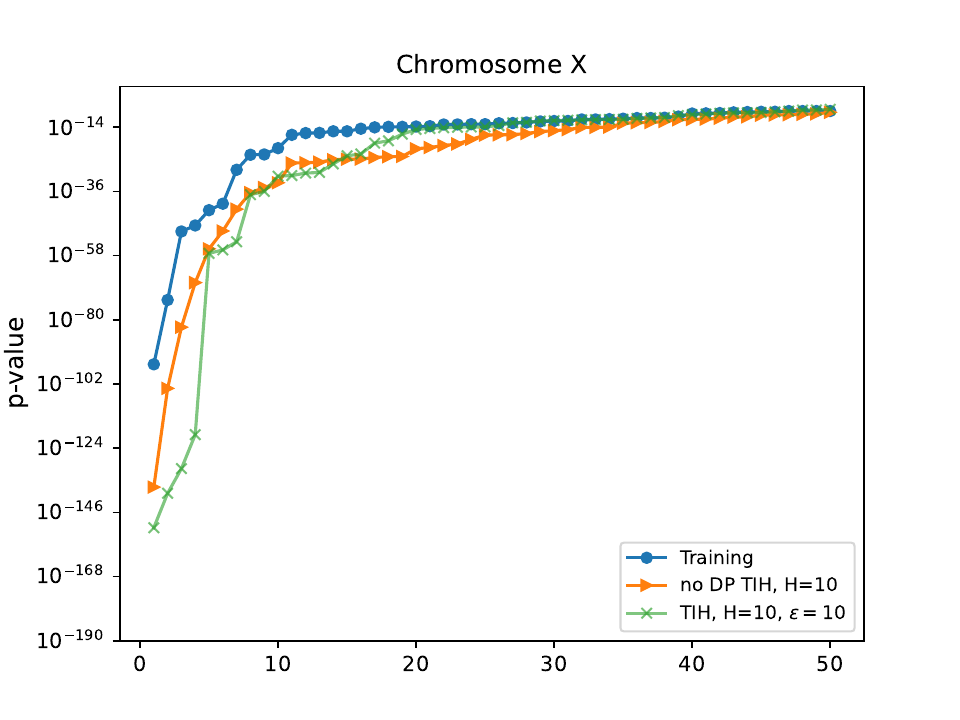}
    \end{minipage}%
    \hspace{-0.5cm}
    \begin{minipage}[b]{0.5\textwidth}
        \centering
        \includegraphics[width=\textwidth]{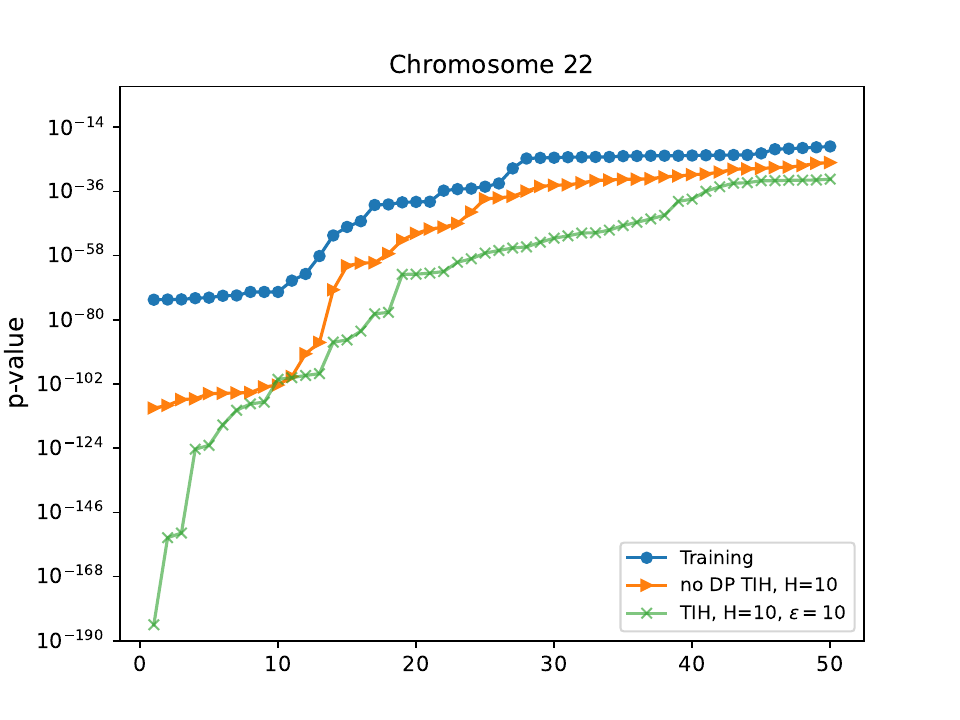}
    \end{minipage}%
    \caption{$p-$value distribution for top-$k$ associated SNPs of both training datasets.}
    \label{fig:pvalues}
\end{figure*}

To analyze this effect in more detail, we plot in Figure~\ref{fig:pvalues} the distribution of $p$-values for the top-50 SNPs obtained using the real dataset, samples from the non-private TIH model, and samples from one random run of the DP-trained TIH model. For clarity, the exact values are also provided in Table~\ref{tab:pvalues-chrX-chr22}. The results suggest that the artificial phenotyping mechanism yields more challenging association patterns for chromosome~22. In particular, while the top-1 SNP on chromosome~X displays a distinctly small $p$-value, chromosome~22 exhibits much smaller separations between the $p$-values of its leading SNPs. Consequently, the first few associated SNPs on chromosome~22 appear statistically similar, making it more difficult for the model to replicate the subtle differences between case and control groups.  

Given that our setting assumes a central data holder trains the HMM models with private data and subsequently releases both the models and synthetic datasets, we recommend that diagnostics such as $p$-value distributions be evaluated prior to release. Based on these evaluations, the data holder can provide guidelines regarding the reliability of the synthetic outputs for downstream tasks. For instance, for chromosome~X, the clear separation in $p$-values suggests that the TIH model can reliably recover the top-1, top-3, top-5, and top-10 SNPs. In contrast, the tighter clustering of $p$-values observed for chromosome~22 indicates that the model’s predictions are more reliable only around approximately the top-10 SNPs, before which caution is warranted.  

We note that such diagnostic guidelines must be provided with care. While releasing exact $p$-values or detailed statistics from the private dataset would risk leaking sensitive information, high-level guidance (e.g., specifying that top-$k$ SNPs are more reliable for certain chromosomes) can be reported without compromising privacy. In practice, this form of aggregate recommendation is comparable to publishing utility benchmarks of a DP mechanism and does not reveal individual-level data.

%%%%%%%%%%%%%%%%%%%%%%%%%%%%%%%%%%%%%%%%%%%%%%%%%%%%%%%%
%%%%%%%%%%%%%%%%%%%%%%%%%%%%%%%%%%%%%%%%%%%%%%%%%%%%%%%%
%%%%%%%%%%%%%%%%%%%%%%%%%%%%%%%%%%%%%%%%%%%%%%%%%%%%%%%%
\section{Pairwisse Correlation of SNPs}
\label{sec:app-ld-correlations}

\begin{figure*}[ht!]
    \centering
    \includegraphics[width=0.29\textwidth]{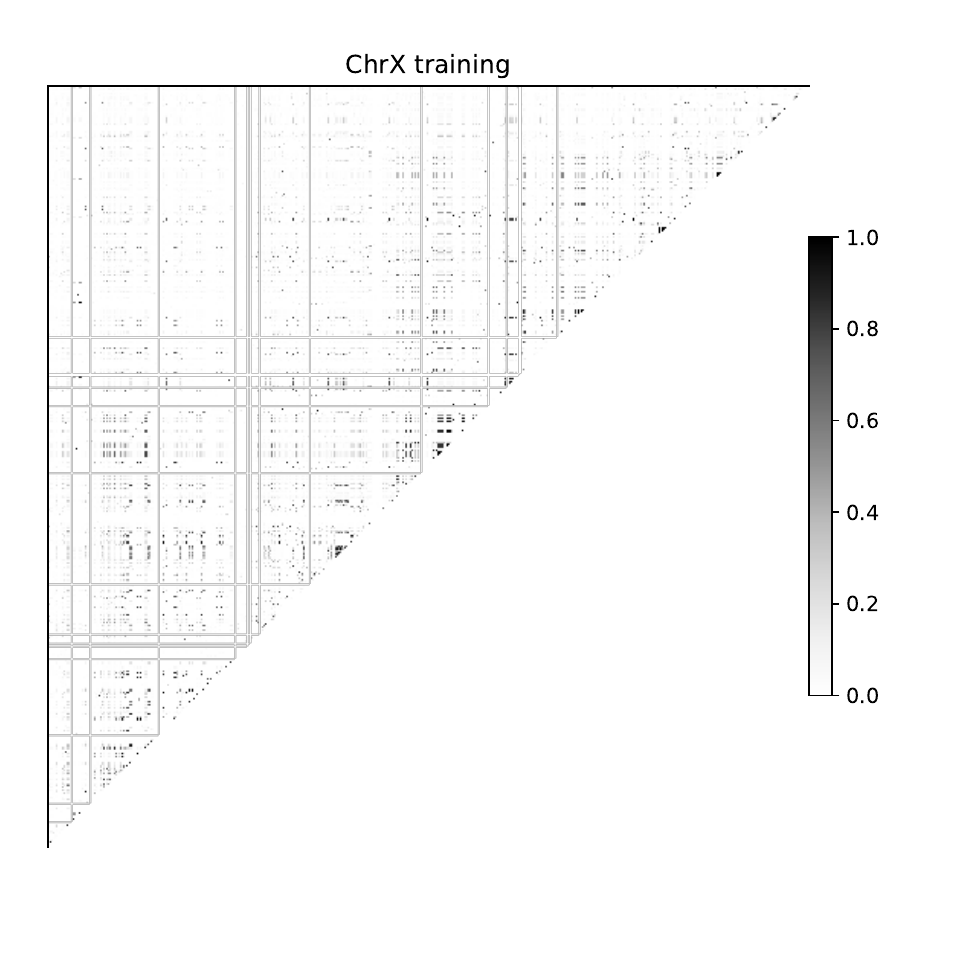}
    \includegraphics[width=0.29\textwidth]{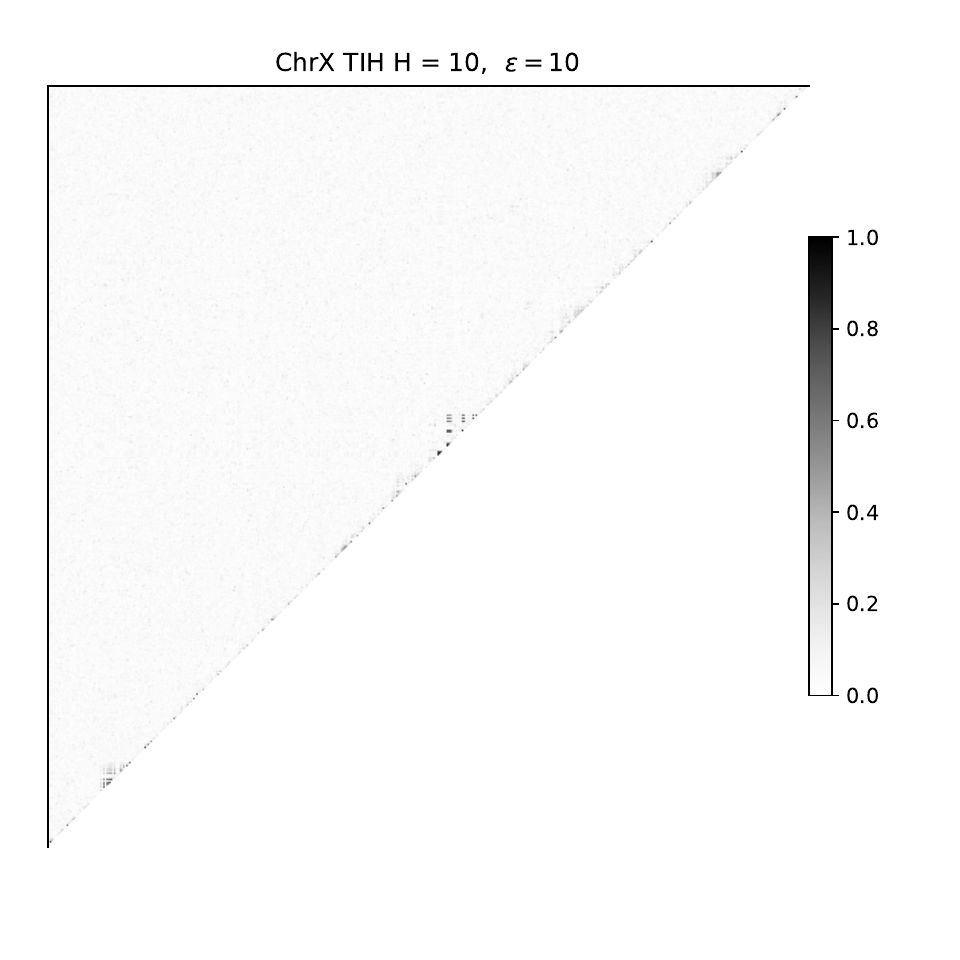}
    \includegraphics[width=0.29\textwidth]{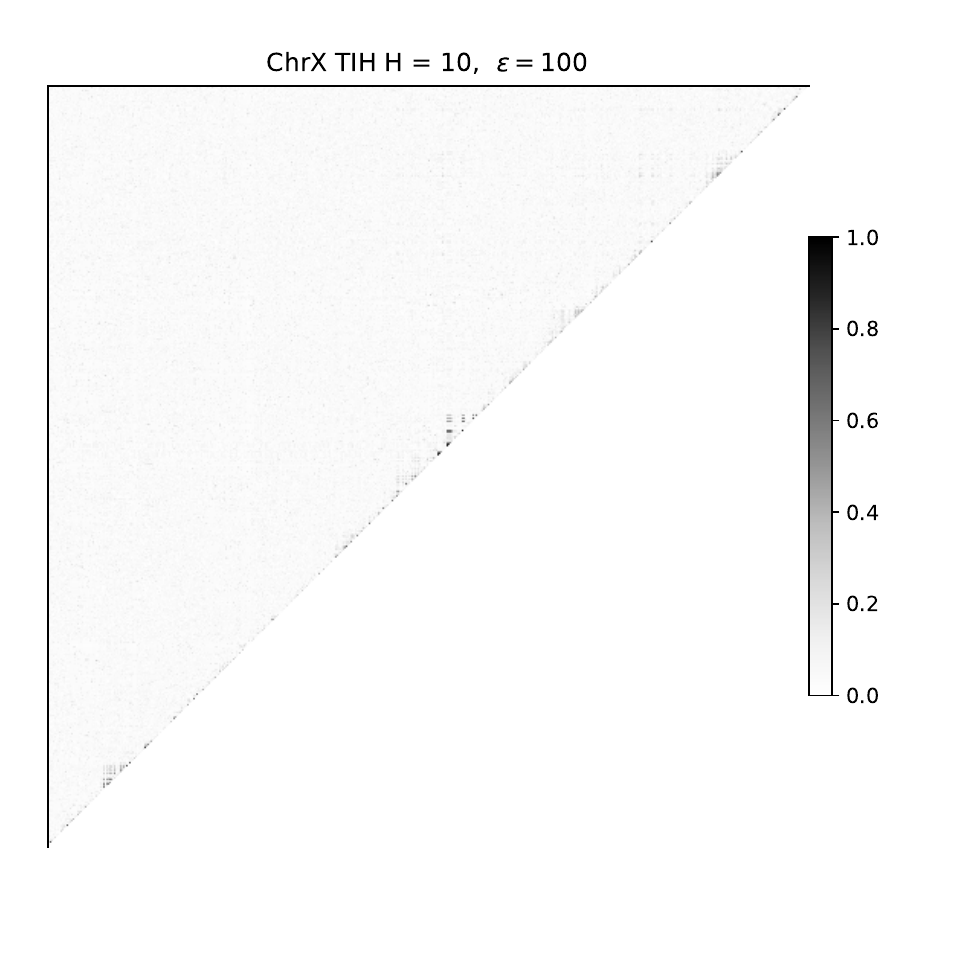}
    \includegraphics[width=0.29\textwidth]{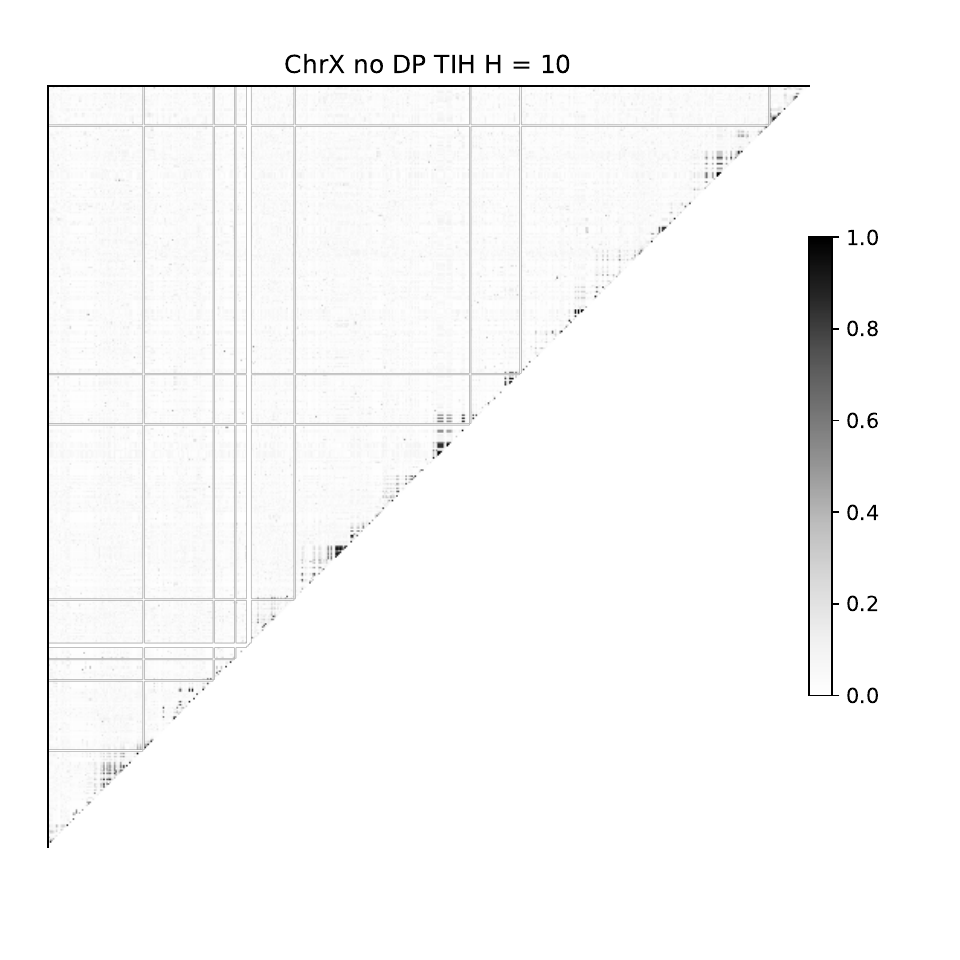}
    \includegraphics[width=0.29\textwidth]{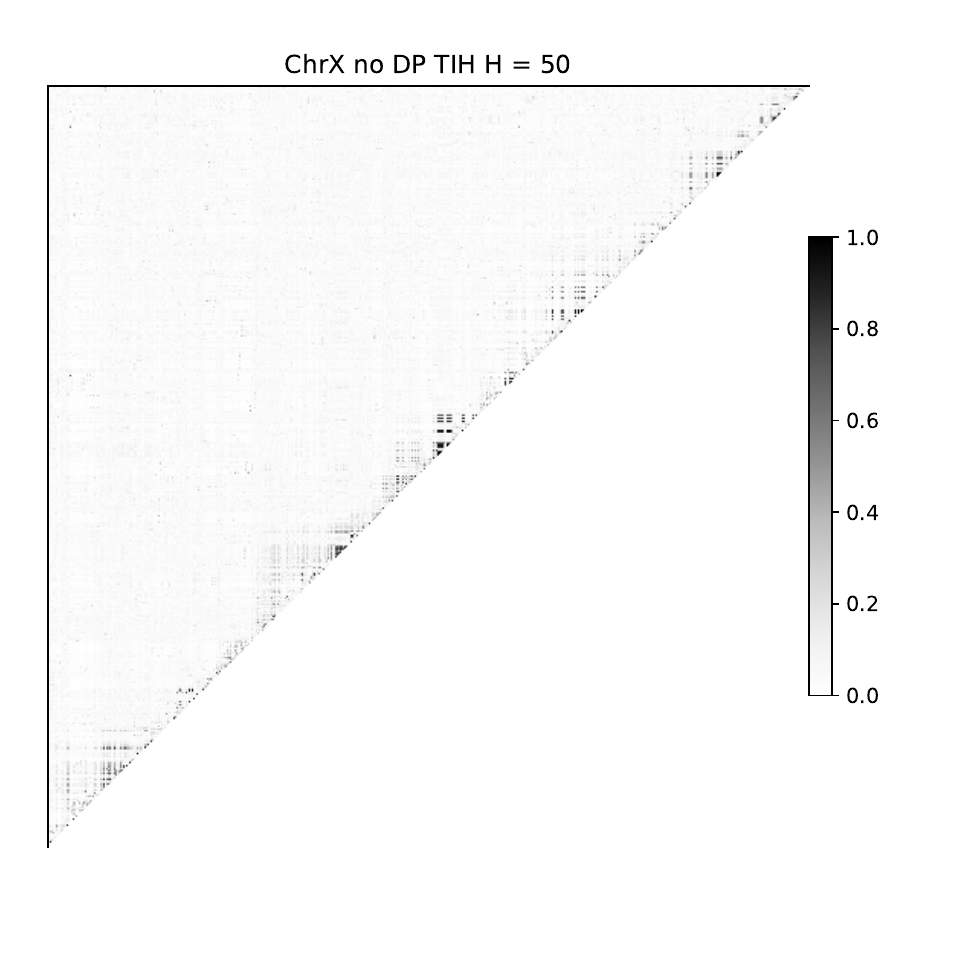}
    \includegraphics[width=0.29\textwidth]{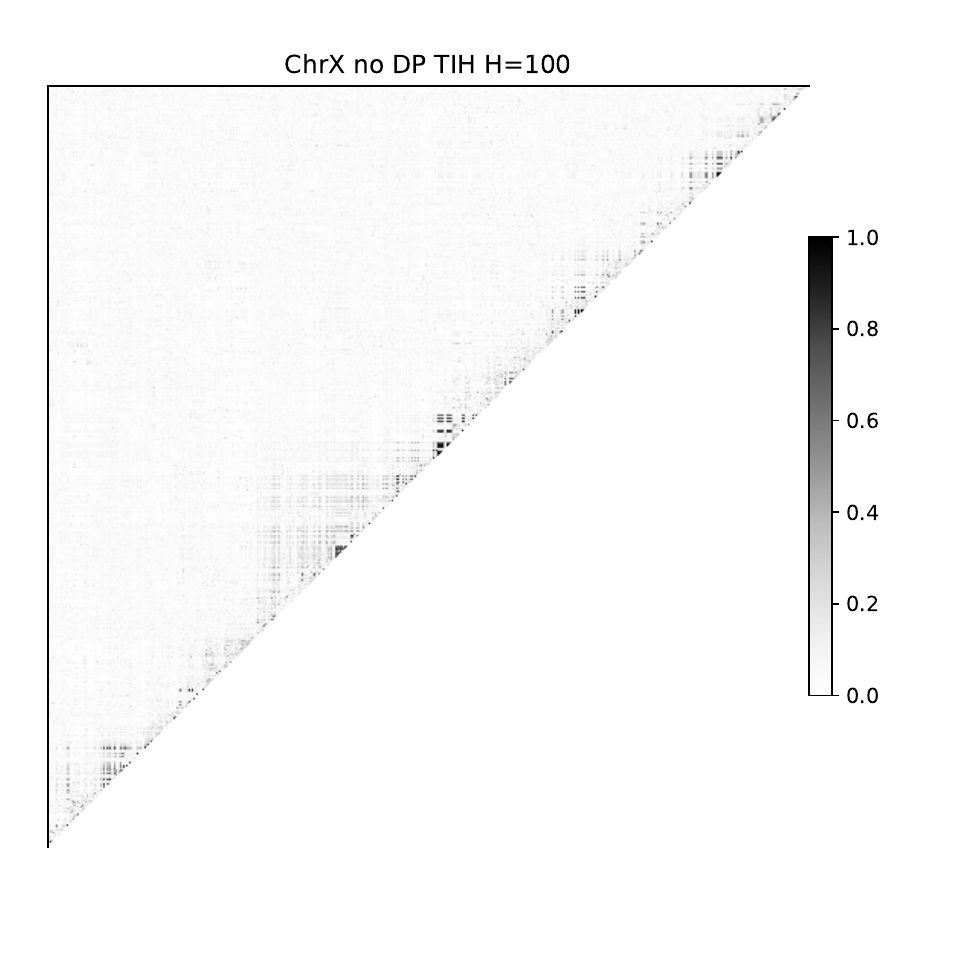}
    \includegraphics[width=0.29\textwidth]{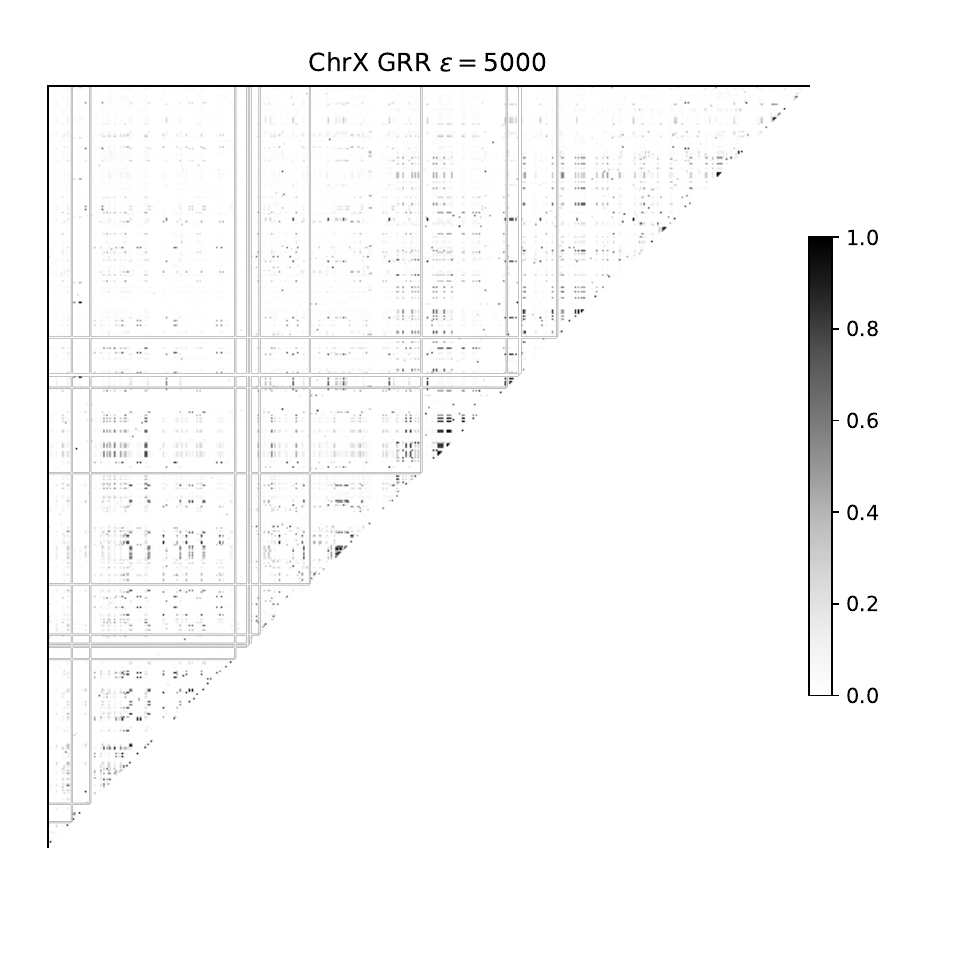}
    \includegraphics[width=0.29\textwidth]{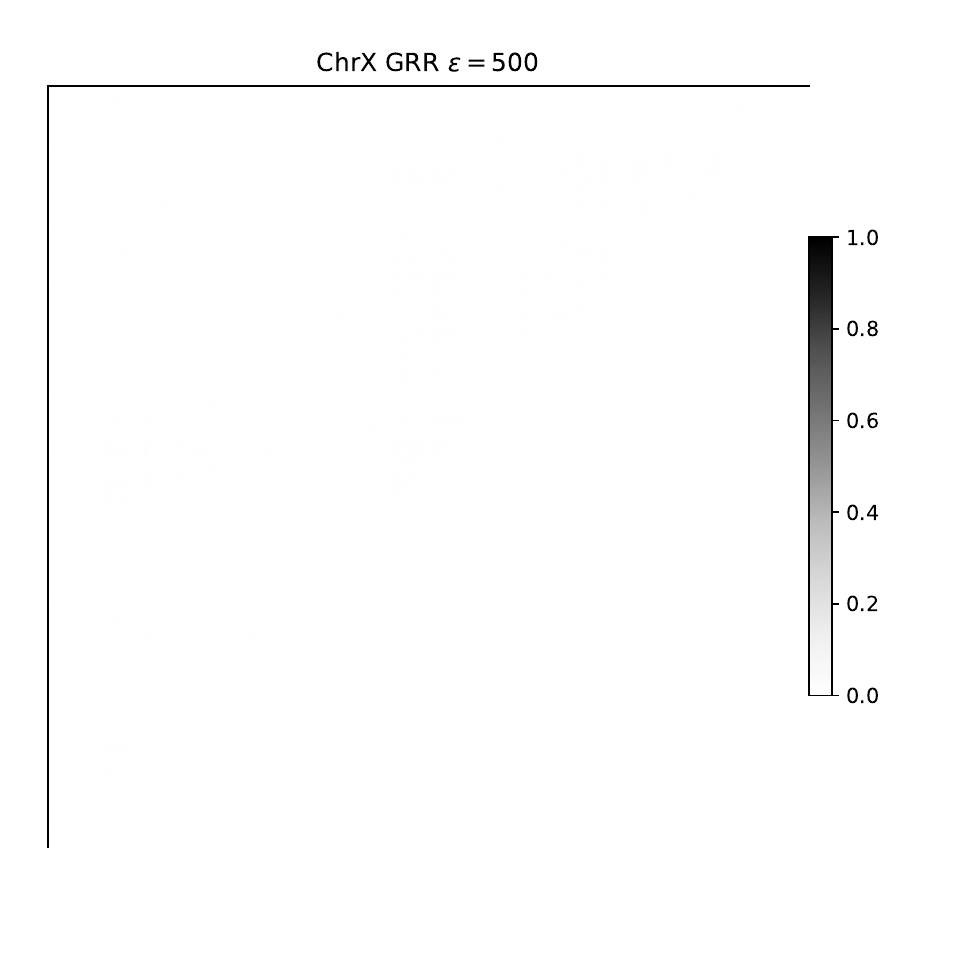}
    \includegraphics[width=0.29\textwidth]{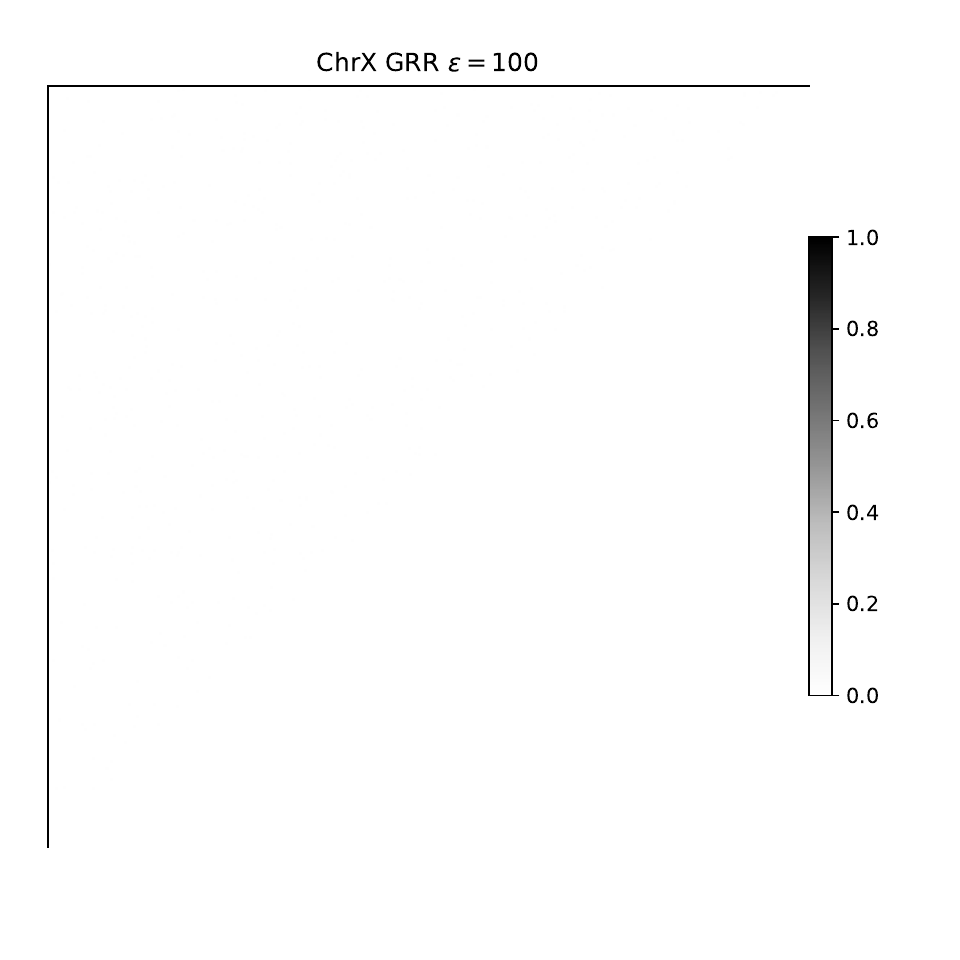}
    \caption{Pairwise LD correlations of the first 500 SNPs for chromosome X.}
    \label{fig:chrX-ld-plots}
\end{figure*}

\begin{figure*}
    \centering
    \includegraphics[width=0.29\textwidth]{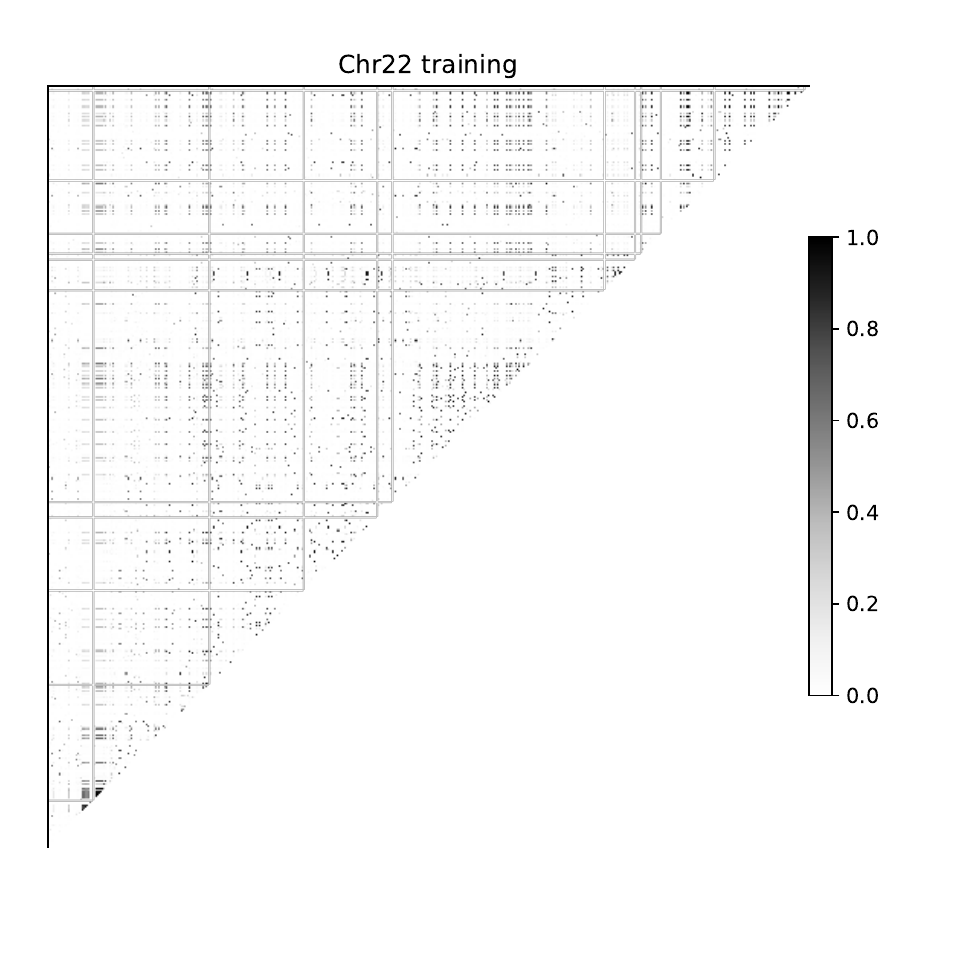}
    \includegraphics[width=0.29\textwidth]{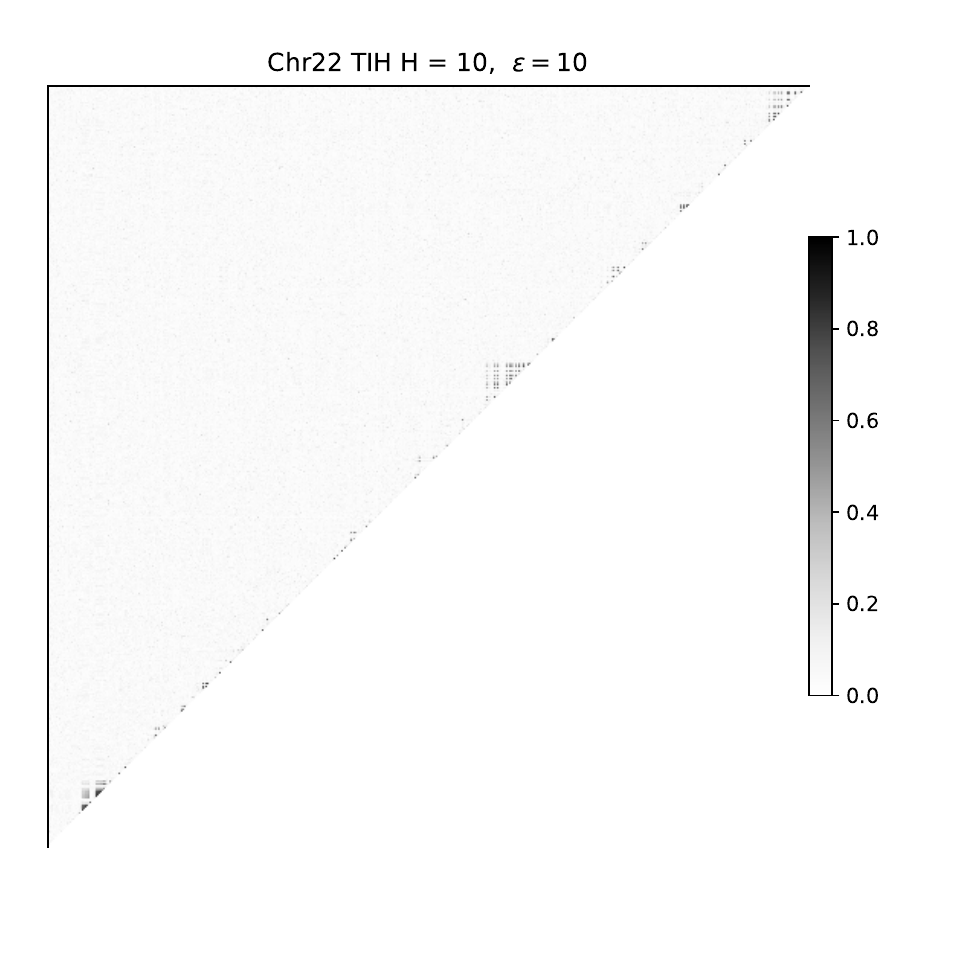}
    \includegraphics[width=0.29\textwidth]{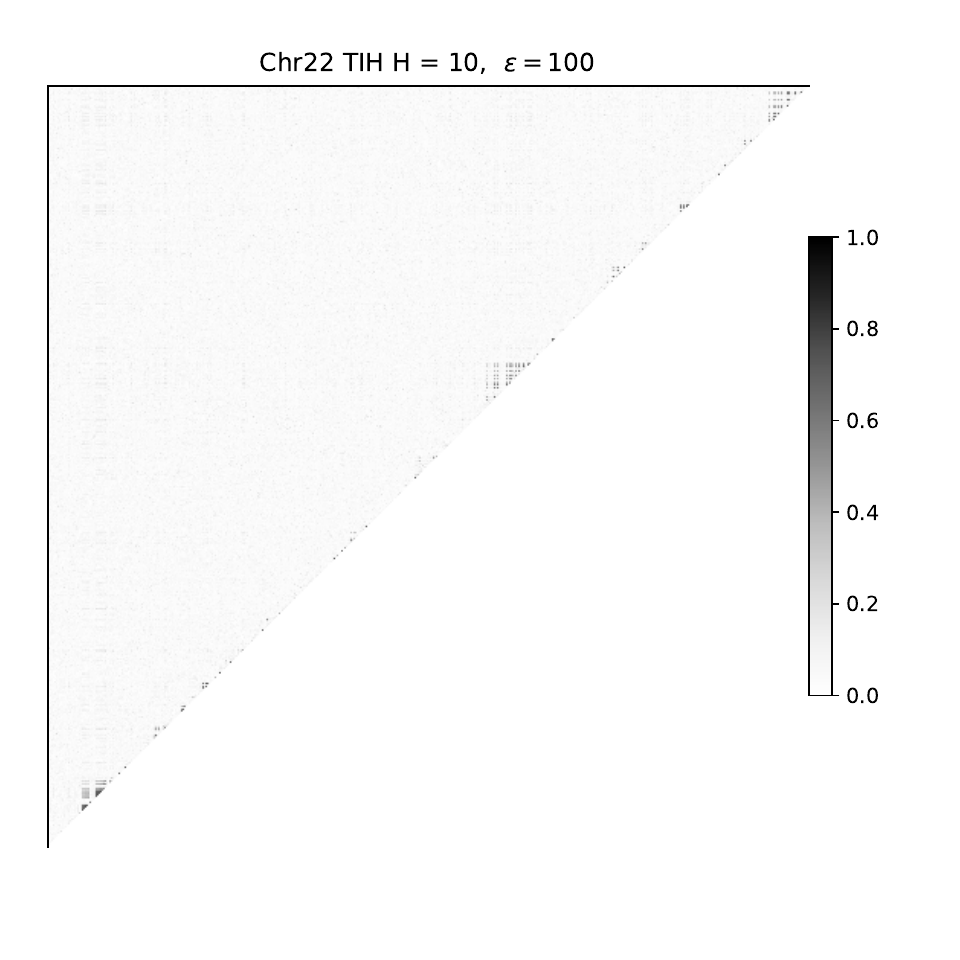}
    \includegraphics[width=0.29\textwidth]{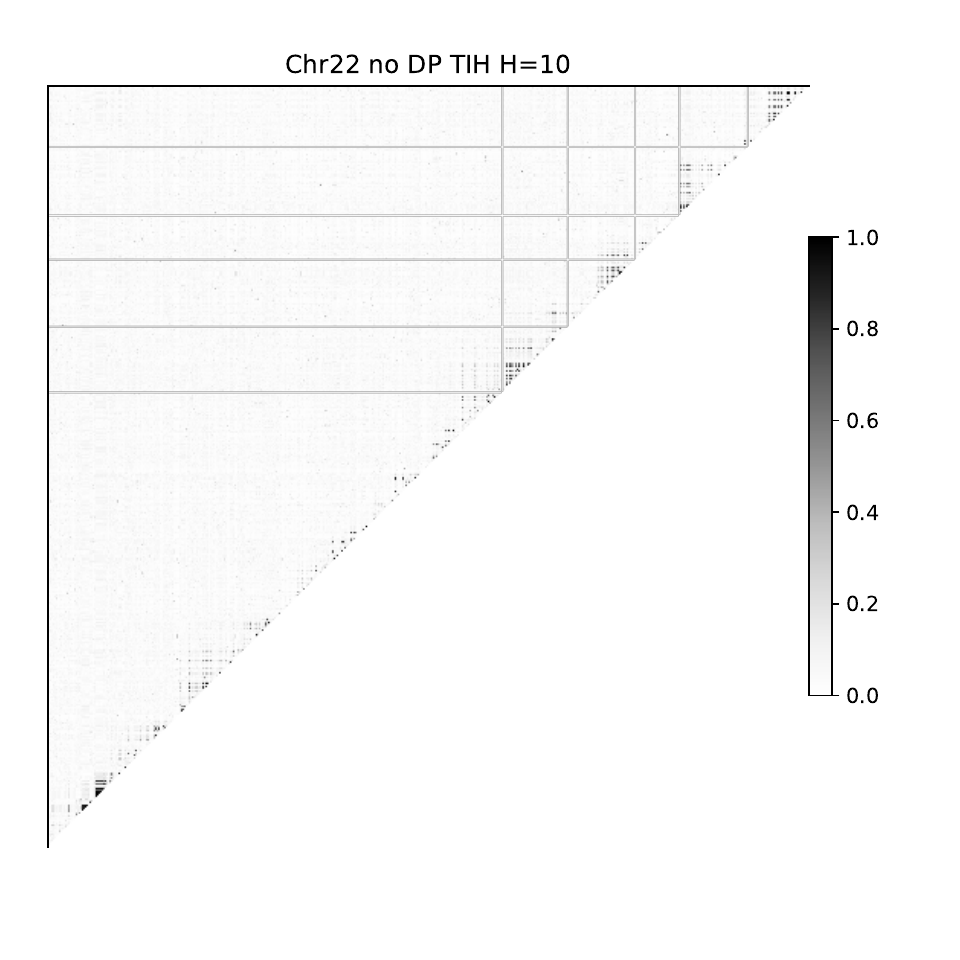}
    \includegraphics[width=0.29\textwidth]{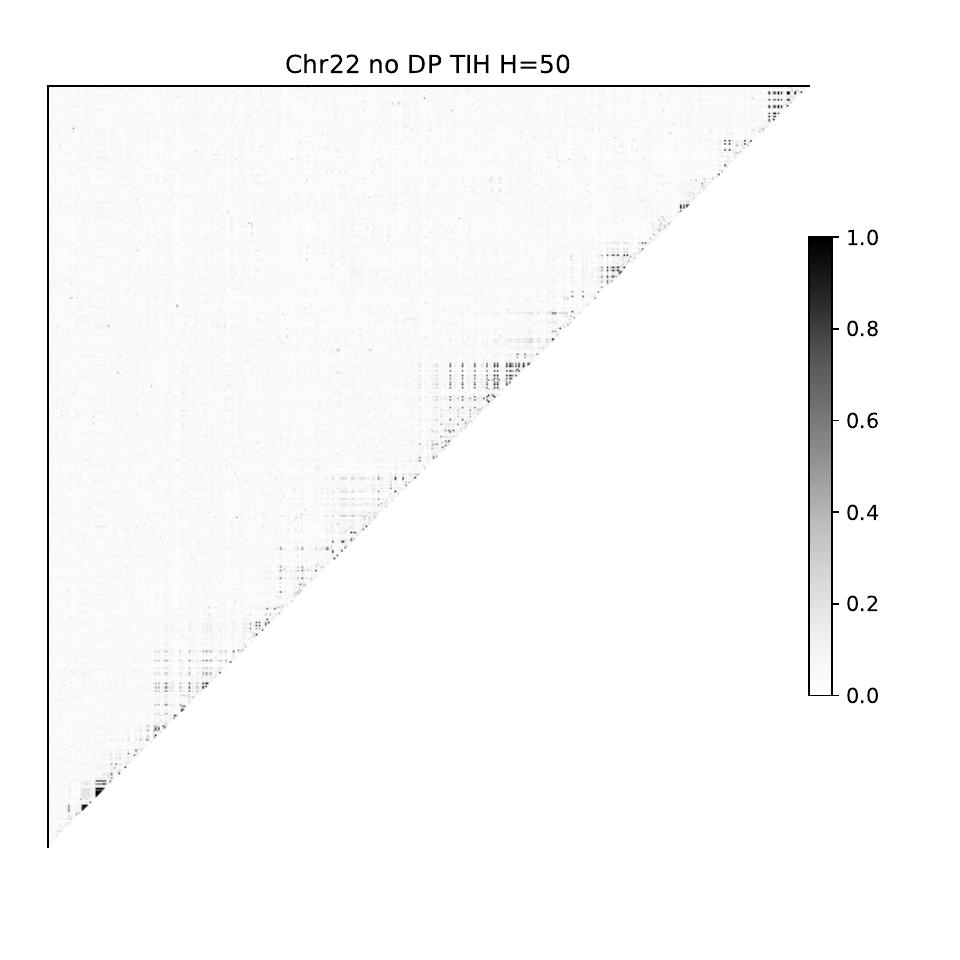}
    \includegraphics[width=0.29\textwidth]{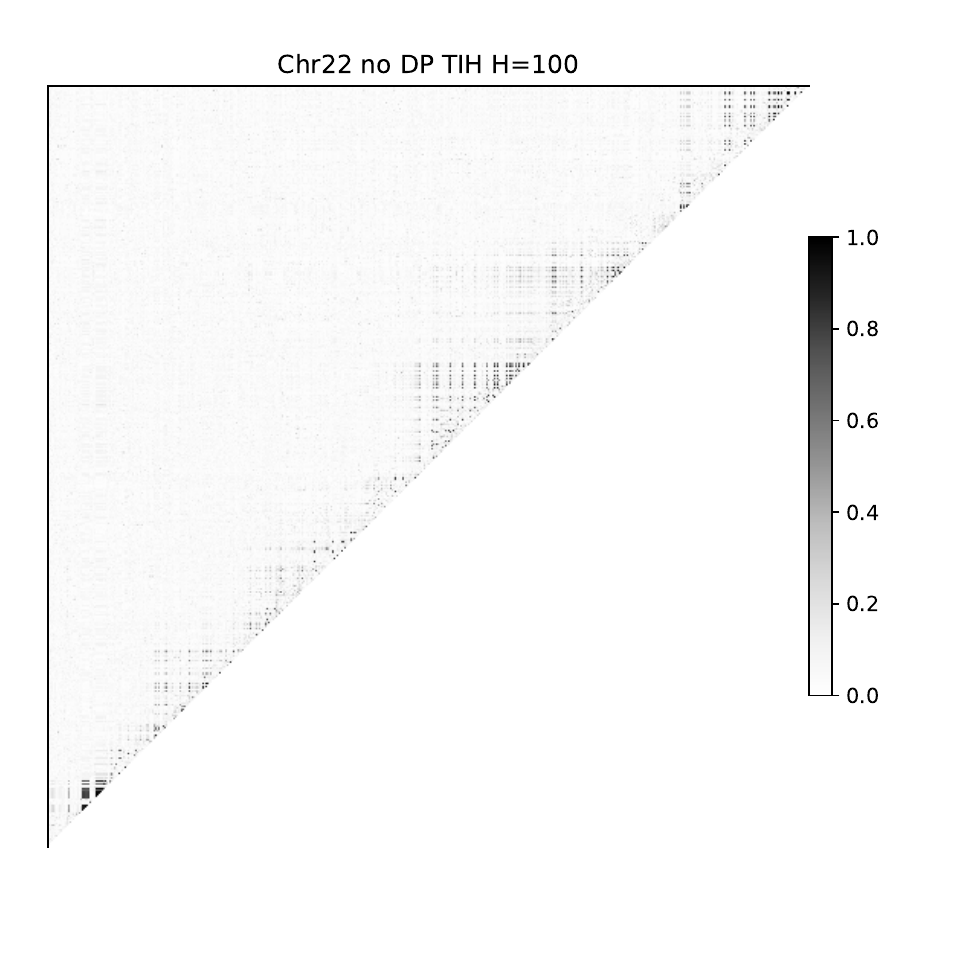}
    \includegraphics[width=0.29\textwidth]{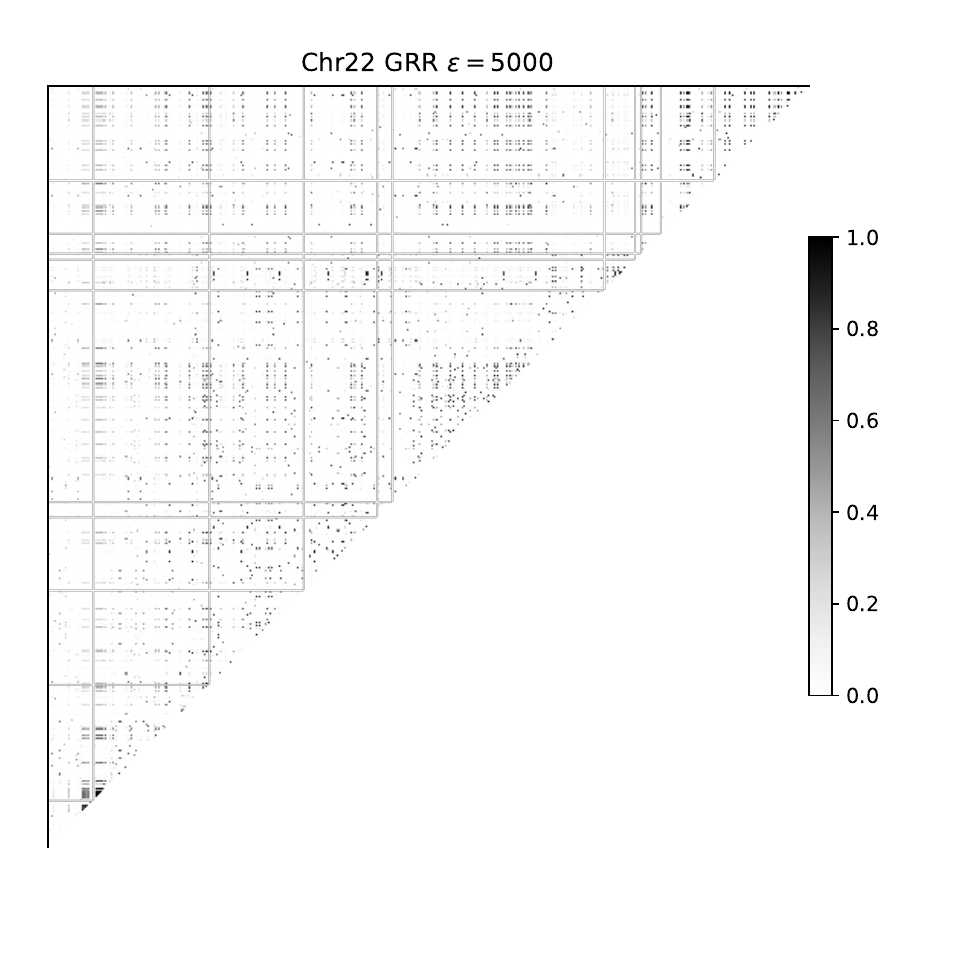}
    \includegraphics[width=0.29\textwidth]{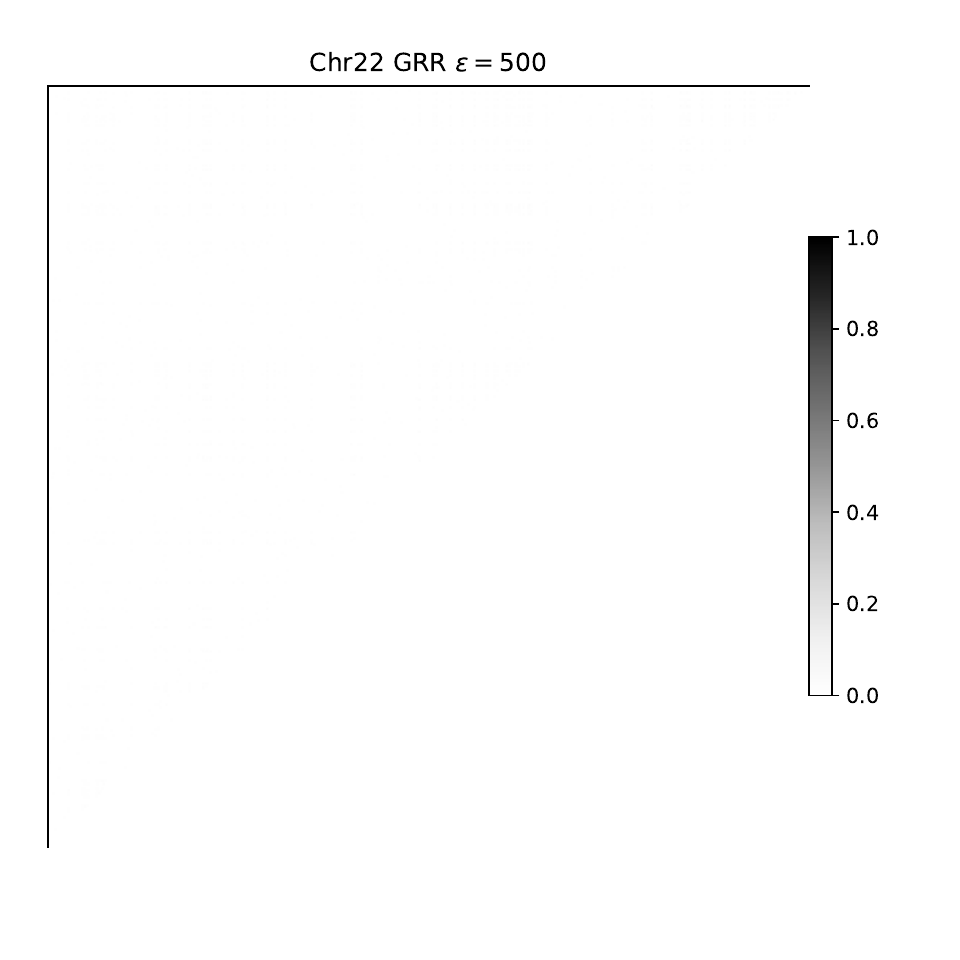}
    \includegraphics[width=0.29\textwidth]{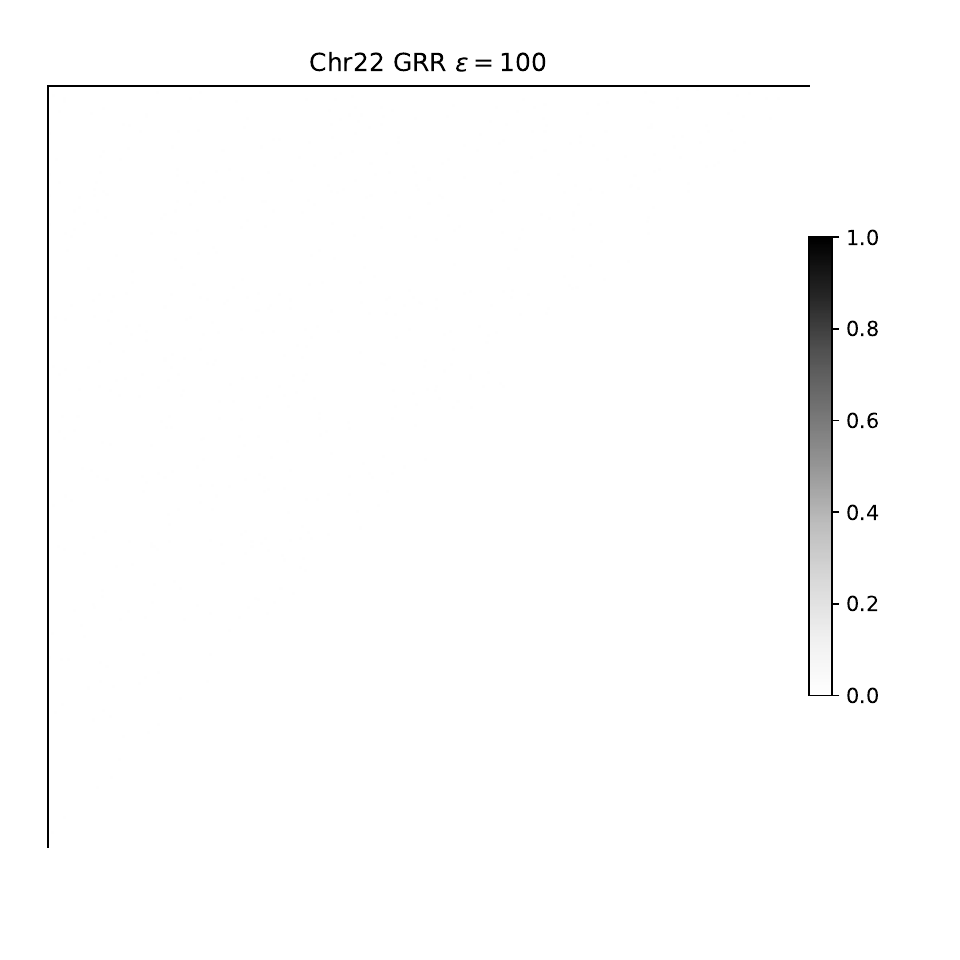}
    \caption{Pairwise LD correlations of the first 500 SNPs for chromosome 22.}
    \label{fig:chr22-ld-plots}
\end{figure*}

Table~\ref{tab:chrx-btss-exact} and Table~\ref{tab:chr22-btss-exact} summarize the results of our correlation-matching experiments, with the corresponding LD panels shown in Figure~\ref{fig:chrX-ld-plots} and Figure~\ref{fig:chr22-ld-plots}. For the DP mechanisms, we plot the results from a single random seed. To improve the visual clarity of the correlation heatmaps, we scale each cell by $(r^2)^{0.4}$ for TIH model results. 

These visualizations highlight that the TIH model consistently preserves short-range, near-diagonal correlations, which are the most prominent features of linkage disequilibrium patterns. However, long-range correlations are not faithfully maintained; which is expected given its reliance on locus-dependent transitions. Extending the model to higher-order Markov dependencies could potentially alleviate this issue by allowing transitions that span more distant loci. 

Interestingly, larger models ($H=50,100$) recover more of the long-range correlation signal. Nevertheless, they do not outperform smaller models in our quantitative similarity metrics (BTSS and exact match rate). A likely explanation is that the larger models trade off local accuracy for global structure. By spreading capacity to capture distant correlations, they reduce their fidelity in reconstructing the very close, near-diagonal correlations that dominate the evaluation metrics. In other words, smaller models achieve higher apparent performance by specializing in local LD, whereas larger models spread capacity across both local and distal signals, lowering their scores under certain metrics. We further observe that relaxing the privacy constraint to $\epsilon=100$ does not yield systematic improvements. Importantly, the imperfect preservation of complex correlation patterns in DP-trained models implies that state-of-the-art membership inference and reconstruction attacks~\cite{deznabi2017inference}, which rely on LD patterns, are unlikely to succeed on our private synthetic datasets.

\begin{table}[h!]
\centering
\caption{Comparison of BTSS and Exact Match for different mechanisms for training chromosome X.}
\begin{tabular}{lcc}
\hline
 & BTSS & Exact Match \\
\hline
GRR $\varepsilon=100$ &  $0.20\pm 0.01$&  $0.13 \pm 0.01$\\
GRR $\varepsilon=500$ &  $0.23\pm 0.01$&  $0.25 \pm 0.01$\\
GRR $\varepsilon=5000$ &  $0.97\pm 0.01$&  $0.96 \pm 0.01$\\
TIH $H=10,$ no DP & $0.67\pm 0.00$ & $0.61 \pm 0.00$ \\
TIH $H=50,$ no DP & $0.54 \pm 0.00$ & $0.46 \pm 0.00$ \\
TIH $H=100,$ no DP & $0.50 \pm 0.00$ & $0.45 \pm 0.00$ \\
TIH $H=10, \varepsilon=10$ & $0.34 \pm 0.01$ & $0.31\pm 0.01$ \\
TIH $H=10, \varepsilon=100$ & $0.35\pm 0.01$ & $0.31 \pm 0.02$ \\
\hline
\end{tabular}
\label{tab:chrx-btss-exact}
\end{table}

\begin{table}[h!]
\centering
\caption{Comparison of BTSS and Exact Match for different mechanisms for training chromosome 22.}
\begin{tabular}{lcc}
\hline
 & BTSS & Exact Match \\
\hline
GRR $\varepsilon=100$ &  $0.24\pm 0.01$&  $0.13 \pm 0.02$\\
GRR $\varepsilon=500$ &  $0.27\pm 0.01$&  $0.25 \pm 0.02$\\
GRR $\varepsilon=5000$ &  $0.98\pm 0.01$&  $0.98 \pm 0.01$\\
TIH $H=10,$ no DP & $0.66\pm 0.00$ & $0.60 \pm 0.00$ \\
TIH $H=50,$ no DP & $0.55 \pm 0.00$ & $0.45 \pm 0.00$ \\
TIH $H=100,$ no DP & $0.53 \pm 0.00$ & $0.44 \pm 0.00$ \\
TIH $H=10, \varepsilon=10$ & $0.41\pm 0.03$ & $0.34\pm 0.04$ \\
TIH $H=10, \varepsilon=100$ & $0.38 \pm 0.02$ & $0.32\pm 0.03$ \\
\hline
\end{tabular}
\label{tab:chr22-btss-exact}
\end{table}

